\documentclass[pdflatex]{sn-jnl}

\jyear{2022}%

\usepackage{parskip}
\usepackage{rotating} 
\usepackage{longtable}
\usepackage{pdflscape}
\usepackage[T1]{fontenc}
\usepackage{mathptmx}
\usepackage{newtxmath}\usepackage[scaled=.78]{noto-mono}\usepackage[scaled=.88]{noto-sans} 
\usepackage{multirow}
\usepackage{array}
\usepackage{enumitem}
\usepackage{graphicx}
\usepackage{subfigure}

\usepackage{xltabular}

\usepackage{fancyhdr} 
\usepackage{url}
\usepackage{natbib}
\setcitestyle{numbers,square,comma} 

\definecolor{paleyellow}{HTML}{FFEC7F}

\long\def\comment[#1]#2{\par\colorbox{paleyellow}{\llap{\textbf{#1:\quad}}%
    \parbox[t]{\textwidth}{\setlength{\parskip}{1ex plus 0.2ex minus 0.2ex}#2}}}

\def\void{}
\newcommand{\mcomment}[2][\void]{\marginpar{\raggedright\footnotesize\ifx\void#1\else\textbf{#1:\\}\fi#2}}

\newcommand{\littleparagraph}[1]{\medskip\par\noindent\emph{#1}}
\begin{document}
\thispagestyle{empty}

\title[Movement Analytics]{Movement Analytics: Current Status, Application to Manufacturing, and Future Prospects from an AI Perspective}


\author*{\fnm{Peter} \sur{Baumgartner}*}\email{Peter.Baumgartner@data61.csiro.au}
\author{\fnm{Daniel} \sur{Smith}}
\author{\fnm{Mashud} \sur{Rana}}
\author{\fnm{Reena} \sur{Kapoor}}
\author{\fnm{Elena} \sur{Tartaglia}}
\author{\fnm{Andreas} \sur{Schutt}}
\author{\fnm{Ashfaqur} \sur{Rahman}}
\author{\fnm{John} \sur{Taylor}}
\author{\fnm{Simon} \sur{Dunstall}}

\affil{\orgdiv{Data61}, \orgname{CSIRO}, \country{Australia}}

\newenvironment{summary}[1]{\subsubsection*{Summary: #1}\bgroup}{\egroup\par\noindent}

\abstract{
Data-driven decision making is becoming an integral part of manufacturing companies. Data is collected and commonly used to improve efficiency and produce high quality items for the customers. IoT-based and other forms of object tracking are an emerging tool for collecting movement data of objects/entities (e.g. human workers, moving vehicles, trolleys etc.) over space and time. \emph{Movement data} can provide valuable insights like process bottlenecks, resource utilization, effective working time etc. that can be used for decision making and improving efficiency.
 
Turning movement data into valuable information for industrial management and decision making requires analysis methods. We refer to this process as \emph{movement analytics}. The purpose of this document is to review the current state of work for movement analytics both in manufacturing and more broadly.
 
We survey relevant work from both a theoretical perspective and an application perspective. From the theoretical perspective, we put an emphasis on useful methods from two research areas: machine learning, and logic-based knowledge representation. We also review their combinations in view of movement analytics, and we discuss promising areas for future
development and application. Furthermore, we touch on constraint optimization.
 
From an application perspective, we review applications of these methods to movement analytics in a general sense and across various industries. We also describe currently available commercial off-the-shelf products for tracking in manufacturing, and we overview main concepts of digital twins and their applications. 
}






\maketitle
\clearpage\newpage

\thispagestyle{empty}
\tableofcontents
\clearpage\newpage

\setcounter{page}{1}

\section{Introduction}
\label{sec:introduction}


Manufacturing is a large and complex process, and significant resources are required to efficiently produce goods that meet quality standards. There are many challenges such as in the areas of quality control, fault detection, maintenance, planning and logistics. Companies need to simultaneously maximize sales revenues, minimize production costs and maximize customer service levels, all while providing high standards of quality~\cite{altiok_performance_2012}. Data analysis can provide valuable insight by diagnosing problems and informing decision-making to improve processes~\cite{harding_data_2005, nagorny_big_2017,altiok_performance_2012}.

IoT-based and other forms of object tracking are an emerging tool for data collection, capable of capturing vast quantities of data. 
These systems typically involve fitting ‘tags’ to people or objects and collecting
position information over time. Other systems use computer vision to record location and
activity. Some systems, including Embedded Intelligent Platforms~\cite{csiro_secure_nodate} have smart tags that exchange information between each other. This provides additional data on the state and action of objects/entities during their trajectories. The movement and trajectory data acquired by these systems over time can be turned into valuable information for industrial management and for automation and algorithm design.

To extract knowledge from this tracking data, we need analysis methods to process the data and provide meaningful solutions to decision-makers. We refer to this process as \emph{movement analytics}. While many modeling techniques and algorithms exist for analyzing tracking data, more work can be done to better apply them in a manufacturing context. There are many challenges to overcome working with tracking data, such as noise and missing data. Furthermore, the sheer volume of data can be difficult to manage. 

The purpose of this document is to review the current state of work for movement analytics both in manufacturing and more broadly. We present relevant work from both a theoretical perspective (useful methods and algorithms) and an application perspective (current implementations in manufacturing and other industries). Before going
into the details, it is helpful to set the stage and characterize the kind of issues
we had in mind while preparing this review and the methods that we think are
relevant to address them. 

\paragraph{Problem Space}

The movement analytics problems we are interested in solving contain four key ingredients. Firstly, we are interested in solving \emph{advanced} movement analytics problems. Secondly, we focus on problems that involve movement, so problems that make use of data that tracks objects or people are in scope. Thirdly, we are interested in problems where the state of an object is important. Finally, we are interested in problems that are relevant to manufacturing, though we broaden our scope to include problems from other application areas if we can see how they could arise in a manufacturing context.

In this review, we focus on complex problems that require advanced movement analytics methods. Analyzing trajectories can be straightforward, such as tracking an object over time and some simple data aggregation, e.g., deriving speed as movement over time.
However, we are more interested in problems that require
sophisticated data analysis and/or integrating additional knowledge or information
sources. See \citet{bian_trajectory_2019-1} for a related overview. Specifically, we focus on problems that combine the use of trajectory data and state information.

A core part of the problems we are interested in solving is that they involve \emph{trajectories}. Following \citet{zheng_trajectory_2015-1}, a (spatial) trajectory is a trace generated by a moving object in geographical spaces, usually represented by a series of  chronologically ordered points, e.g., $p_1 \to p_2 \to \cdots \to p_n$ where each point consists of a geospatial coordinate set and a timestamp such as $p = (x, y, t)$ (or $p = (x, y, z, t)$). It is by having such a trajectory that we restrict to problems that involve movement.

The meaning behind an object's trajectory is inextricably linked to its \emph{state}. The state of an object could refer to the type of object, e.g. a forklift or a worker, or a certain property of an object, e.g. a truck that is full or empty of cargo. On the one hand, knowing what type of object traces a certain trajectory allows for inference of the meaning behind its movement. For example, knowing that a truck has traveled from a manufacturing plant to a warehouse could help us infer that it contains finished products. On the other, the trajectory itself could be used to infer the state of an object. For example, trajectories of objects that visit the canteen or toilets could be inferred to be humans. 
A good example of a class of problems that make use of trajectories and state are those that require \emph{behavioral analysis} (see \citet{lei_framework_2016} for an overview).

Examples of potential manufacturing processes that require (anomalous) behavior analysis are \emph{searching}, \emph{shuffling}, and \emph{reworking}. Searching behavior is when a worker looks for a misplaced item, something which could be detected from their trajectory if they are circling or revisiting places within a short period of time. Shuffling and repacking behavior occur when a production facility is not well organized and more reorganization than necessary is required when new items come into the storage area. Again, this could be detected by analyzing trajectories of products which would be useful for identifying inefficiencies in storage processes. Reworking refers to when something goes back and forth between work stations because it does not pass quality control requirements. Identifying which parts or workstations are involved in reworking could help identify problems with the manufacturing process.

It is these kinds of problems we had in mind when conducting this review. Next we describe the scope of methods we investigated for solving them.

\paragraph{Methods in Scope}
Confining the scope of methods narrowly to ``(indoor) movement manufacturing'' bears the risk of missing
technology that is potentially useful, even if not directly or obviously
related. We therefore considered, more generally, methods for spatio-temporal data analysis as relevant, and included them guided by the following criteria:


\begin{description}
\item[Indoor vs.\ any location type.] While most of ``manufacturing'' happens indoor, we
  include problems and methods that carry over independent of location type, e.g., anomalous behavior patterns.
\item[Manufacturing vs non-manufacturing specific.] While our key application area is manufacturing, we include problems outside of manufacturing if they could be transferable to a manufacturing context. For instance, flow analysis of urban traffic,  anomaly detection in maritime
scenarios, robot movement planning, and visual language navigation (Section~\ref{sec:knowledge-graphs-VLN}) can provide structurally
similar phenomena and were considered in scope.
\item[Movement vs non-movement.] 
A paper or method can even be relevant if it is not directly on ``movement'' but is utilized in documented movement applications.
A good example is temporal logic,
which is movement-agnostic but can be used to specify safety, liveness or planning
constraints for robot (or any kind of) movement.  
Another example are digital twins, which we see relevant for providing simulated data sets
for movement analysis.
\end{description}

We put an emphasis on two research areas:
machine learning (ML) on the one hand, and logic-based knowledge representation on the
other hand.
There is a history of using ML methods on trajectory data for various purposes, including
behavior classification, collision avoidance and identification of anomalous
trajectories.  ML methods excel when there are huge volumes of data, but it is not always
practical to gather large data sets.  Logic-based methods, on the other hand, can capture
domain knowledge and infer otherwise unknown information, but often do not scale well and are labor intensive in design.

There are approaches that integrate ML and logic to make the most of them, in
combination. We will also review these combinations in view of movement analytics, and we
will offer some speculation about which ones seem most promising for future development and
application.
Constraint optimization is another area we touch on, as this area can
provide optimal strategies for decision making.


\paragraph{Structure}
This literature review is structured as follows. In
Section~\ref{sec:fundamental-techniques} we review fundamental techniques for
spatio-temporal data analysis from our application perspective. We cover classical logic
and knowledge representation as well as its probabilistic extension, ML,
trajectory methods, the integration of logic with ML and constraint
optimization. In Section~\ref{sec:applications-movement-data-analytics} we review analytic
techniques that have been applied to movement data. In
Section~\ref{sec:industrial-applications-and-commercial-systems} we give examples of
movement analytics across 
various industries and describe currently available commercial off-the-shelf products for
tracking in manufacturing. We also overview main
concepts of digital twins and their applications in manufacturing more generally.
Finally, in Section~\ref{sec:conclusions} we give a summary of the findings and offer some
ideas for future research. 

\begin{figure}[htpb]
  \centering
  \includegraphics[width=0.95\textwidth]{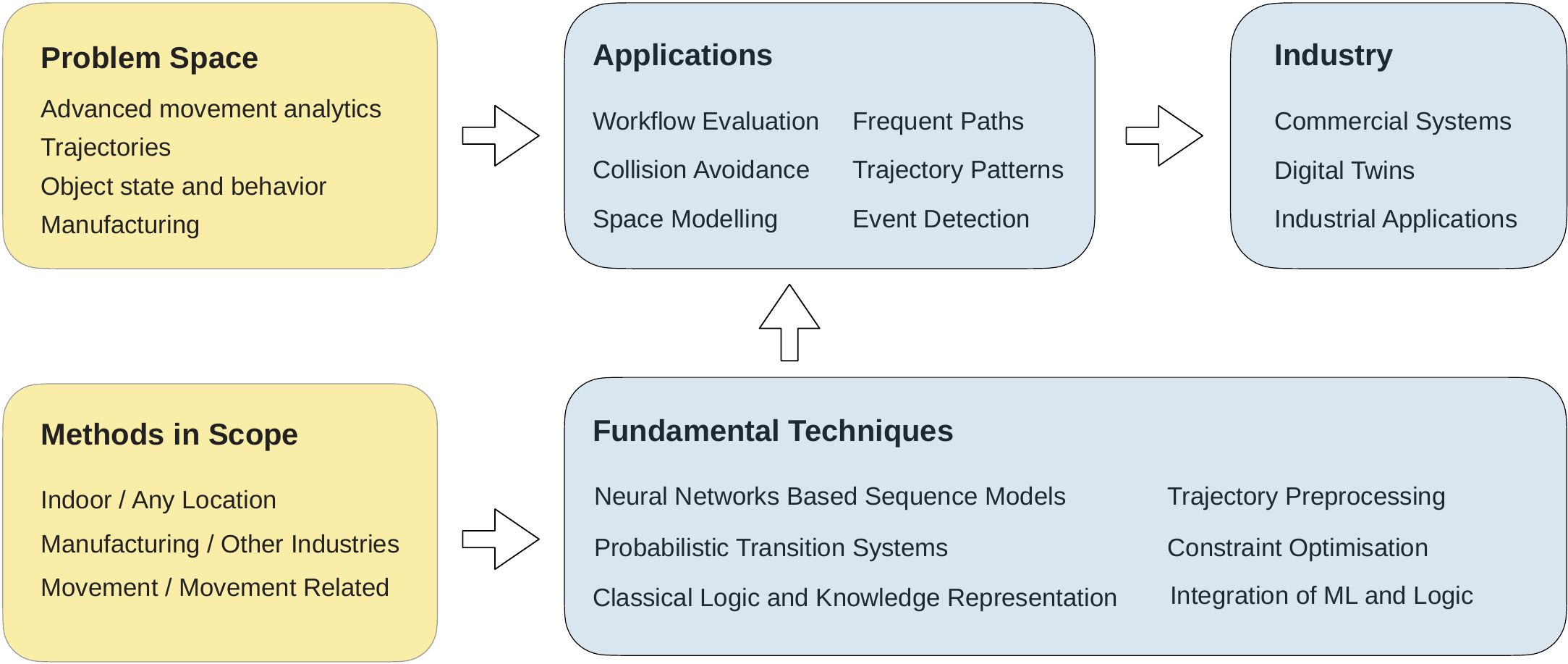}
  \caption{A structured view of the topics of this review.}
  \label{fig:topics}
\end{figure}

Figure~\ref{fig:topics} depicts a structured summary of the topics touched upon above.

\section{Fundamental Techniques} 
\label{sec:fundamental-techniques}
In this section, we review fundamental methods for analyzing movement and spatial trajectories. We distinguish between methods based on classical logic and related knowledge representation formalisms, probabilistic transition systems, and neural network based ML. We first discuss these methods separately, which reflects the (by and large) historical disjointness of their underlying research areas.
Following that, we turn to efforts of combining logic and ML. 
We also include an overview of preprocessing techniques for trajectories, which are ``fundamental'' in the sense that some sort of data cleaning or aggregation is often needed in preparation of any of these methods.
Finally, we give a brief description of the relevance of constraint optimization to movement analytics.

\subsection{Classical Logic and Knowledge Representation} 
\label{sec:classical-logic-KR}
Following common usage, \emph{classical logic} refers to a family of formal logical languages that includes propositional logic and first-order logic. 
Propositional logic supports formulas over Boolean variables, while first-order logic supports existentially and universally quantified formulas over structured objects.
Classical logic is equipped with a standard (``Tarski'') semantics that enables push-button style automation of reasoning tasks like entailment (theorem proving) and consistency checking (diagnosis). Classical logic and variations have many applications in computer science~\cite{halpern_unusual_2001}. 


However, propositional logic is often not expressive enough for knowledge representation
in real-world applications, and, in general,  first-order logic is not amenable to
efficient automated reasoning (important reasoning services like ``theorem proving'' are not
even decidable). This is why research in automated reasoning has developed
specialized reasoning procedures for specific applications or classes of
applications. A prominent and successful example is
\emph{Satisfiability Modulo Theories (SMT)}~\cite{barrett_satisfiability_2018}, which
generalizes propositional satisfiability solving (SAT solving) to more complex formulas involving
real numbers, integers, and/or various data structures such as lists, arrays, bit vectors,
and strings.  SMT solvers typically only support first-order logic  quantifiers in a
restricted way, for decidable and efficient reasoning. Main applications are in software and hardware verification, but not so much in areas requiring spatio-temporal reasoning.

In the following, we focus on specialized approaches that are relevant for movement
analytics.

\subsubsection{Space and Time}
\label{sec:space-and-time}
\citet{allen_maintaining_1983} introduced an influential calculus for qualitative reasoning on time intervals (e.g., ``interval $A$ overlaps with interval $B$''). It  provides a composition table that can be used, e.g., for consistency checking and for deriving implied temporal relations. Allen's calculus is, in fact, a certain relational theory that can be expressed in first-order logic. 

The family of \emph{temporal logics}~\cite{goranko_temporal_1999} provides operators for stating temporal relations between worlds characterized \emph{as formulas} (formula ``$A$ must hold \emph{before} formula $B$'', ``\emph{Every} request must \emph{eventually} be acknowledged''). 
Temporal logics play a major role in software/hardware verification and runtime
verification~\cite{bauer_runtime_2011}. For the latter,  temporal logic formulas typically specify safety
(non-anomality) conditions to be tracked during runtime, e.g., that certain milestones
must be reached or resource constraints must be met within certain time limits~\cite{li_satisfiability_2019}. Specific applications for robot navigation planning and safety under temporal logic constraints have been proposed, e.g.,  by~\citet{yoo_provably-correct_2013},~\citet{akin_asymptotically_2015} and by \citet{li_reinforcement_2017}.

The region-connect-calculus (RCC) enables qualitative spatial reasoning with symbolic relations between regions (``connects with'', ``overlaps with'', etc). Like Allen's temporal calculus, the RCC relations can be  axiomatized in first-order logic. See \citet{cohn_qualitative_2008} for an in-depth overview. 
\citet{galton_spatial_2009} provides a more general overview of qualitative combined spatial and temporal reasoning. 

\citet{ge_towards_2018} describe a qualitative theory of object movements with a  qualitative spatio-temporal approach.  Their main application is explaining causes for observed changes of qualitative relations between composite objects changing shape over time (e.g., towers of blocks forming an arc which then collapses).
\citet{li_spatio-temporal_2020} develop a spatio-temporal logic that combines temporal
logic with a spatial region calculus and prove decidability properties. The logic is mainly motivated by verification of
cyber-physical systems (e.g., performance guarantees of train emergence braking
system). The paper also contains an overview of related work in that area.

\subsubsection{Description Logics}

Description logics~\cite{baader_description_2008} (DLs) are a family of knowledge representation languages with first-order logic semantics. At their core, DLs support specification of \emph{ontologies} in terms of \emph{is-a} and \emph{has-a} relations. Much of the research on description logics is fundamental in nature but is strongly motivated by practical applications. In contrast to first-order logic, DLs variants are (mostly) designed so that important
reasoning services like consistency and query answering are decidable, i.e., they are
guaranteed to terminate with a (correct) result on any request.

This lesser generality compared to first-order logic gives rise to numerous DL variants that are tailored for specific purposes.
For instance, extensions exist for RCC-style spatial reasoning~\cite{lutz_tableau_2007} and temporal logic~\cite{lutz_temporal_2008}.
DLs and their implementations serve as rigorous theoretical and practical tools for semantic web languages such as OWL~\cite{bechhofer_owl_2009}.   

The DL area of ontology-based data access (OBDA)~\cite{faber_ontology-mediated_2015} is concerned with implementing systems that collect data at runtime for recognizing certain predefined
situations and triggering adaptations. Technically, the OBDA approach consists of augmenting classical query answering in databases by adopting the open-world assumption and including domain knowledge 
provided by a DL ontology. 
OBDA can be temporalized.
For instance, \citet{borgwardt_temporal_2015} propose a query language extended by temporal logic. \citet{lutz_stream-temporal_2014} designed an SQL-like OBDA query language for streamed data (time windows).

Interestingly, with the advent of the Yago ontology~\cite{harth_yago_2020} a connection to big data mined from the Internet (Wikipedia) is given. See also Section \ref{sec:knowledge-graphs-VLN} Knowledge Graphs.

More at the  meta-level, \citet{palmer_ontology_2018} propose an ontological framework for risk assessment in supply chains. With its focus on strategic issues, such approaches may well be transferable to indoor movement scenarios.

\subsubsection{Logic Programming and Event Calculus} 
\label{sec:logic-programming-event-calculus}
Logic programming is a programming paradigm that separates the knowledge to be used for
solving a problem from the control component, which determines the strategy for how this
knowledge is to be used~\cite{kowalski_algorithm_1979}. While logic programming is largely
based on first-order logic, it differs by assuming a closed-world semantics (roughly:
``what is not known to be true, is false''). Like many other
non-monotonic formalisms, this makes logic programming often more suitable for real-world
knowledge representation, 
which requires drawing conclusions even from incomplete knowledge.  At the same time, 
important reasoning tasks become more intractable than under standard first-first order
logic semantics. In general, satisfiability testing and theorem-hood are not even
semi-decidable anymore. 

In other words, strong expressivity and good computability properties are competing
goals. This dichotomy is traditionally
reflected in the two main paradigms of logic programming, ``Prolog-style'' and ``Answer
Set Programming''. The former emphasizes expressive power (Turing-completeness) and
leaves much responsibility to the programmer, like in traditional programming. The latter
enables highly declarative specifications of, typically, search problems over finite
domains that largely relieve the programmer from specifying control. Technically, Prolog-style logic
programming is about answering queries by backchaining if-then rules towards facts;
Answer Set Programming does not need a query and is more about computing logical models of
sets of if-then rules.
\citet{apt_logic_1994} and \citet{baral_logic_1994} provide overviews.

Logic programming is application agnostic and can be utilized and specialized in many ways. For instance, \citet{benzmuller_answer_2018} and 
\citet{calimeri_aspmtqs_2015} equip logic programming with spatio-temporal
\emph{quantitative} reasoning along the RCC calculus and report on experiments.
The \emph{event calculus}~\cite{kowalski_logic-based_1986} (EC) is a widely studied
formalism that utilizes logic programming
as a host formalism for reasoning about time and events. An EC model specifies the
consequences of an event $e$ happening at a time $t$ in terms of \emph{fluents}. Fluents
are properties of objects that can change over time. When an event $e$ causes a fluent $f$ to become true (false) at time $t$ then $f$ remains true (false) until changed later.
\citet{hutchison_reactive_2012} developed a ``reactive'' version of the EC for dynamically extending a narrative represented so far. 
\citet{konev_combining_2021} has similar motivation but also incorporates description
logic reasoning for added expressivity. The approach has been applied, among others, for anomaly
detection in a food supply chain~\cite{baumgartner_anomaly_2021} and for maritime
surveillance~\cite{pitsikalis_event_2020}. 
\citet{skarlatidis_probabilistic_2015} define a probabilistic version on top of the
probabilistic logic programming language ProbLog (see
Section~\ref{sec:statistical-relational-learning} for ProbLog). They apply it to problems of recognising long-term activities from short-term activities in trajectory data. This can be seen as behavior recognition. 

Logic programming has a long-standing connection to relational database
technology~\cite{ceri_logic_1990}. 
The most prominent framework is \emph{Datalog}, which is both a declarative logic programming language and a lightweight
deductive database system for expressing queries and database updates.
An overview is available~\cite{greco_datalog_2015}.

Logic programming and Datalog have received revived interest  
for combining logic and ML, see Section~\ref{sec:logic-programming-and-machine-learning}.

\subsubsection{Logic-Based Stream Processing}
\label{sec:logic-based-stream-processing}
Another line of research subsumes modeling systems that evolve over time under the
terms of \emph{stream processing}  or \emph{complex event recognition}. These approaches
aim at devising systems for recognizing high-level events from a stream of low-level events coming from, e.g., sensor networks such as radio frequency identification (RFID) tags~\cite{valle_its_2009}. 
\citet{tsilionis_incremental_2019}, for example, describe an approach for sea vessels movement data analysis.
The implementation is by way of event calculus and scales up to realistically sized data.

Stream processing can be be formulated in a logic programming framework (Section~\ref{sec:logic-programming-event-calculus}). A sophisticated system is LARS~\cite{beck_lars_2018}, which extends logic programming by temporal operators for
time windows for data monitoring.
\citet{artikis_logic-based_2012} provides an overview over \emph{logic-based} approaches
in general, \citet{alevizos_probabilistic_2017} provides an overview over
\emph{probabilistic} methods, including logic and event calculus, and
 \citet{giatrakos_complex_2020} overviews the area even more broadly from a big data perspective.



\subsubsection{Symbolic/Semantic Trajectories and Databases}
\label{sec:symbolic-trajectories-non-ML}
\emph{Symbolic trajectories} \cite{guting_symbolic_2015} are trajectories enhanced with time-dependent labels, e.g., for transportation modes like ``walk'', ``car'' or ``bus'' along a person's commute to work~\cite{xu_moving_2019}. Their storage and querying is often backed by database systems~\cite{valdes_framework_2019}.
 
\citet{guting_modeling_2006} propose reusable abstract data types for moving objects and their environment. The approach provides a semantic abstraction of $(x,y,z,t)$ in terms of points, lines, regions and an SQL-like query language over symbolically named objects, geometric properties, and time.
 The main application considered by \citet{guting_modeling_2006} is road
 networks. Abstract data types can describe interchanges and their constraints on usage of lanes. The paper uses logic for defining geometric properties of objects, e.g., ``inside''.

\citet{li_toward_2020} present a method for aggregating movement sensor data into triples carrying ``what/when/where'' information. Their main application is analyzing customer behavior in a shopping center based on WiFi sensor input of the form $(x,y,\textnormal{floor},t)$. 
Semantic geographical regions, e.g. ``Nike shop", are predefined. Their method has a three-layered architecture: cleaning, annotation, and complementing. Cleaning uses constraints for plausibility checks with respect to movement speed, e.g., walking. Annotation turns raw data into semantic annotation sequences of triples (movement type, shop type, time). Two methods have been tried: clustering (ST-DBSCAN) and semi-supervised learning using logistic regression and domain knowledge. These sequences are often incomplete due to incomplete WiFi sensing. 
The complementing layer infers the missing points based on the (indoor) topology and movement patterns studied in the area of human mobility prediction. This domain knowledge is hard-coded into algorithm. Technically, ``complementing'' is a max-posteriori classification problem under a Markov-assumption. 

In many cases, the exact location of objects is uncertain. Such cases can be handled by \emph{probabilistic knowledge bases}~\cite{tao_indexing_2005}. More recent work by \citet{parisi_integrity_2014,parisi_knowledge_2016} supports representing atomic statements of the form ``object $\mathit{id}$\ is/was/will be inside region $r$ at time $t$ with probability in the interval $[l, u]$'' and integrity constraints on such statements.
\citet{tawfik_temporal_2000} discuss the treatment of time, causality, and the representation of events, effects and interactions more broadly. 



\begin{summary}{Classical Logic and Knowledge Representation}
There is a plethora of logics rooted in first-order logic that support fundamental needs for movement analysis such as 
representation of internal structure of objects, time and space.
Research in these areas often has fundamental character and aims to deliver theoretical results
such as correctness and complexity properties of important reasoning services.
The literature is rich with such results.

On the practical side, application-oriented subfields have produced implemented systems for automated reasoning and knowledge representation but some caveats apply.
Some of the main issues revolve around applicability and extendibility. 
For example, conjunctive query answering over description logics extended with temporal operators is well understood and supported by efficient implementations.
If, however, a seemingly ``minor'' extension is needed, say, for a quantitative treatment of time, tool support might no longer be available.
It could lead to undesirable complexity or decidability properties, or the particular extension has not yet been studied, or an implementation does not support the needed feature.

Nevertheless, automated reasoning tools have been successfully embedded into larger
systems or methodologies for solving real-world problems in industry. They seem
particularly useful for problems that are either highly complex in terms of the number of
alternatives to be explored (if-then reasoning) or in terms of data size, but not in
combination. Noteworthy areas are software verification, e.g., application of SMT solvers at Microsoft for
Windows driver verification~\cite{ball_slam_2004}, and for smart contract verification in the Azure blockchain. See~\cite{tolmach_survey_2022} for a survey on that topic. NASA employed model checkers and other logic-based methods as part of its verification and diagnosis approach for spacecraft control software~\cite{nelson_formal_2003,markosian_program_2007}. 
More recently, classical logic automated theorem provers play a significant role in Amazon Web Services to increase the security assurance of its cloud infrastructure~\cite{cook_formal_2018}. \citet{backes_reachability_2019} report on successful application of this technology to network reachability analysis of up to 10000 nodes. Another example with large data is SNOMED CT, the Systematized Nomenclature of Medicine - Clinical Terms, a comprehensive, multilingual terminology for the electronic health record. Its 311,000 concepts were formalized in a limited expressive Description Logic and analyzed automatically for design flaws~\cite{schulz_snomed_2009}. 

Systems like these have matured into industrial strength quality and there is no reason they could not be applied to related problems in movement analytics.

Logic programming has been motivated as a more versatile approach for general
algorithm development. For movement analytics, we expect in particular
probabilistic versions of interest and when combined with other relevant approaches.
Among these are \emph{Dynamic Bayes Networks} (Section~\ref{sec:dynamic-bayes-networks}),
which employ fluents in the same sense as the EC, however not via a logic programming
setting. \citet{mantenoglou_online_2020} describe a probabilistic version of the event calculus.
In Section~\ref{sec:statistical-relational-learning} we cover probabilistic logic
programming as part of a larger sub-field of AI.

Returning to classical logic, spatial and temporal logics can also be of value in supporting roles.
For example, temporal logic has been used as a sub-system for
safety constraints in robot movement planning and monitoring.
Similarly, plausibility constraints on, e.g., movement speeds of
pedestrians could be stated in a suitable logic or with logic
programming. (We found approaches with such
constraints hard-coded.)

The literature that we reviewed suggests that logic-based spatio-temporal methods are theoretically well understood  but rarely used as stand-alone methods. Moreover ``time'' is far more prominent than ``space''. Logical languages with a built-in notion of time do play a role for querying and integrity constraints on (clean) data sets.
Logic-based approaches appear not relevant for directly handling imperfect sensor data. For that, hierarchical combinations make more sense, where, e.g., statistical or ML methods deal with cleaning and aggregating sensor data for downstream analysis by logical methods.
\end{summary}




\subsection{Probabilistic Transition Systems}
\label{sec:probabilistic-transition-systems}
Probabilistic transition systems generalize finite or infinite state transition systems (automata) by
probabilistic transition relations.  They can be used for modeling systems and phenomena
that appear to develop over time in a stochastic manner. 
For movement analysis, states could represent, for instance, points in space and state transition
sequences could represent trajectories under uncertainty of object locations as they
move.

\subsubsection{Markov Chains and Markov Decision Processes}
\label{sec:MC-and-MDP}
The most basic kind of probabilistic transition systems are \emph{Markov chains}. In
a Markov Chain, the transitions outgoing from any state are weighted by probabilities and form, in sum, a distribution over its successor states. Typically, the distribution models the response of an environment to some event.  
Markov Decision Processes (MDPs)~\cite{feinberg_handbook_2002} generalize Markov chains by
adding an extra ``action'' layer in between transitions. Typically, actions are under user
control (e.g., robot control input) and come with costs (e.g., time, fuel). An MDP needs to be equipped with a
\emph{policy}, which specifies what action to
take in what state. Policies can be deterministic or probabilistic,  and can take action history
into account (or not).
One of the main reasoning tasks for MDPs is computing a policy so that given objectives
are satisfied.
Objectives are typically stated in terms of value maximization (typically in expectation),
where the value of policy is aggregated from individual rewards for each transition or state . Also,
temporal logic constraints are possible~\cite{ding_mdp_2011}. 

The  robotic applications mentioned in Section~\ref{sec:space-and-time} by \citet{yoo_provably-correct_2013},
\citet{akin_asymptotically_2015} and by \citet{li_reinforcement_2017} are formulated in an
MDP framework.  


Both Markov chains and MDPs can be complicated by hidden states that only allow for
stochastic observation of the current state. An important case are Kalman filters, with
canonical application to predict the next state of a moving object given noisy
observations of the current state. See Section~\ref{SSM} below for their application to movement analytics.

\subsubsection{Reinforcement Learning}
\label{sec:reinforcement-learning}
Reinforcement learning (RL) is the task of learning behavior by trial and error by positive or negative feedback from MDPs. The RL
task, then, is to synthesize an optimal policy, as just stated in
Section~\ref{sec:MC-and-MDP}, however under a priori unknown rewards.
RL is different to supervised and to unsupervised learning in that it does not
need a pre-defined labelled or unlabelled set of data.  It instead relies on
exploiting previous experience and exploring untried alternatives. RL, in general, needs
to sample a space of probabilistic state transitions that grows exponentially with the
state space. To address this problem, policy approximations may be required, for instance by using neural
networks (``deep reinforcement learning'').

An overview of RL that also takes psychological aspects into account is
available~\cite{collins_beyond_2020}. For a comprehensive introductory book see
\citet{sutton_reinforcement_2018}.

As for industrial applications, RL has
been deployed in production systems for process optimization and reducing reliance
on human experience. \citet{panzer_deep_2021} conducted a literature review on this topic.
(Deep) RL has been applied to robot motion problems, see, e.g.,
\cite{li_reinforcement_2017,lee_reinforcement_2022}. (Probabilistic) temporal logic can be brought into
the picture quite naturally, for specifying (un)desirable properties of robot movement
plans~\cite{camacho_ltl_2019}. \citet{liao_survey_2020} wrote a survey on reinforcement
learning with temporal logic constraints.

For numerous other applications see the overviews by
\citet{arulkumaran_deep_2017} and by \citet{li_deep_overview_2018}. 

\subsubsection{Dynamic Bayes Networks} \label{DBN}
\label{sec:dynamic-bayes-networks}
A \emph{Bayes network}~\cite{pearl_probabilistic_1988}  is a directed acyclic graph whose nodes represent domain variables
of interest and whose edges represent conditional (informational or causal) dependencies between a node and its
parents. A Bayes network supports the computation of the
probabilities of any subset of variables given evidence about any other subset.
In practice, Bayes networks are often used for taking an event that occurred and
predicting the likelihood that any one of several 
possible known causes was the contributing factor. 
However,  Bayesian networks do not model temporal relationships between variables. 

\emph{Dynamic} Bayes networks (DBN) ~\cite{dean_model_1989,carbonell_learning_1998} extend
Bayes networks with a temporal dimension.  A DBN is a time-indexed sequences of Bayes
networks where the network at each state depends (usually) only from the previous
one. Like in the untimed case, the dependencies must form a directed acyclic graph, but
additional edges from variables from immediately preceding timepoints are permitted now.
One of the main inference tasks is state monitoring, the task to estimate the current
state of the world given the observations (evidence) made up to the present.
\citet{carbonell_learning_1998} provide an overview over inference and learning
procedures of DBNs.


Regarding applications, \citet{roos_dynamic_2017} propose a DBN approach
for forecasting short term passengers flows within an urban rail
network given their capacity to model the dynamic system under
the condition of incomplete passenger flow data. Passenger flows were predicted from
the DBN based upon observations in their local spatio-temporal neighborhood.

DBNs are rather expressive. They cover commonly used models like Hidden Markov models,
Kalman filters, time series clustering, auto regressive model and extensions thereof. See
the following Section~\ref{SSM} for more details and application of those.



\subsubsection{State Space Models}\label{SSM}
State Space Models (SSMs) are probabilistic graph models used to represent  
a dynamic system by a set of
differential equations and latent states that are associated with the observed data.
The objective of state space modeling is to estimate the optimal latent states using the observed data and knowledge encapsulated within the system equations. Given that SSMs are stochastic, inference becomes computationally expensive, unless certain assumptions are made. Classic time series models, such as Hidden Markov Models (HMM)~\cite{baum_le_statistical_1966} and Auto-Regressive Integrated moving Average (ARIMA)~\cite{box_george_time_2008}, can be formulated in state space, making the assumption that the dynamic system is linear, Gaussian distributed and only dependent upon state at the previous time step (Markov assumption). This makes latent state inference tractable using approaches such as Kalman filtering~\cite{kalman_re_new_1960} or the Viterbi algorithm~\cite{viterbi_andrew_error_1967}, however, this comes at the expense of having a model with less expressivity. For instance, the Markov assumption limits the ability of classic time series models to represent trajectories, given they can only capture the relationship between consecutive samples. 

If a more expressive, non-linear model is to be formulated, an exact solution for the latent states cannot be estimated. In this case, sequential importance sampling, commonly referred to as particle filtering~\cite{del_moral_pierre_non_1996}, is commonly used to approximate the latent states.


In terms of their application, SSM have commonly been used as a pre-processing step to reduce the noise in spatial trajectories. Linear dynamic models, based on the Kalman filter~\cite{parent_semantic_2013} and Gaussian kernel based regression models~\cite{hutchison_hybrid_2010} have been used to smooth out trajectory noise. Furthermore,~\citet{gustafsson_particle_2002} used a particle filter in conjunction with spatial context (in the form of 
maps) to reduce the influence of measurement noise. Position estimates from the particle filter were constrained by the spatial context provided by maps. For example, estimates of car position were constrained to locations on the given road network and aircraft position estimates were constrained by altitude maps.

HMMs have been used in the map matching process~\cite{quddus_current_2007} (see Section~\ref{sec:trajectories-preprocessing} for a description of map matching), which involves transforming trajectories of raw position data into a semantic trajectory of known spatial landmarks.~\citet{mohamed_accurate_2017} employed a HMM to infer the discrete spatial landmarks from the raw spatial coordinates of trajectories subject to significant noise.  


\begin{summary}{Probabilistic Transition Systems}
Probabilistic state transition systems are suitable for analyzing stochastic time series
  data, through generative 
  modeling. In generative modeling, each event
$e_i$ updates the state of the system from 
$s_i$ to $s_{i+1}$, which then determines the distribution for drawing the next event
$e_{i+1}$.
(See Section~\ref{Generative} below for generative modeling in a neural network context.)
  Obvious applications to trajectory
analysis are on an operational level, e.g., to predict the current or near-future state, or to recommend
a reward-maximizing action at  a current state, e.g., for 
collision avoidance.

When there is no causal relationship between certain events and the system
state then accurate modeling becomes difficult. This problem could be countered by
assessing and filtering out events in context of states with the help of a domain model.
Some of the combination methods in Section~\ref{sec:ML-non-ML-integration} could be useful
for that. 

Reinforcement learning (Section~\ref{sec:reinforcement-learning}) does not
seem to play a major role for \emph{analytic} tasks like anomaly detection within given
trajectories. Well-explored applications of RL for controlling robot movements may suggest
their transfer to movement analytics. This could be done depending on the degree of
integration into an overall process:
\begin{itemize}
\item Low-level analytics, e.g., robot learns how to pick up an object.
  \item Mid-level, e.g., robots/systems work together to yield assembly of a complex
    object.
    \item High-level, e.g., factory operations including all movements are optimized.
\end{itemize}
Ultimately, a concept of adaptive reinforcement learning might turn out to be useful,
where the best solution 
changes with time. 

\end{summary}

\subsection{Trajectory Preprocessing Techniques}
\label{sec:trajectories-preprocessing}
Trajectories of raw position data often need to be pre-processed before additional modeling can be performed. There are a number of different issues to be addressed with data sets of spatial trajectories including noise, non-uniform sampling rates, uneven trajectory lengths and trajectories of an unmanageable length. To address these issues, there are four categories of
pre-processing techniques that are commonly applied to 
trajectories that will be outlined: noise reduction, segmentation, semantic mapping and harmonization.

\subsubsection{Noise Reduction} 
Trajectories are commonly represented as a sequence of inaccurate position measurements that are 
the result of the noise in the underlying sensor technology. These errors are sometimes tolerable depending upon the application at hand, whilst in other scenarios, it is important to reduce the influence of measurement noise prior to modeling. 

Filters are the simplest noise reduction approaches for spatial trajectories. Mean, median or moving average filters transform spatial trajectories by computing their respective statistic across sliding windows. These filters were used to reduce the influence of noise by smoothing the spatial trajectories across a local neighborhood of samples~\cite{lee_trajectory_2011}. State space models (which we define in Section \ref{SSM}) representing the linear dynamics of motion ~\cite{parent_semantic_2013} have been used for noise reduction. Furthermore, more complex models of noise reduction, representing the non-linear motion dynamics have been developed with a particle filter ~\cite{gustafsson_particle_2002}, Gaussian kernel regression model~\cite{hutchison_hybrid_2010} and Savitzky-Golay filter\cite{savitzky_smoothing_1964}. 

Anomaly detection methods have also been used for noise reduction. In~\citet{jing_yuan_t-drive_2013}, a simple heuristic was proposed to detect anomalies as the samples where trajectory velocity were deemed to be implausible for the moving object. In~\citet{hu_anomaly_2020}, a kernel function was used to model the probability density function (pdf) of trajectory samples with respect to a window of neighboring samples. Anomalous samples were selected as low density points in the trajectory's probability density function that were then removed.

\subsubsection{Segmentation} Spatial trajectories are often partitioned into a set of sub-trajectories prior to being used by ML or data mining methods. This process is known as segmentation where each segment represents the maximal subsequence of a trajectory with samples that comply with a given predicate~\cite{parent_semantic_2013}. Commonly, this predicate might be associated with a particular behavior, activity or geographical location that the moving object is associated with. Trajectory segmentation is commonly used to improve the performance and computational efficiency of downstream modeling tasks~\cite{parent_semantic_2013}.

Segmentation approaches either exploit the statistical properties of trajectories or semantic knowledge of the objects being tracked. Statistical segmentation approaches include methods that utilize changes in the trajectory shape ~\cite{zheng_trajectory_2015-1} or an information theory based criteria~\cite{lee_trajectory_2007}. For instance, ~\citet{lee_trajectory_2007} used an entropy based criteria, the Minimum Description Length (MDL), to segment trajectories such that they were maximally compressed. 

Semantic approaches to trajectory segmentation often involve stop point detection, identifying the points at which moving objects appear to become stationary~\cite{atluri_spatio-temporal_2018,zheng_trajectory_2015-1}. These stop points can be used to partition trajectories into segments without object movement (the stop segments), whilst intervals between these stop points became the movement segments. These segmentation methods are often essential for route optimization applications, for instance, taxi trip optimization.~\citet{jing_yuan_t-drive_2013} partitioned vehicle trajectories into individual trips prior to them being clustered and applied to route optimization algorithms. 
Furthermore, stop points and social media information can be combined to segment tourist
trajectories ~\cite{alvares_model_2007}. Stop points were detected at points with minimal
tourist motion and combined with coinciding Point of Interest 
check-ins to generate  a semantic trajectory of tourist attraction visits.

\subsubsection{Semantic Trajectories} 
\label{sec:semantic_traj}
Location information can often be enriched with contextual information enabling the raw position data of trajectories to be translated into a semantic trajectory of 
annotations meaningful to the application at hand. This semantic translation can be useful to eliminate the noise that is present within raw measurements and to compress trajectories to improve the computational efficiency of modeling. Furthermore, it can provide an enriched representation with a stronger and more meaningful connection to the application at hand. 

Map matching is an example of a semantic sequence transformation where the spatial coordinates of a moving object are projected onto a map network to infer the sequence of spatial landmarks that have been traversed. Map matching methods can be categorized as either geometrical or topological based~\cite{quddus_current_2007}. Geometrical methods find routes by utilizing distance measures to match trajectory coordinates to landmarks represented in the spatial network. These measures can either be point-to-point, point-to-curve or curve-to-curve. Dynamic programming is commonly used to find the optimal semantic trajectory as the set of spatial landmarks with the shortest distance to the raw trajectories. More recently, modern deep learning architectures have been designed for geometry based map matching~\cite{zhao_deepmm_2019}. These approaches have issues with inferring the optimal path of spatial landmarks (i.e. roads) that do not have spatial continuity. In contrast, topology based methods~\cite{blazquez_simple_2005} have been developed to exploit the physical topology of a network to ensure that spatial continuity and connectivity is maintained within the semantic trajectories. 

An issue with using semantic trajectories with ML, data mining or probabilistic methods is their sparse representation, which is not appropriate for such methods. Consequently, specialized neural networks (embedding networks) have been proposed to transform the sparse representation of semantic trajectories into a dense, continuous valued representation~\cite{a_de_freitas_using_2021}.
See Section~\ref{sec:symbolic-trajectories-non-ML} for other approaches to producing semantic trajectories.

\subsubsection{Harmonization} 
Data sets often contain spatial trajectories of differing length and/or non-uniform sample resolutions, which are a problem for temporal based ML and statistical models. Firstly, interpolation methods can be used to harmonize trajectories by estimating the signal across a consistent set of time points. \citet{li_deep_2018} and \citet{yao_trajectory_2017} have harmonized data sets of noisy trajectories by training neural network architectures to transform raw inconsistent trajectories into a set of consistent, compact latent representations. Harmonized latent representations of the trajectories can then be used by models. 
Secondly, conventional distance measures (i.e. Euclidean distance, cosine distance) compute the match between equivalently indexed samples of trajectories, and hence, are inappropriate to use when comparing trajectories of a varying length. Alternative distance measures, such as dynamic time warping, allow a non-linear warping of the sample indices being matched. Such distance measures can be used with spatial trajectories of different length, however, they are more computationally expensive than conventional measures.

\subsection{Neural Network Based Sequence Models}
\label{sec:machine-learning}
A majority of common ML models, including Decision Trees (DT), Support Vector Machines (SVM) or Feed Forward neural networks have traditionally been used to represent data sets of independent and identically distributed (iid) samples. The trajectories of movement applications, however, possess spatio-temporal dependencies that are not represented with these models. Early movement applications used such models to represent the trajectories. For instance,~\citet{de_vries_machine_2012} applied an SVM to the spatial trajectories of marine vessels to identify trajectory anomalies and infer the ship class (i.e. cargo, tanker, fishing),~\citet{zheng_understanding_2010} used a DT classifier to infer the transportation modes of individuals using handcrafted features of their movement trajectories, and~\citet{niu_deepsense_2014} predicted traffic flow within cities using a generative neural network (Restricted Boltzmann Machine) and support vector regression model. 

There is a class of neural network architectures that have been developed to represent
sequence data with sample dependencies. As will be outlined in
Section~\ref{sec:applications-movement-data-analytics}, manufacturing based movement
applications in the literature have not commonly utilized this class of neural sequence
models. Consequently, we introduce the most common and contemporary architectures for spatio-temporal data and refer to where they have been used for movement applications in alternate domains to manufacturing.

\begin{figure}
\centering
 \includegraphics[width=.8\linewidth]{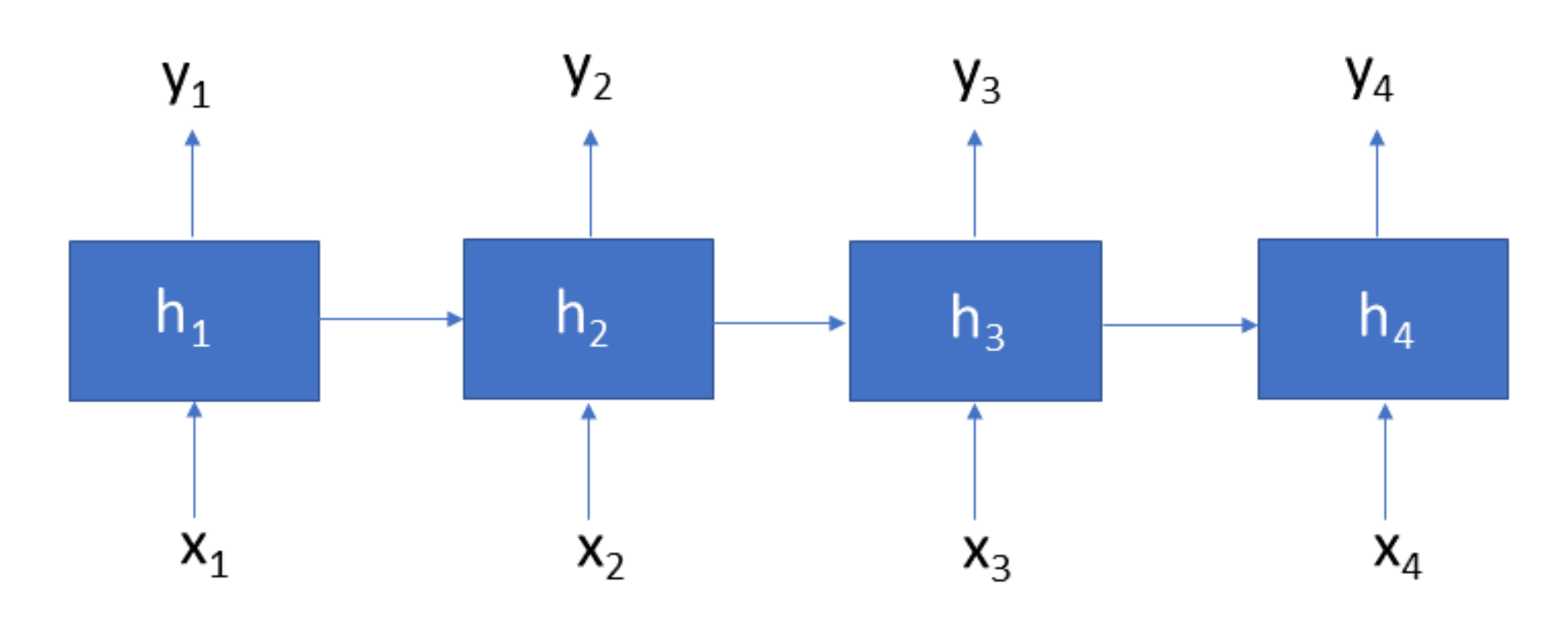}
 \caption{A recurrent neural network architecture that was used to model a sequence $x$ as an output sequence $y$. The latent state $h_{n}$ (representing the past sequence, $[x_{1},...,x_{n-1}]$) is updated for each sample $x_{n}$ using the previous latent state $h_{n-1}$.}
\label{fig:RNN}
\end{figure}

\subsubsection{Recurrent Neural Networks}\label{RNN} 

Recurrent Neural Networks (RNN) are a class of network architectures~\cite{rumelhart_learning_1986} that have been developed to represent sequential data. Unlike feed forward architectures where data is passed in a forward direction only, RNNs utilize a recurrent feedback structure to memorize the previous samples in the sequence. Figure \ref{fig:RNN} shows an RNN architecture where the latent state ($h_{t}$), representing previous samples in the sequence, is computed by updating the previous latent state ($h_{t-1}$) with the current input sample ($x_{n}$). 

One major issue of RNN architectures is that they can be difficult to train. As training
errors are backpropagated through a sequence, the error gradients continue to shrink
between consecutive steps until their values become neglible. Such steps have a minimal
influence on modeling. Consequently, the Long Short Term Memory
(LSTM)~\cite{sepp_hochreiter_long_1997} and Gated Recurrent Units%
~\cite{chung_empirical_2014} have been proposed to extend the memory capacity of RNN architectures. Both architectures utilize specialized memory and gating cells to ensure recurrent training is less susceptible to the vanishing gradient problem, and hence, a longer horizon of samples can be modeled.

LSTMs have been the neural sequence model most commonly used with spatial trajectories for classification~\cite{song_deeptransport_2016,gao_identifying_2017,yu_tulsn_2020,zhou_fan_self-supervised_2021} and prediction tasks~\cite{song_deeptransport_2016,zhou_fan_self-supervised_2021}. Trajectory to User Linking (TUL) is a classification problem where the aim is to classify the user responsible for generating a trajectory of social media check-ins. The spatial trajectories are semantic trajectories given the samples correspond to discrete spatial locations with additional meaning (i.e. a tourist destination). The TUL problem was addressed in~\citet{gao_identifying_2017} by segmenting the semantic trajectories and then mapping the segments into a continuous valued embedding space. During inference, each trajectory embedding was then fed to an LSTM based classifier to identify the user responsible for it.
~\citet{yu_tulsn_2020} proposed a new embedding architecture that provides a scalable and data efficient solution to the TUL problem. A Siamese architecture, composed of a pair of LSTM encoders with shared weights, was used to learn a latent embedding space where trajectories of the same user were more compactly distributed in latent space than different users. TUL was then performed by applying a $k$-nearest neighbor classifier to the embeddings of the semantic trajectories.~\citet{a_de_freitas_using_2021} proposed a trajectory classification model where the spatial and temporal embeddings of segmented sub-trajectories were used in conjunction with an LSTM model.

To enhance downstream classification and prediction tasks,~\citet{zhou_fan_self-supervised_2021} used self-supervised representation learning (SSRL) to harmonize trajectories of noisy and non-uniform length. A Siamese architecture consisting of a pair of LSTM encoders was used to map non-uniform trajectories to a compact, fixed size latent representation. It was shown using SSRL in the TUL problem achieved state-of-the-art classification performance after a relatively small quantity of labeled trajectory data was used to fine tune the self-supervised network. 

~\citet{alahi_social_2016} proposed social LSTM, the first model for human trajectory prediction that incorporated the spatial interactions of individuals within a crowd. The spatial trajectories of each individual were modeled by a separate LSTM and a shared pooling layer was used to connect the latent states of each LSTM. ~\citet{song_deeptransport_2016} proposed the DeepTransport architecture to solve multiple tasks (multi-task learning) upon human mobility trajectories. The movement and transportation mode of each individual were simultaneously predicted using a hierarchical network of LSTMs to represent human mobility across different temporal scales. Two LSTM based encoders were utilized to represent the inputs of each task separately, two LSTMs were used to create a shared feature representation and a pair of LSTM decoders were used to generate the outputs of each task.


\begin{figure}
\centering
 \includegraphics[width=.75\linewidth]{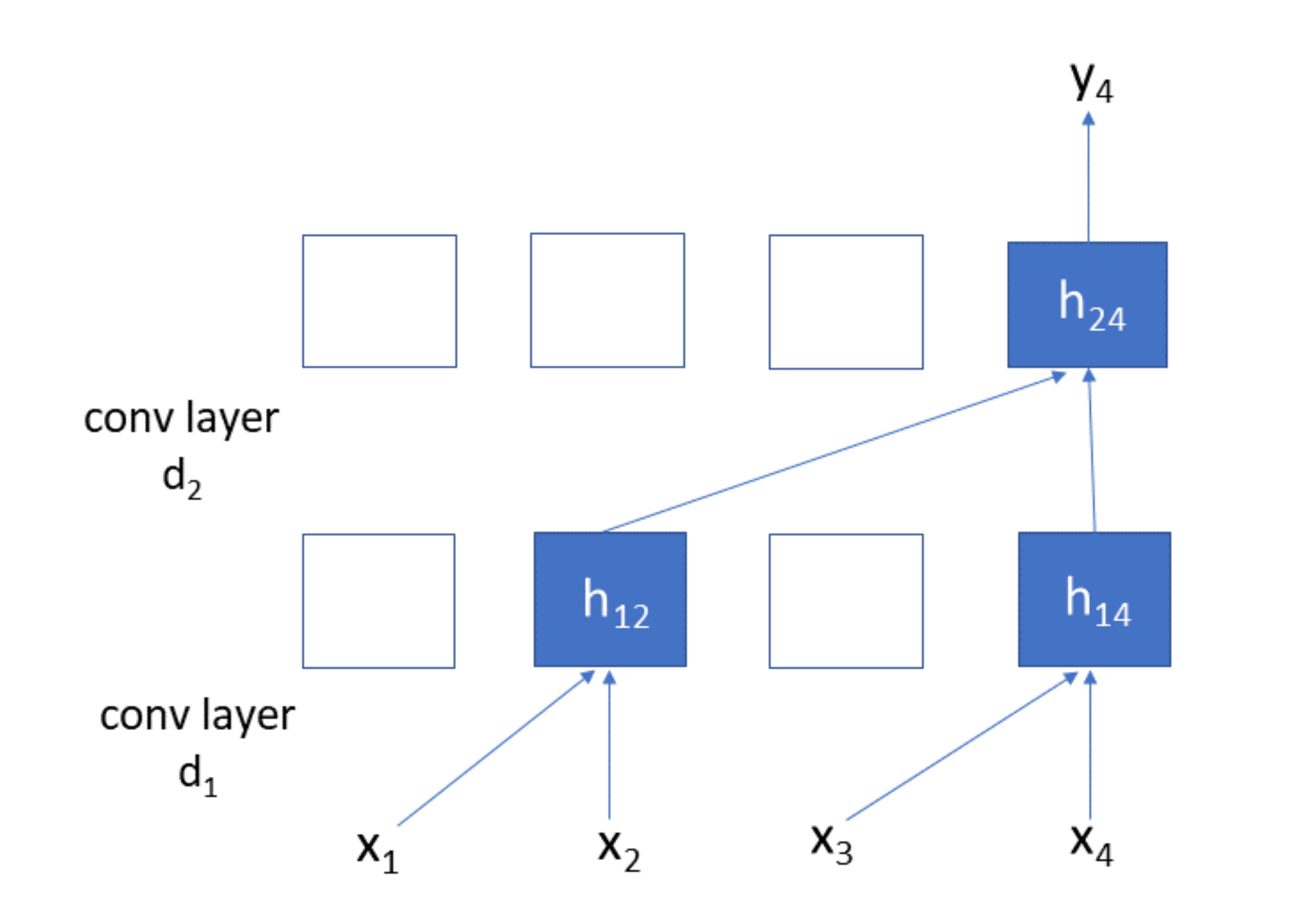}
 \caption{A neural network composed of two stacked convolutional layers with dilated convolution operations (dilation rate of 2) and kernel filters of size 2. Unlike conventional convolution, where the kernel is applied to consecutive samples, the kernel filter was applied to every second sample of the input $x$ and feature space $h1$.} 
\label{fig:dec_cnn}
\end{figure}

\subsubsection{Convolutional Neural Networks} \label{CNN}

Given that the training issues of RNNs (i.e. vanishing gradients) are not encountered with feed forward networks, feed forward networks have been proposed for sequence modeling. Given that feed forward networks are unable to represent sequences of a dynamic length (unlike RNNs), commonly sequences must be partitioned into fixed length blocks. Convolutional Neural Networks (CNN)~\cite{lecun_gradient-based_1998} are a class of regularized feed forward networks where only a local region of adjacent neural layers are connected. In each CNN layer, the receptive field of its neurons are restricted by convolving the data with a set of short filters (kernel filters) that are trained to represent the local structure of the data. 

CNNs have been used for classification~\cite{chen_ship_2020,ljunggren_using_2018} and prediction tasks~\cite{lv_t-conv_2018,leal-taixe_convolutional_2019} with trajectories. In these CNN approaches, 2D kernels are used to model the spatio-temporal trajectories that have been projected onto a two-dimensional spatial grid. The CNNs represent the local structure of spatio-temporal trajectories (in a similar manner to how images are modeled) to classify the general movement categories~\cite{chen_ship_2020} of marine vessels or vessel types~\cite{ljunggren_using_2018}. CNN architectures have also been used for predicting the future positions of taxis from their GPS (Global Positioning System) trajectories~\cite{lv_t-conv_2018,leal-taixe_convolutional_2019}.



The issue with using standard CNN architectures for spatio-temporal trajectories is their limited capacity to capture long-term spatio-temporal dependencies. This is a result of the kernel filters operating upon a small number of consecutive samples. To model temporal sequences across a wider horizon, however,~\citet{oord_wavenet_2016} used dilated convolution operations in their CNN layers. Figure \ref{fig:dec_cnn} shows a CNN architecture using dilated convolution operations, which perform the convolution between the kernel filters and non-consecutive data samples. Dilated layers enable the receptive field, and hence, the memory capacity of a network to expand exponentially with respect to its depth as opposed to linearly with standard CNN layers.

\citet{zhou_graph_2022} used a stack of dilated CNN layers to represent the temporal dependencies of a spatio-temporal graph network. The spatial dependencies between graph nodes, where each node was used to represent a unique location, were captured by the network edges. The graph network was conditioned upon the temporal dependencies at each location using a dilated CNN. Furthermore,~\citet{tran_learning_2015} jointly modeled the spatial and temporal dependencies of video data using volumetric CNNs; this was a 3D extension of the standard 2D CNN architecture. One issue with such an architecture is that spatio-temporal data can be sparsely distributed within discrete 3D space; this is often problematic when attempting to accurately represent a system of trajectories.

\subsubsection{Transformers}\label{Transformer}

\begin{figure}
\centering
 \includegraphics[width=0.98\linewidth]{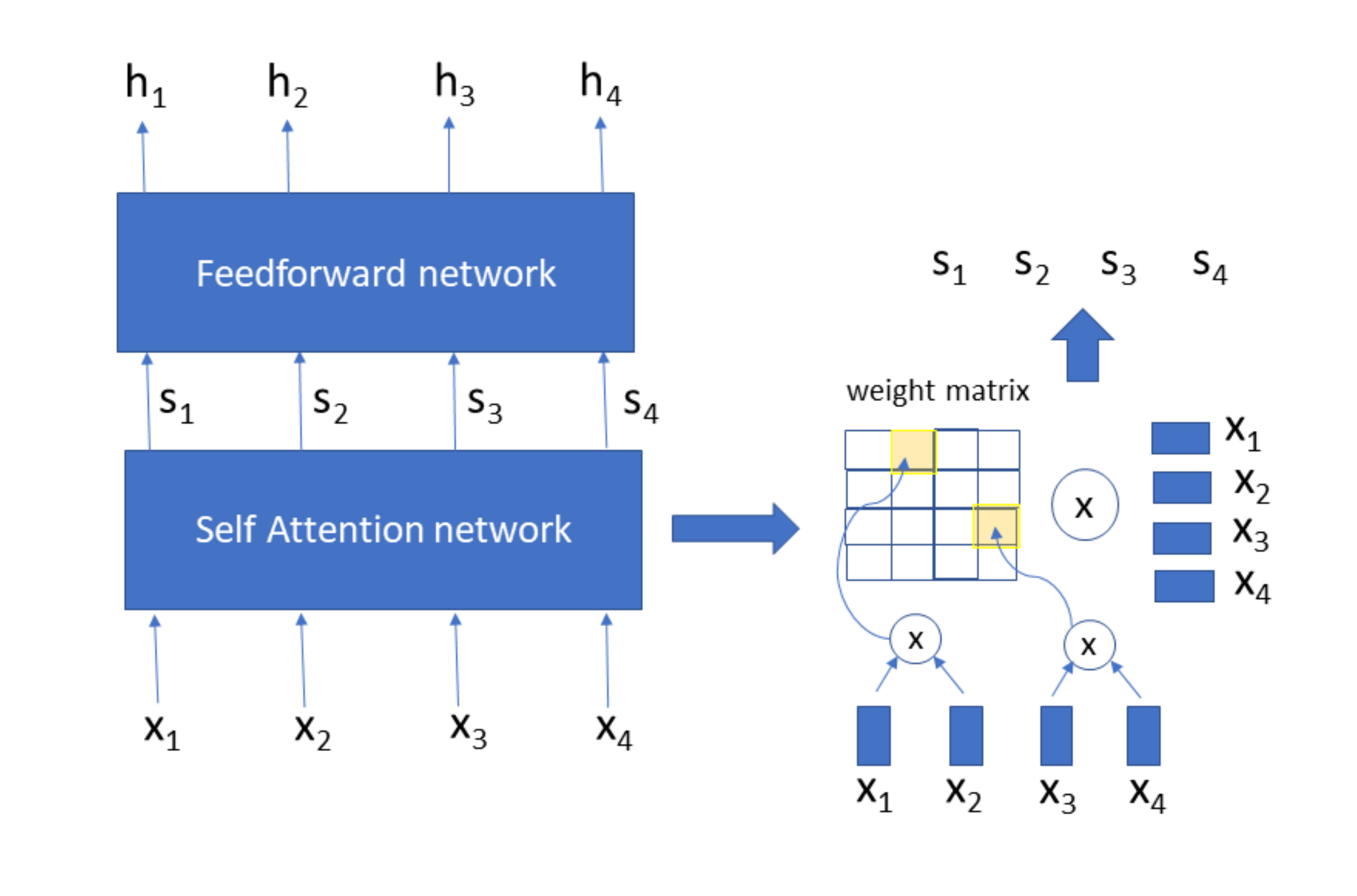}
 \caption{A Transformer network which is comprised of a self attention mechanism and feed forward network. The self attention operation explicitly models each sample pair in the sequence $x$. The resulting self attention matrix is multiplied by the sequence to generate the representation $s$. The representation $s$ is passed through a feed forward network to generate a latent encoding $h$ of the same length as the input sequence $x$.}
\label{fig:transformer}
\end{figure}

Transformers~\cite{vaswani_attention_2017-1} are currently the state-of-the-art approach for sequence modeling. Figure \ref{fig:transformer} shows a transformer network which is composed of a self-attention mechanism and a feed forward network. Self attention explicitly models the dependencies between each pair of samples in the sequence (self attention matrix), independent of their temporal proximity to one another, and hence, has an unlimited memory capacity. This comes at the cost of a quadratic order of memory and processing complexity, (e.g. $O(N^{2})$ where $N$ is the sequence length), which restricts the ability to train longer sequences. This becomes a technical challenge for modeling spatial trajectories that commonly represent sequences of longer duration and higher sampling frequencies. 

A number of approaches have been proposed~\cite{kitaev_reformer_2020,choromanski_rethinking_2021,beltagy_longformer_2020} to reduce the memory and processing complexity without experiencing a drop off in performance. Low rank or sparse approximations of sequences have commonly been used in self-attention computation. ~\citet{beltagy_longformer_2020} used a fixed prior of sparsity to define which subset of samples were used in the self attention computation. A fixed approach is likely to be sub-optimal, however, given it fails to consider the underlying sequence structure in sample selection. The Reformer~\cite{kitaev_reformer_2020} utilized a neural network to learn the most relevant samples to utilize in the self attention computation of each sequence. Furthermore, in~\citet{choromanski_rethinking_2021}, the commonly used softmax kernel was replaced by a fast and scalable  to produce a Transformer with linear and processing memory complexity.

\citet{giuliari_transformer_2021} proposed a Transformer based model to predict pedestrian movement using the spatial trajectories of individuals extracted from video. Unlike most contemporary trajectory prediction models that exploit the spatial interactions between pedestrians, the movement of each pedestrian was predicted solely from its own historical sequence data.~\citet{vedaldi_spatio-temporal_2020} modeled the spatio-temporal properties of a crowd by coupling a Transformer network, modeling each individual's movement, with a spatial graph network modeling the spatial interactions of the crowd.~\citet{wu_graph_2019} used a self attention mechanism to predict non-periodic traffic flow based upon the spatial trajectories of vehicles. A graph network representing the spatial structure of the road network was combined with a self attention mechanism (a core component of the Transformer) to capture the temporal structure of traffic status across multiple road segments.

\subsubsection{Sequence to Sequence}\label{Sequence to Sequence}

\begin{figure}
\centering
 \includegraphics[width=.8\linewidth]{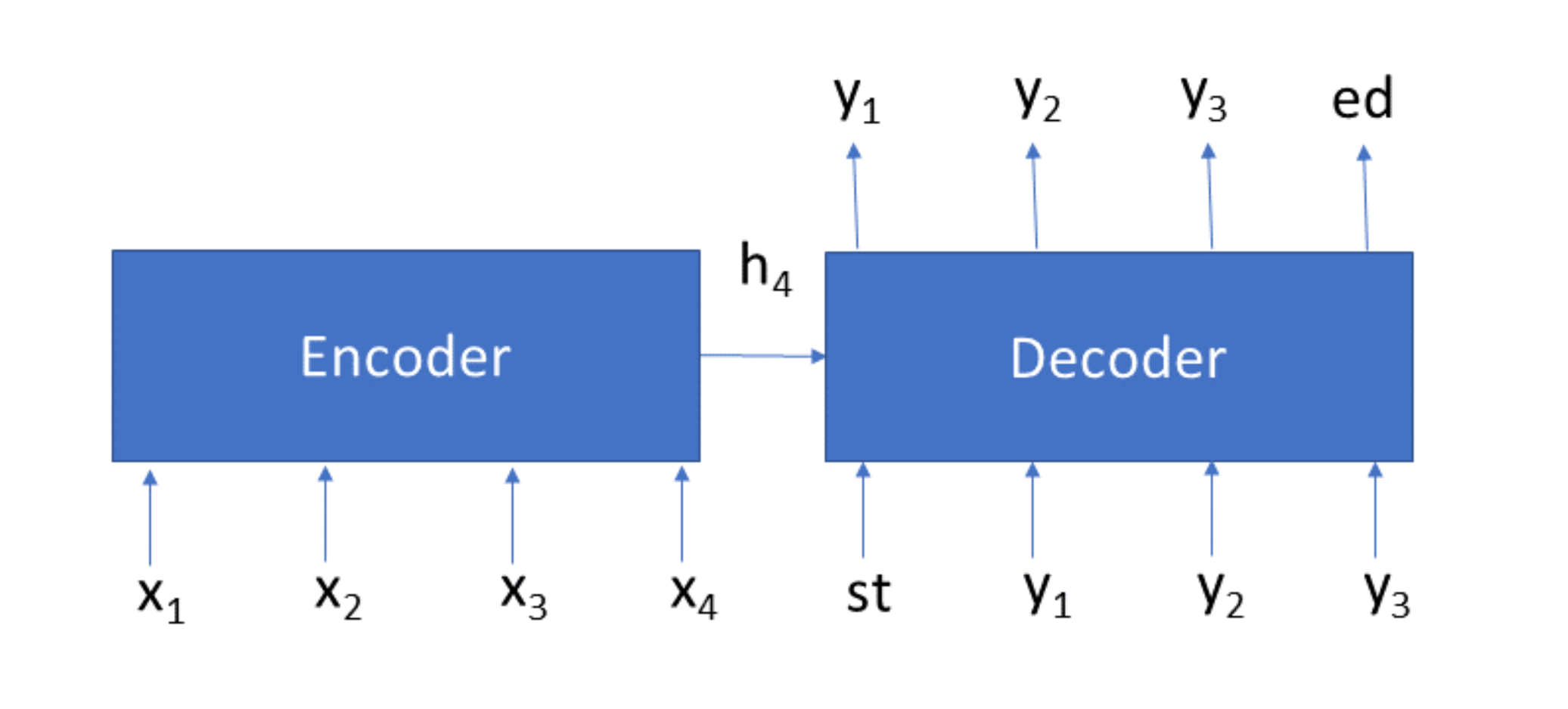}
 \caption{A sequence to sequence architecture which is comprised of a pair of encoder and decoder networks. An input sequence $x$ is encoded into a latent vector ($h$) that is used in conjunction with the decoder to generate an output sequence $y$, which can be a different length to $x$. A start symbol $st$ and end symbol $ed$ were used by the decoder to model the output sequence. }
\label{fig:seq_to_seq}
\end{figure}

Sequence to sequence (seq2seq) architectures~\cite{sutskever_sequence_2014} are commonly used to reduce the dimension of sequence data, or to produce an output sequence with a different length to the input sequence. The seq2seq architecture, as shown in Figure \ref{fig:seq_to_seq}, is comprised of a neural network pair: an encoder, which compresses the input sequence into a low dimension latent vector ($h_{N}$), and a decoder, which uses the latent vector to generate the output sequence. The type of neural network used in the seq2seq architecture is non-prescriptive, and hence, as we will outline, different sequence networks can be employed for trajectory based applications.  

A common application of seq2seq architectures is to harmonize data sets of spatial trajectories for downstream analysis tasks, such as classification~\cite{li_deep_2018} or clustering~\cite{yao_trajectory_2017}. For instance,~\citet{li_deep_2018} used an encoder and decoder pair to learn a latent vector of fixed dimension from a data set of noisy and heterogeneous trajectories (i.e. trajectories of uneven lengths and/or different sampling rates). The inputs to the sequence encoder were generated by randomly augmenting high quality trajectories into low quality versions by introducing noise and randomly dropping samples. The encoder and decoder pair were trained to reconstruct high quality trajectories from the compact latent vector that represent its low quality versions of the trajectories. ~\citet{yao_trajectory_2017} utilized a similar SSRL principle to ~\citet{li_deep_2018}, however, in this case, the latent encoding was used to reconstruct the input, as opposed to the higher quality version of the input. The latent representation of the trajectories were then used for clustering.

Seq2seq architectures can also be used for trajectory prediction. For instance, both~\citet{cinar_position-based_2017} and~\citet{park_sequence--sequence_2018} used a seq2seq architecture to learn a latent encoding of spatial trajectories that were then decoded into a sequence of predicted future samples. Furthermore,~\citet{zhao_deepmm_2019} used a seq2seq architecture to perform geometric map matching of spatial trajectories (an application outlined in Section \ref{sec:trajectories-preprocessing}) using a pair of LSTM based encoders and decoders to learn the mapping between augmented versions of raw spatial trajectories and its semantic sequence of road segments.

The Variational Auto-Encoder (VAE)~\cite{kingma_introduction_2019}, is a variational Bayesian method with structural similarities to the auto-encoder. It is also comprised of a encoder and decoder pair that are used to reconstruct the input sequence, but in the VAE, the latent (encoded) space is also regularized by representing it with a prior latent distribution. This makes the VAE a suitable model for sample generation by drawing samples from this latent distribution.~\citet{chen_trajvae_2021} utilized the VAE architecture with an LSTM based encoder and decoder pair to generate spatial trajectories. Furthermore,~\citet{zhou_trajectory-user_2018} used a VAE to address the social media based TUL problem outlined in Section~\ref{RNN}. Given semantic trajectories from social media are often sparsely sampled (i.e. due to infrequent check ins), the VAE was sampled to generate new trajectory instances. These generated, unlabeled trajectory instances were then leveraged to train a neural network classifier in a semi-supervised fashion for the TUL task.

\subsubsection{Generative Models}\label{Generative}

Generative models are unlike a majority of the ML models we have reviewed to date (apart from the VAE and Bayesian networks outlined in sections \ref{Sequence to Sequence} and \ref{DBN} respectively) given they are trained to represent the joint probability of all the variables as opposed to the supervised learning model that are trained to represent a conditional probability distribution of the annotated labels. 

In addition to the VAE models, Generative Adverserial Networks (GANs)~\cite{goodfellow_generative_2014} are currently the other type of generative model that are commonly used by the ML community. The GAN is an architecture comprised of a generator network, which is used to generate new data samples, and a discriminator network, which is used to discriminate between the generated samples and the real data. The GAN is trained in a competitive process where the generator network attempts to produce data samples that fool the discriminator network into thinking they are authentic samples of the original data set, whilst the discriminator network attempts to ensure the generated data can be distinguished from the authentic data. Once the generator network has been trained to adequately confuse the discriminator network, it can then be used to generate high quality samples of the original data set by sampling from it.       

GANs have been used to predict the future movement of a human conditioned upon their past trajectory observations \cite{gupta_social_2018,kosaraju_social-bigat_2019}. This architecture is known as a conditional GAN (cGAN), given the generation process was conditioned upon previous movement. ~\citet{gupta_social_2018} used a cGAN to generate trajectory predictions from the past samples of its sequence. The generator network was composed of an LSTM based auto-encoder that was used to generate trajectory predictions and an LSTM based discriminator network that determined if the movement predictions were socially acceptable. The social interactions were modeled with a shared pooling network that captured the spatial dependencies between the individual's trajectories.~\citet{kosaraju_social-bigat_2019} extended the cGAN to generate a more diverse set of movement predictions
by modeling the spatial trajectories as a multi-modal distribution. The generator network was an LSTM based autoencoder similar to~\citet{gupta_social_2018}, however in contrast,~\citet{kosaraju_social-bigat_2019} used two discriminator networks that operated at different scales; at the individual's level and across the entire crowd. A graph based attention network was used to model social interactions within the crowd by representing the spatial dependencies of trajectories.

\begin{summary}{Sequence Based Models}
The use of neural based sequence models has potential value to movement based manufacturing applications in the areas of pre-processing, classification, prediction and anomaly detection. Their main benefit is their capacity to model long term spatio-temporal dependencies of movement data, unlike the other methodologies outlined in this review (i.e. logic, probabilistic and state space models). Whilst each of the network models have their own memory and computational limitations, there are some new methods, such as the transformer variant of ~\citet{choromanski_rethinking_2021} with the capacity to model a sequence's entire past with only linear time complexity. Such new methods are an interesting option to consider when modeling trajectory data.

Movement data sets are often irregularly sampled and subject to measurement noise, which means 
they often need to be harmonized prior to modeling. This is another important task that can be addressed with neural based sequence models. Existing solutions to this problem are either computationally expensive, given non-linear distance functions are used, or can introduce additional noise when signal processing methods are used to harmonize the trajectories. SSRL methods enable raw spatial trajectories to be mapped into a compact feature space of size to address the noise and non-uniform sampling problems.

One of the areas where ML still has some limitations is its ability to include domain knowledge. Domain knowledge can be included into ML models in numerous ways; at the input, within the model structure, or as part of the cost function. There are numerous examples of domain knowledge being used by ML models in this review; one such area is map matching, where the spatial context provided by a map network is used to place constraints upon the model output. One general insight about the ways domain knowledge is included into ML model is that it can be rather ad hoc; more systematic approaches to integrating different knowledge classes would be beneficial. Furthermore, there are still restrictions in the types of knowledge that can be integrated into ML models. As an example, there remains an open question about how symbolic knowledge (i.e. logic programming, production rules) can be effectively represented, aligned and fused with a neural network. The next subsection discusses how this question is currently being considered with neural networks.

\end{summary}

\subsection{Integration of Logic-Based Methods with Probabilistic/ML Methods}
\label{sec:ML-non-ML-integration}
In this section, we briefly touch on how ML and logic-based methods can complement
each other to solve problems in movement analytics.
Generally speaking, ML methods perform
best when there is a large volume of data. ML models are intended to learn new features from data and
provide insights that are not possible otherwise.
 In contrast, logic-based models can be built solely based on domain knowledge and without
requiring any training data. 
Expressive logical languages are intended for representing structured data, i.e., objects with their
properties and relations among such objects. Logical inference of deductive,
inductive or abductive kind allows for deriving inevitable consequences,
conjectures and plausible explanations, respectively (Section~\ref{sec:classical-logic-KR}).\footnote{It can be non-obvious whether to apply ML or ``logic''. For example, a typical house
  floor plan places the entrance, living room and kitchen along this path. (This knowledge
  can be helpful for visual language navigation, see
  Section~\ref{sec:knowledge-graphs-VLN}.) This fact could be
  learned from video (real data), from plan drawings (simulation) or be axiomatized in a
  space-capable logic. Learning house floor plans in a different
  culture, when nothing is known a priori, could feasibly be done by learning from real
  world data, or by turning an expert's advice into logic, but perhaps not easily by
  simulation.}

At the same time, classical logic is often not suitable for modeling situations with significant amounts of data.  Classical
logic is interpreted under an open-world semantics. As a consequence, the \emph{only}
conclusions that can be drawn are those that 
follow inevitably from a set of given facts. However, in practice, it is often not practicable or
impossible to obtain all such required facts.

As a simple example, consider trajectories of items to be moved from $A$ to $B$, e.g.,
goods in a manufacturing setting.
Every trajectory can be ``normal'' or ``abnormal'', e.g., ``on time'' or ``late'' in terms of
movement time span. With classical
logic, one can attempt a formalization for proving that any given trajectory is either normal or abnormal. When
there are many trajectories or when knowledge is incomplete, this could be a difficult task.
The closed-world assumption ingrained into non-monotonic knowledge
representation methods (such as logic programming, cf.\
Section~\ref{sec:logic-programming-event-calculus}) enables more compact specification. It
enables assuming trajectories are normal \emph{by
  default}, and only the abnormal ones need to be singled out
explicitly by logical proof. (Or the other way around, but usually abnormal ones are expected to
be far fewer.) In practice, such axiomatization might still be impossible, too complicated to come up
with, or just not needed for the task at hand.

Many ML models are based on statistical analysis of historic data and
make it easy to classify trajectories based on relative probabilities (e.g., Bayesian
approaches). However, understanding (ab)normal trajectories in a
more analytical way might require 
contextual properties that are difficult to integrate (other than in some ad-hoc way, perhaps).
For example, it could be normal for an electric powered vehicle pausing for a long time when at
a charging station or at a weekend, otherwise it is abnormal; or relations with other
(ab)normal trajectories could be needed (e.g., on school excursions, younger students 
might be tolerated more ``erratic'' trajectories as normal compared to
elder students or teacher trajectories from $A$ to $B$). There might also be opportunity to apply
 optimization: what is an expected best trajectory given a history of abnormal
 trajectories from $A$ to $B$?

Such considerations motivate ``combining logic with ML'' for a best of both worlds result.
As for the method of doing that, one could take a logic formalism as a starting point and ``add probabilities'' to it. This idea goes back to (at least) 
\citet{nilsson_probabilistic_1986}. The main inference problem studied there is determining the probability of
an arbitrary query formula $Q$ given a set of logical formulas $F_i$ and their
probabilities $P(F_i)$.  
The probabilities can be seen as soft constraints on interpretations satisfying these formulas. This is different to probabilistic logic programming, see Section \ref{sec:statistical-relational-learning},  which assigns probabilities to relational instances given or derived by a probabilistic logic program. One could also start from a probabilistic framework and ``add logic'' to it. Or, one could try hybrid black-box style combinations. 
\citet{davis_symbolic_2020} presents an in-depth overview of issues and options
around combining logic and learning starting from such high-level considerations.



The recently introduced notion of ``Cognitive AI'' provides motivation at a more strategic level.
\citet{singer_rise_2021} discusses capabilities and limitations of deep learning
(DL). DL excels in the classification of broad and shallow data (for example, a sequence of words
or pixels/voxels in an image) and indexing very large sources (such as
Wikipedia) and retrieving answers from the best matching places.
Principled limitations are centered around model scalability and missing competencies
for abstraction, context, causality, explainability, and intelligible reasoning.
He proposes a new phase of ``Cognitive AI'' to achieve these competencies through the
integration of DL with symbolic reasoning and deep knowledge.
This will require reasoning over deep knowledge
structures, including facts and deep structures of declarative (know-that), causal
(know-why), conditional/contextual (know-when), relational (know-with), and other types of
models. (These are capabilities that are well within the scope of ``logic''.) He expects significance of this approach in robotics, autonomous
transportation, logistics, industrial, and financial systems.
\citet{marcus_next_2020} argues in a similar direction to \citet{singer_rise_2021}.

The Cognitive AI approach could possibly be supported by a
high-level workflow with ongoing adjustments based on feedback:
\begin{list}{}{}
\item sensors $\to$ ML $\to$ logic $\to$ decisions $\to$ actions $\to$ sensors.
\end{list}
One could also consider combination architectures for decision making on top
of data and experiences learnt over time (perhaps similar to how people do). 
A suitable ML framework for that could be reinforcement learning.

In the following, we overview some established approaches for combining ML and logic.
We need to emphasize, though, that the scope is somewhat confined to techniques that seem useful for movement analysis. \citet{marcus_next_2020} is a good starting point for considerations on combining ML and symbolic reasoning more generally.
We will conclude it with an assessment of the current state of the art
for application to movement analytics and ideas for future research.


\subsubsection{Statistical Relational Learning}
\label{sec:statistical-relational-learning}
Statistical Relational Learning (SRL) is a research area that is concerned with domain models that exhibit both uncertainty and complex, relational structure. It utilizes first-order logic, for modeling relational structure, and probabilistic methods, mainly graphical methods such as Bayes networks and Markov networks, for probabilistic inference and learning. 
What separates SRL from "non-relational" learning algorithms, e.g., decision trees,
support vector machines, Bayesian networks,
is their ability to take the relationship between the individuals as well as their internal structure into account. As~\citet{koller_probabilistic_2009} explain under the
notion of  \emph{template models}, SRL
supports more complex modeling than is possible with a fixed set of propositional
variables. In a temporal setting, for example, one can express general rules for how
system state distributions evolve over time, which then can be applied to individuals
(e.g., current battery charge as a function of battery capacity and aggregated usage). In a non-temporal
setting, they can express general rules or relations that apply to all or certain classes
of individuals (minimum battery capacity by
vehicle type and mission requirements) or combined. Sometimes these relationships can be
discovered automatically (learning) or the can be given as background knowledge for the
purpose of informing learning and/or probabilistic inference.  
 
 As an early application example,~\citet{natarajan_early_2013} used SRL for mining non-obvious statistical dependencies in health record databases for patient risk assessment.
Applications of SRL methods for object tracking have been developed by \citet{limketkai_relational_2005} and by \citet{nitti_relational_2014}.

The books by \citet{koller_probabilistic_2009} and by \citet{raedt_statistical_2016} are
comprehensive introductory texts and overviews (the latter with a focus on probabilistic
logic programming) covering SRL languages and semantics, probabilistic inference, and parameter and structural learning.

In the following we briefly summarize some of the main approaches to SRL with relevance to movement analytics.

\littleparagraph{Relational Dynamic Bayes Networks}~\cite{sanghai_relational_2005} generalize dynamic Bayes networks (see Section \ref{sec:dynamic-bayes-networks}) to relational data. Relational properties are to be specified with first-order logic. The formulas are interpreted on probabilistic facts whose distributions are presented in a certain way (first-order probability trees). 
\citet{vlasselaer_efficient_2014} express Relational Dynamic Bayes Networks as a probabilistic logic program. They argue for better scaling behavior by exploiting structure.  


\littleparagraph{Markov Logic Networks}~\cite{richardson_markov_2006} combine Markov 
networks and ﬁrst-order logic by
attaching probabilities to ﬁrst-order formulas, similar to~\citet{nilsson_probabilistic_1986}.
An Markov Logic Network  program $L$ is a set of pairs $L = (F_i, w_i)$
where $F_i$ is a formula and $w_i$ is a real number representing its weight.
Positive/negative weights increase/decrease probability
that the formula holds true in a world. Formulas act as templates, i.e., under a finite Herband semantics, thus allowing reduction to Markov Networks. 
(A Markov network is similar to a Bayesian network in its representation of dependencies, the differences being that Bayesian networks are directed and acyclic, whereas Markov networks are undirected and may be cyclic.) 
See \cite{domingos_unifying_2019} for a more recent overview article.


\littleparagraph{Probabilistic logic programs (PLPs)} are logic programs extended with
facilities to express probabilistic facts or conclusions. Roughly speaking, traditional logic programs~\cite{kowalski_algorithm_1979} generally consist of if-then rules and atomic facts. The rules typically contain variables for the elements of the domain of interest. Rules act as templates then. Rule conclusions can be disjunctive, or not, and the conditions can use ``default negation'' for enabling a closed-world semantics. Facts are statements of unconditional truths. The main reasoning tasks are query answering, akin to database query answering, and model computation. Model computation means to extend the given facts to an interpretation that also satisfies the relations defined by the rules (``model''). These tasks require computationally rather different methods and are typically used for rather different purposes, giving rise to corresponding sub-paradigms.  

In PLPs, facts are equipped with probabilities. A PLP then induces a joint distribution over all predicates (relations) according to the (generally accepted) distribution semantics~\cite{sato_statistical_1995}. Reasoning services include computing joint distributions, marginalizing, and query answering (posteriors). Some PLP languages are rather general and can express, e.g., Bayes networks, (hidden) Markov models, and Dynamic Bayes Networks.
See also Section~\ref{sec:dynamic-bayes-networks} for Dynamic Bayes Networks and their expressivity.

An advanced implementation of PLP is the ProbLog system~\cite{de_raedt_problog_2007,dries_problog2_2015}.
\citet{kifer_survey_2018} provide an overview over probabilistic logic programming.

In Section~\ref{sec:logic-programming-and-machine-learning} we return to logic programming
combined with ML.

\littleparagraph{Lifted Inference}~\cite{poole_first-order_2003} is a term that characterizes
methods that aim at inference directly at the first-order logic level, rather than through
exhaustive instantiation (cf.\ Markov Logic Networks above, or the ``weighted model counting'' approach by
\citet{gogate_probabilistic_2016}). The quantifiers of first-order logic enable reasoning with collections of
individuals as a whole, which can be much more efficient (and general) than reasoning on the individual level. \citet{riguzzi_survey_2017} provide a survey of lifted inference for probabilistic logic programming.

\littleparagraph{Learning.} 
All of the above approaches support ``learning'' in one way or the other. 
Parameter learning in context of probabilistic logic programming, for instance, means learning the probabilities attached to facts. This can be done, for instance, with Monte-Carlo simulations and gradient descent optimization. 

Structural learning means learning a logic program. The underlying area of inductive logic programming is among the most traditional ones in ML.
\citet{muggleton_inductive_1994} and
\citet{de_raedt_probabilistic_2008} provide overviews.
It has gained renewed interest in conjunction with knowledge graphs, see below.

\citet{davis_symbolic_2020} provides a brief but in-depth overview of issues and options around combining logic and learning.

\subsubsection{Knowledge Graphs and Visual Language Navigation}
\label{sec:knowledge-graphs-VLN}
Knowledge graphs are edge-labeled graphs for representing triples subject/predicate/object (cf.\ ``semantic web''). Known for a long time, there is renewed interest for application to mined data sets stemming from, e.g., social networks. Knowledge graphs are typically large, incomplete and noisy. 
Dealing with these problems by ML has become a vast area by itself.  \citet{liu_network_2021} provide an overview.
\citet{zhou_graph_2022} cover the more general area of graph neural networks.

One important reasoning task, among others, is link discovery. For example, if $A$ and $B$ are known to be co-authors of a certain paper, $B$ is student, and $A$ is professor then, with
some high probability, one should conclude ``$A$ supervises $B$''.
Such rules can be learned, see, e.g.,~\cite{gu_towards_2020,pellissier_tanon_completeness-aware_2017}.

\citet{chekol_time-aware_2019} propose Markov logic networks for computing marginal probabilities in large knowledge graphs, the Yago ontology~\cite{harth_yago_2020}.
\citet{gutierrez-basulto_reasoning_2020} bring together probabilistic reasoning for knowledge graphs and logic programming. They utilize a certain 
dialect with attractive computational features (warded Datalog) for bottom-up model computation in a relational database framework (tuple-generating dependencies, Chase algorithm). Their method targets computing probability distributions of interest but does not support learning so far.

Knowledge graphs have become relevant for \emph{visual language navigation (VLN)}. 
VLN aims to enable embodied agents to navigate in realistic environments using natural language instructions. It requires machine-understanding of video-sensed trajectories on-the-fly as a robot moves along through known or unknown territory. Current research recognizes that background knowledge can be helpful. For example, a typical house plan puts the living room between the entrance and the kitchen; often, printers are located in the office. The approach by \citet{wu_bayesian_2019} utilizes such prior knowledge for landmark-based navigation planning. 

Considering VLN research in context of movement analytics may seem far-fetched. However, as said, VLN navigation planning requires (on-the-fly) derivation of semantic trajectories, so that the robot's plan execution can be aligned with its current position, all in symbolic terms. Techniques for that seem well usable  for industrial movement analytics in applications with on-board video.  


\subsubsection{Logic and Deep Learning}
\label{sec:deep-learning}
Logic can be a helpful mechanism to capture domain knowledge in deep learning
architectures. \citet{dash_review_2022} 
distinguishes several ways to integrate such knowledge: Introducing background knowledge into deep network  
(a) by transforming data; (b) by transforming the loss function L; and (c) by transforming the model (structure and parameter). The following are some examples of these different categories.

  
\citet{wan_neural_2018} falls in category (a) and present an approach to add a set of auxiliary inputs to help interpret the outcome of a neural network. Input data is passed through a neural network to generate the auxiliary inputs to the next network. The number of outcomes is incremented by one to cater for a conflict class. Logic can be used instead of the first network to generate the auxiliary inputs. \citet{al-shedivat_contextual_2020}  devised a method for classification dependent on context and falls in category (a). The main point is to provide explainability to a human user. The method integrates a context-encoder as the front-end to neural networks. The context encoder is supposed to produce human-understandable classification. The example given in the paper classifies a household as ``rich'' or ``not rich'' dependent on location context, the household location determined by satellite image.

\citet{hu_harnessing_2016} falls in category (b) and present an  approach to integrate logic and ML. Two networks are trained in parallel---a teacher network and a student network. The student network learns from the data but the loss function is guided by the teacher network. The teacher network is constrained by the logic-based domain knowledge and the loss goes up if domain constraints are not satisfied. This way the student network is indirectly influenced by the domain knowledge based constraints expressed as logical expressions.

\citet{li_augmenting_2019} falls in category (c) and present an approach to add extra layers of logic neurons in an existing structure of neural network. The inputs and outputs of such neurons approximate binary propositional logic functions. To make the functions differentiable, a T-norm representation of the logic gates is presented. The paper is achieving better results on a number of ML problems. \citet{shi_neural_2020} also falls in category (c) and present an approach to represent logical expressions as neural networks. The method itself is not an integration of logic with ML. However, the logic expression structures proposed in the paper can be used as part of constructing logic driven neural networks.





\subsubsection{Logic Programming and Machine Learning}
\label{sec:logic-programming-and-machine-learning}
While inductive learning and parameter learning is not something new for (probabilistic)
logic programming (Sections~\ref{sec:logic-programming-event-calculus}
and~\ref{sec:statistical-relational-learning}), there are more crossover areas,
  some rather recent.

In one area, Datalog is used as a specification language to express ML
applications~\cite{bu_scaling_2012,li_mlog_2017,wang_formal_2021}.
\citet{bu_scaling_2012} put it concisely and observe that training ML architectures builds
on a core set of capabilities for \emph{search}, \emph{iterative refinement} and
\emph{graph computation} on big data sets.  The common theme behind the cited papers then
is a proposal for Datalog (or a related language) as a declarative alternative to
traditional imperative programming for realizing these capabilities. The main motivation
is simplicity and conciseness without loosing efficiency.

Another area could be called ``Deep Logic Programming''. 
This term is meant to subsume hybrid methods that integrate logic programming with 
neural networks with ``many'' layers. In contrast to the first area, logic is used this time for capturing domain
properties, not for algorithm development. To date, there are only a few developments in this
direction.

\citet{mei_neural_2020} address the problem of how to predict future events from patterns of
past events, which is difficult when the set of possible event types is large (risk of
overfitting). For example, a pattern of people traveling between cities does in general
not depend on event types like soccer goals and wheat sale. In their hybrid approach, a
domain expert writes down the rules of a logic program that tracks the possible
\emph{relevant} event types and other Boolean facts over time. This logic program is then
used to automatically construct a deep recurrent neural architecture for every pattern
derivable from facts and rules (every derived fact provides a new layer). This neural
network can be trained by backpropagation to recognize event types having happened and
predict the next event. In terms of the movement example above,  it can learn a neural probabilistic model of
object movements while relying on a discrete symbolic deductive database to cheaply and
accurately record what  is where.

\citet{manhaeve_neural_2021} develop
DeepProbLog, an extension of ProbLog (Section~\ref{sec:statistical-relational-learning})
that integrates Probabilistic Logic Programming
with deep learning. In ProbLog, the probabilities of all random choices are explicitly
specified as part of probabilistic facts. DeepProbLog generalizes
ProbLog's facts to ``neural predicates'' whose probabilities are parameterized by neural networks. 
In this regard, DeepProblog is loosely related to the
\citet{mei_neural_2020} approach (which however, is not based on \emph{probabilistic} logic).
The neural network represents a classifier for the probability of a derived fact being
true. A good example for illustration is character recognition, where neural predicates associate
probabilistically hand-written characters with their actual symbolic interpretation.
DeepProbLog supports probabilistic inference and various learning schemes.







\begin{summary}{Integration of Symbolic/Logical Methods with Probabilistic/ML Methods}
\littleparagraph{Logic Programming and Probabilistic Methods.} Existing research in SRL
(Section~\ref{sec:statistical-relational-learning}) already has a lot to offer:
representing structures of objects and their relations, native support of time (e.g.,
dynamic Bayes networks), probabilistic reasoning and parameter/structure learning.
However, it is not obvious if the existing methods offer all needed reasoning services for spatio-temporal applications. It is also not obvious if the first-order language fragment they offer for specifying background knowledge is expressive enough. Integrating numerical domains or other background theories is a topic that received considerable attention in automated reasoning in first-order logic, 
and we see scope for making SRL language more expressive.
These could be done via natively integration or in a modular way by combining SRL methods with classical logic or related methods on spatio-temporal reasoning. Given the rich state of the art in this area (cf.\ Section~\ref{sec:classical-logic-KR}) such combinations could be worthwhile to explore. 
Probabilistic logic programming appears to be an appropriate framework for that.

Research in this direction seems to gain momentum. 
In Section~\ref{sec:logic-programming-and-machine-learning} we reviewed two hybrid logic
programming and neural network combinations which do support learning. 
The mentioned papers by \citet{mei_neural_2020} and by \citet{manhaeve_neural_2021} address the challenge of making neural networks classification less brittle by adding logical domain knowledge to the picture. These approaches look very appealing for movement analysis, e.g., for robust classification of movement patterns. 

The LARS~\cite{beck_lars_2018} system mentioned in
Section~\ref{sec:logic-based-stream-processing} technically is a logic programming system
with dedicated support for time-windows based reasoning on data streams. The LARS 
language has recently been extended by 
\emph{weights} attached to its formulas (rules)~\cite{eiter_weighted_2020}. Weights can be
given a probabilistic interpretation similar to Markov Logic Networks, see
Section~\ref{sec:statistical-relational-learning}. The extended LARS language also
generalizes the ProbLog probabilistic logic programming language, see
again~\ref{sec:statistical-relational-learning} (but not its ``learning'' aspects).
Interestingly, \citet{ferreira_deep_2021} recently utilized deep learning architectures
for time-series analysis to efficiently approximate LARS inferencing. Research on those aspects currently seems to be focused more on fundamental results and first implementations.

The most integrated approach we are aware of is by  
\citet{katzouris_online_2021}. They implemented an answer-set logic programming system
that  is capable of combining temporal reasoning via the event calculus under uncertainty
via probabilistic logical inference, with online structure and parameter learning.  The
intended application area is stream processing
(Section~\ref{sec:logic-based-stream-processing}). The approach has been tested on large ship
movement data and is capable of \emph{learning} rules  for anomaly
detection in trajectories.

PLP rules of the form ``if $\phi$ then $y$ holds with probability $p$'' represent conditional probabilities $P(y \mid \phi)
= p$ (but are more general). In a Bayes model, the joint probabilities of generative models are computed
from conditional probability tables. In the simpler cases, a non-probabilistic logic
programming rule ``if $\phi$ then $y$'' can be taken as a PLP rule with probability $p =1$. The
 answer-set model (Section~\ref{sec:logic-programming-event-calculus}) then contains a joint probability query if and only if the query probability is $1$.  These considerations suggests a bridge from
 knowledge-representation practice and experience to the probabilistic case. It might be
 worth exploring for movement analytics.

On the more speculative side is a connection between movement analysis and graph formalisms. As observed in Section~\ref{sec:knowledge-graphs-VLN}, research in visual language navigation recognizes the importance of background knowledge for robot navigation tasks and is exploring knowledge graphs for that. 
This opens opportunities to exploit knowledge graphs technology as a tool for mining semantic trajectories from video.

\littleparagraph{Deep Learning.}
``Integration of Machine Learning with Symbolic AI'' is a popular even if not a very precise term.
Industry has promoted a similarly vague the term ``Contextual AI'' as an overarching term for the next
``third'' phase of AI~\cite{launchbury_darpa_2017}. 
Contextual AI is advertised as the integration of the two major earlier phases, classical AI (rule-based,
logical reasoning) and statistical  AI (machine learning). One of the aims of contextual
AI is to add \emph{explainability} to ML. In image classification, for example, an
ML-recognized ``cat'' could be augmented by explanations like ``has fur'', ``has whiskers'', etc. 
Few academic papers seem to explicitly mention the term ``contextual AI'', though.


To achieve these objectives, the technical backbone on the ML side is commonly understood as neural networks equipped with deep learning. Indeed, this is a hot topic, see Section~\ref{sec:deep-learning} for some recent work.
Our assessment from these works is as follows.

The maturity of the current state of the art depends on the architecture of the integration. Hierarchical methods with a clear-cut interface are easier to achieve both theoretically and for practical applications. Much harder is the tight integration of logical inference as an integral part of the neural network architecture. For example, approaches based on dedicating neurons for representing propositional Boolean functions work only under rather strong assumptions and seem to have limited applicability.

In summary, we assess that integrating first order logic as part of the deep learning framework for incorporating domain knowledge is still an emerging area as a whole, and not widely explored for movement analytics at all. 



\end{summary}

\subsection{Constraint Optimization}
\label{sec:constraint-optimisation}

Constraint optimization for manufacturing is about formalizing the real-world problem as a
set of constraints and decision variables forming the model, and solving this model by optimizing
an objective function (or multiple objective functions), so that all constraints are satisfied in the model.
For example, a classical problem is the production scheduling, in which a schedule is sought that
minimizes the project end by deciding the job-process order on each machine.
Constraint optimization is an established research field
with the notable sub-fields Operations Research, Constraint Programming, and Mixed-Integer Programming
and has attracted much research in the modern area.
Due to that there is a plethora literature about different optimization techniques and 
solution methods for various manufacturing problems.
We refer the keen reader to these surveys and books as a starting point (see, e.g., 
\citet{baptiste_constraint-based_2001, schwindt_handbook_2015, schwindt_handbook_2015-1, ouelhadj_survey_2009, liu_survey_2018, wari_survey_2016}).

The remainder of this sub-section focuses on optimization aspects related to
big data and movement analytics.
We note that to best of our knowledge constraint optimization is used as a ``consumer'' of
data from big data and movement analytics rather than a ``producer'' or an ``enhancer'' of such
data.

\subsubsection{Production Scheduling}

The majority of the problems in production scheduling are on the tactical and operational level.
The former one is concerned with generating a ``baseline'' or ``master'' schedule before the start
of production and the latter one to react to unfolding events, e.g., machine failures, 
resource unavailabilities, delays, and new orders, during the production.
As~\citet{ovacik_exploiting_1994,ovacik_decomposition_1997} point out,
one of the main issues for optimization is the lack of accurate information, e.g., the length of
the processing times, which results in uncertainty in a solution.
In order to decrease the uncertainty on both levels, recorded real-time information,
which can come from, e.g., sensors, scanners, bar-codes, and computer terminals,
is used to improve estimated values of some data, e.g., worker performances and job processing
times~\cite{cowling_using_2002}.
These improved data are then fed to the standard solution approach for project scheduling.

On the operational level, the real-time information is digested in a timely manner and fed into
a framework, which decides what action to take or not.
\citet{cowling_using_2002} developed a general framework, in which \emph{events} (i.e., the arrival of 
new real-time data) are put through a four-stage process:
\emph{detection} (via, e.g., sensors, scanners, bar-codes, and computer terminals), 
\emph{classification} (classifies the event and decides whether the event is handled automatically or
manually), 
\emph{identification} (regular or not, and reasons), and
\emph{diagnosis} (decides whether to take no action or perform a limited repair or reschedule from
scratch). 
Constraint optimization is especially important for the last stage, for which an optimal action strategy
is seek, and the repair and the reschedule are optimization problems.
In general, the action strategy is based on quantitative measures of utility
(i.e., the improvement in ``baseline''/``master'' scheduling objectives due to schedule revision)
and stability measurement
(i.e., the disruption caused by schedule revision).
\citet{ouelhadj_survey_2009} presented a review of the optimization techniques used for real-time optimization using the data analysis (movement and disruptions) in the manufacturing space.
Moreover, \citet{dobler_supporting_2020} has combined big data with optimization for optimizing 
job assignment by comparing the static optimization against a real-time situational awareness 
digital avatar where real-time situational awareness will inform about events such as, e.g., disruptions, 
machine failures.

\subsubsection{Factory Layout}

The factory layout problem concerns the spatial positioning of work
stations, machines, tools, and/or functional areas in order to increase the efficiency of the
production flow while respecting the application specific constraints,
e.g., building boundaries and pathway dimensions.
There is a significant amount work on optimizing the layout by simulation
of production schedules on a virtual digital factory environment on simulated data (see, e.g., \citet{centobelli_layout_2016-1, lee_construction_2011-1, kanduc_optimization_2015, herr_bluecollar_2019}),
which can be partly based on real-time data and/or movement analytics on the current factory setup.
Different objectives or combination of have been studied, e.g.,
minimum travel time/distance~(see, e.g., \citet{dzeng_application_2014,kanduc_optimization_2015, herr_bluecollar_2019}) and 
minimum production lead time~(see, e.g., \citet{centobelli_layout_2016-1}).

For instance, \citet{dzeng_application_2014} use movement analytics for allocation of functional space in a
facility while minimizing the travel distance required by the worker during their daily activities and
incorporating preference of space sizes.
For movement analytics, the worker's movement are tracked through the facility using RFID 
technology.
Then the tracking data is mined to determine movement patterns and the relation values between
functions.
This information is fed into an optimization algorithm, here a genetic algorithm, to solve the problem
at hand.

There is also other related work on factory layout that focus on building construction
in general using movement analytics.
\citet{duan_emerging_2020} review how RFID technology can be used to integrate the movement analysis
for better tracking and hence making efficient decisions in construction related and other decisions
at different stages of the life cycle of a building. While they focused on the use of RFID
technology, \citet{li_real-time_2016-1} reviewed all the use of different real-time locating systems
developed in construction sector to identify and track the location of an object in both indoor and
outdoor environments and support the decision making. 
\citet{du_gaps_2020} has also presented the review of how movement data can be used in the construction
of building and deciding optimal layout. 
However, their study focused on the work done in optimizing the energy consumption.


\begin{summary}{Constraint optimization}
Constraint optimization is a consumer of curated data from big data and movement analytics to make better 
decisions in production scheduling and factory layout problems.
The curated data can provide following things or combination of as an input for optimization methods:
\begin{itemize}
    \item more realistic value estimation, e.g., job processing times,
    \item real-time information about values and disruptions for rescheduling and/or repairing a schedule in action, and
    \item discovery of work patterns, e.g., worker's movement patterns between different functional spaces.
\end{itemize}
Since the optimization methods are the consumer of such information, standard optimization techniques 
can be applied on it.

For instance, a worker requires a tool to process a job, but always has to leave their work area
to obtain it, which significantly contributes to the processing time.
By tracking the movement of the worker and what job they process, movement analytics can
identify that the worker always leaves their work area for some time when processing this particular job.
This knowledge can be used to, e.g., a permanent placement of the tool at the work area or nearby
to decrease the processing time of the job, or
a creation of a new job for the retrieval of the tool preceding the
actual processing job to reflect the reality better.

Movement analytics along optimization has been applied for better path planning of unmanned vehicles
to avoid obstacle collisions and to minimize path times and turns (see~\citet{yang_obstacle_2022, lutz_temporal_2008}) in the transportation area.
These developed methods for transportation are also applicable in manufacturing under the assumption
of a fixed factory layout, when transportation of raw material, intermediate, and final products
through and around a factory are critical of the operation of the factory.

An interesting area is to explore architectures for combined ML, logic and constraint
optimization techniques. For instance, logic programming could be tried to 
bridge a semantic gap between ML methods for, e.g., trajectory segmentation and
classification on the one hand, and constraint 
optimization on the other hand.
\end{summary}

\section{Applications of Movement Analytics}
\label{sec:applications-movement-data-analytics}
In this section, we review research-intense applications of movement analytics in a manufacturing context. We focus on
approaches that are based  
on positioning data, but we briefly touch on non-positioning data as well
(Section~\ref{sec:applcations-non-positioning-data}).  

Position data-based production tracking has the potential for optimizing production processes
in manufacturing.
We focus on domain specific applications or case
studies considered, models applied, trajectory data representation and cleaning process,
and data sets used (if any).
In contrast, the subsequent Section~\ref{sec:industrial-applications} looks at applications from a more industrial perspective.

Table \ref{tab:refpapers_manufacturing_app} presents a list of prominent research papers in the literature, categorized according to their applications.

\begin{table}[]
    \centering
    \resizebox{\textwidth}{!}{%
      \begin{tabular}{m{3cm}m{0.9cm}m{3cm}m{2.5cm}m{3.6cm}}
    \hline
    \textbf{Application} & \textbf{Ref paper} & \textbf{Approach}  & \textbf{Model} & \textbf{Data processing} \\
    \hline \hline
        \multirow{3}{1.5cm}{Workflow evaluation} &~\cite{arkan_evaluating_2013} & Key Performance Indicators (KPIs) computation & & Layout based trajectory mapping \\
        \cline{2-5}
        &\cite{gyulai_analysis_2020} & KPIs computation & & Trajectory smoothing using S-G filter, and layout based trajectory mapping\\
        \cline{2-5}
        

        & \cite{racz-szabo_real-time_2020-1} & Clustering \& KPIs computaion & k-means &\\
         \hline
         
         \multirow{3} {1.5cm}{Collision avoidance, human-robot collaboration} &  \cite{locklin_trajectory_2020} & Trajectory prediction & LR & \\
         \cline{2-5}
         
         & \cite{cheng_towards_2020} & Trajectory prediction \& plan recognition & LSTM networks, Bayes rule & \\
         \cline{2-5}
         
         & \cite{zhang_recurrent_2020} & Trajectory prediction & LSTM networks & \\
         \cline{2-5}
         
         & \cite{wang_location_2021} & Trajectory prediction & k-NN & Access points based trajectory mapping\\
         \hline
         
         \multirow{3} {1.5cm}{Frequent path or trajectory patterns} & \cite{cai_mining_2017} & Data mining & Apriori algorithm & Trajectory mapping using data cubes \\
         \cline{2-5}
         
         & \cite{bu_data_2018} & Data mining & Apriori algorithm & \\
         \cline{2-5}
         
         &\cite{liu_mining_2012} & Analytical & &  Raw trajectory to binary trajectory using Chebyshev’s inequality\\
         \hline
                  Indoor space modeling & \cite{han_data_2014} & Clustering & GDBSCAN \cite{sander_density-based_1998} & Trajectory segmentation using TRACLUS \cite{lee_trajectory_2007}\\
         \hline
         Event detection &  \cite{flossdorf_unsupervised_2021} & Analytical & DBSCAN &\\
          \hline
           \multirow{5} {1.5cm} {Others} & \cite{syafrudin_performance_2018} & Big data analytics & RF & Data cleaning based on DBSCAN\\
           \cline{2-5}
           
           & \cite{tao_data-driven_2018} & Big data analytics & NNs & \\
           \cline{2-5}
          
           & \cite{zhang_framework_2017} & Big data analytics & k-means, association rule mining & \\
           \cline{2-5}
           
           & \cite{ji_big_2017} & Big data analytics & Rapid-Miner & \\
           \cline{2-5}
           
           & \cite{zhong_mining_2014} & Clustering \& prediction & SVM, DT &\\
          \hline
    \end{tabular}}
    \label{tab:refpapers_manufacturing_app}
     \caption{A list of prominent studies that consider different manufacturing applications}
\end{table}

\subsection{Workflow Evaluation}
Workflow evaluation refers to characterization of dynamic production systems by computing process-related metrics or Key Performace Indicators (KPIs) for implementing situation-aware production control~\cite{gyulai_analysis_2020}. In the era of cyber-physical environments, state-of-the-art tracking  systems monitor, evaluate and control production in smarter ways than ever before. \citet{arkan_evaluating_2013} utilize spatio-temporal data collected by applying Real Time Locating System (RTLS) from a multi-item production system with the aim of improving work-in-process (WIP) visibility within manufacturing. Specifically, they focus on assessing the performance of a semi-automated shop floor for producing passenger car plastic bumpers and spoilers for a manufacturing company. They first applied a filtering method to exclude redundant RTLS data instances from the trajectories of objects/items moving between workstations. The filtering method divides the floor into a set of zones and for each product only keeps data instances when it enters or exits a zone. This helps to significantly reduce the size of trajectories and saves time during analysis, since the multiple data instances when waiting in a zone for a while are ignored. The cleaned trajectories are then used to compute a set of KPIs: cycle time, cycle speed, production time, defect reject ratio, work space utilization, for evaluating the workflow performance. These KPIs are then analyzed to redesign the floor with a simulation tool. 

To develop novel analytics solutions for improving production control and management
process, the correct use of RTLS data is of utmost importance. Consequently,
\citet{gyulai_analysis_2020} present a spatial processing method to clean trajectory data
for the purpose of computing different KPIs (similar to ~\cite{arkan_evaluating_2013}) to
evaluate the performance of production systems. The method applies a discrete event
simulator 
model using Siemens Tecnomatix Plant simulation to create a test bed reflecting the operation of an assembly system consisting of four lines each with fifteen workstations. The simulated trajectory data are cleaned in two stages: noise filtration and mapping trajectories to production route. Firstly, a Savitzky-Golay filter has been applied to spatial data to remove the noise and increase the precision of the data without distorting the signal tendency. Secondly, smoothed data is mapped onto one of the production routes and further corrected using a probabilistic correction method. 

\citet{racz-szabo_real-time_2020-1} study the feasibility of RTLS to support different applications in manufacturing including  production control, quality control, safety, and efficiency monitoring, etc. They present a case study of using RTLS data from an automotive company to: i) identify the bottlenecks in defined production zones, ii) measure the cycle time deviation at the workstations. They also provide a guideline for implementing RTLS based tracking systems for the above mentioned applications. As such, they systematically explain the data cleaning and analysis method involved as part of the workflow. To identify the bottlenecks (in terms of temporary storage or unplanned workstations in the production process), they cluster the trajectory data by applying the k-means algorithm. The cycle time of workstations is measured based on classified zone data that are visualized later to provide real-time information on the status of the production process.

\subsection{Collision Avoidance}
In a dynamic manufacturing environment, different objects such as human workers, robots, and Automated Guided Vehicles (AGV) often work side-by-side. The collaborations among different objects contribute to improve the flexibility and intelligence of automation. To facilitate a safe and effective working environment, it is necessary to predict the future whereabouts of a large number of users in indoor spaces. However, the movement patterns of these objects are stochastic and time-varying in nature. As such, it is quite challenging for the objects to efficiently and accurately identify task plans of others and respond in a safe manner. \citet{locklin_trajectory_2020} considers the task of predicting future positions of human workers with the aid of RTLS data, which can be considered as the general problem of trajectory prediction. Motivated by the law of momentum, they present a method which assumes that the workers cannot change their speed and direction infinitely fast. Consequently, they apply least square fitting of a
second-degree polynomial function to compute the future speed $V_{t+1}$ of the workers using a number of past positions $L_{history}$. The current positions data ($P_t$) and estimated future speed ($V_{t+1}$) are then used to compute the future position ($P_{t+1}$) at next time step as shown in Eq. \ref{eq: trajectory_LR} 

\begin{equation}
\label{eq: trajectory_LR}
P_{t+1} = V_{t+1} \cdot  t+P_t
\end{equation}    
Moreover, to enable human-robot collaboration, 
the robots require various capabilities ranging from fundamental skills such as activity recognition of human co-workers to high level skills including reasoning about intentions
and collaboration in a shared space.  \citet{cheng_towards_2020} develop a unified framework for safe and effective collaboration between agents (human and robots). The framework consists of two main components: human trajectory prediction and plan recognition. Human trajectory prediction aims to predict continuous movement of human activities for safe robot trajectory planning. On the other hand, plan recognition is to infer the correct plan in the human’s mind to help adapt their actions to the human’s work plan. LSTM recurrent networks have been used to model the dynamics and dependencies in sequential movement data and consequently predict the human's next activity. The inputs to the LSTM networks include wrist positions and velocities of selected key points of human fingers. Given the classified motion labels and a history of human pose, the potential plans of human workers have been inferred based on Bayesian inference methods. 

\citet{zhang_recurrent_2020} also present a deep learning method to predict the future
motion trajectory of human operators in a human-robot collaborative car engine assembly
task. The method uses the visual observations of human actions (in terms of $x$, $y$, $z$
coordinates of five parts and four coordination units from the human body) as inputs to
the LSTM model which predicts the next move of a human operator to enable a robot's action
planning and execution. The applied LSTM model includes two types of functional units into
the recurrent structure to parse the evolutionary motion pattern of human body parts as
well as their coordination for improved prediction accuracy. To reduce the
uncertainty-induced robot mis-trigger and enhance the  reliability in interpreting the
human motion, they also apply a probabilistic inference based on Monte-Carlo
dropout. Additionally,~\citet{wang_location_2021} develop a similarity based model for
location prediction by incorporating both spatial and semantics aspects in the indoor
scenario. They represent the trajectory data by the sequence of (access point, time) pairs
where each access point is represented by a unique ID and sub-category of regions it
covers. The developed model applies k-nearest
neighbors 
to the find a trajectory ($T_s$) from the database that is most similar to a given
trajectory $T_y$ based on a distance metric and then predict the next location of $T_y$
from $T_s$. The main novelty lies in the formulation of a distance metric that considers
both spatial and contextual/semantic distance computed based on longest common sub
sequences 
and dynamic time warping, 
respectively. The method is evaluated using a large trajectory data set: 67 access points, 200 defined regions,  34 sub-regions and 261,369 trajectories. Evaluation shows its superior performance over the hidden Markov model (HMM) model.

\subsection{Frequent Path or Trajectory Patterns}
In recent years, ML and data mining based analytic approaches have been used for mining common and frequent patterns from trajectory data collected from manufacturing objects as a collaborative community. \emph{Frequent path} or \emph{trajectory patterns} mining refers to finding groups of trajectories, considering their spatial or temporal similarity or both, for the purpose of traceability and transparency of production process, and to enable control and management of work in process (WIP). It also can help to task scheduling and detect abnormal condition during production planning and execution process. 

\citet{cai_mining_2017} develop a spatio-temporal data model for monitoring IoT enabled
production systems by mining frequent trajectory patterns of WIP. The data model first
maps physical trajectories of WIP into logical trajectories by utilizing the concept of
multi-modal data cubes that consider both spatial and temporal data characteristics to
describe the changing states of WIP throughout the manufacturing process. The logical
trajectories expressed in terms of sequence of data cubes can better represent the logical
features of the manufacturing systems. They also present a method, called a
\emph{process-based} method with a priori detection (PMP),  for mining logical frequent
trajectory patterns, i.e., identifying groups of trajectories according to their
similarity either in the temporal or spatial sense. The proposed PMP method combines the
principle of the Apriori algorithm and depth first search to find the frequent nodes and
subsequently identify the frequent logical trajectory patterns.  The performance of the
PMP method has been evaluated using both a synthetic data and real data set collected from
a manufacturing workshop in China. Evaluation shows that PMP outperforms depyth-first
search,  graph based mining, modified Apriori methods in terms of both accuracy and execution time. 

\citet{bu_data_2018} develops a framework describing: i) the possible applications of RFID technology for tracking objects (materials, robots) during the production process; and ii) a method for mining frequent path patterns from massive amount of tracking data. The applied data mining method basically applies the Apriori algorithm to correlate event time with trajectory data in order to identify the most frequent path patterns during both off-time and peak-time. The frequent path patterns could be useful for different purposes:  readjusting material flow paths, dispatch plans of AGV robots or increasing working efficiency. \citet{liu_mining_2012} study the application of stationary RFID tags for activity monitoring and present an analytical method for mining frequent trajectories of regular activities. In contrast to the traditional RFID based localization methods, the developed object localization method uses the interference on the stationary RF tag signals caused by the activities to detect the activities themselves or unauthorized objects. To identify the interference caused by moving objects, Chebyshev’s inequality is used on the sensitivity of the tags. This helps to map the raw RFID signal into binary time series indicating whether interference is identified at different periods. The mapped binary time series data is later used to mine the frequent trajectory patterns of regular activities and identify movement of anomalous objects. They also present an empirical evaluation of the frequent trajectory mining algorithm using a real data set under different scenarios: single activity, group activities, busy activities, etc. Evaluation shows that the algorithm is fault tolerant and can detect frequent trajectories well given the activities are not very complicated in space.


\subsection{Indoor Space Modeling and Event Detection}
Optimal utilization and customization of indoor spaces (such as shop floor, production floor, etc) are crucial for mass production, efficient utilization of resources, and reducing cycle time in manufacturing. Traditionally, qualitative methodologies such as long term observations, and interviews and questionnaire based surveys are applied to design  and understand the use of indoor spaces. However, the massive volume of RTLS data collected can help to track geo-spatial patterns of users and interactions among them in indoor spaces in a timely and more efficient manner. 

\citet{han_data_2014} aim to quantify the distribution of indoor space utilization patterns over time and predict the future regions of interest. Specifically, they develop a trajectory clustering method to model the indoor space utilization by considering common trajectory movement patterns from multiple users.  They apply a partition-and-group  approach for identifying and grouping sub-trajectories from a trajectory database. From each trajectory, they first identify a set of characteristics points by applying a method called TRACLUS~\cite{lee_trajectory_2007}  which uses the Minimum  Description Length (MDL) principle and subsequently partitions the trajectory into a set of line segments by joining those points. The segments represent the intra-trajectory movement patterns. The segments from all trajectories are then grouped into a set of clusters based on a density based clustering algorithm for spatial data (GDBSCAN~\cite{sander_density-based_1998}). The clustering results help with recognizing the regions that are expected to be utilized heavily and visualize the evolution of utilization over time for the better design of indoor spaces in the future. The method has been evaluated using a case study at the College of Engineering, Penn State. The case study considers modelling of indoor space utilization characteristic using the trajectory data set collected from a student-oriented learning and design facility and subsequently optimising the over/underutilized regions of the design space by quantifying the interactions between users and objects. 

\citet{flossdorf_unsupervised_2021} consider the task of event detection from the trajectory data of objects moving in a production floor at a manufacturing company. Specifically, they aim to identify whether incoming location signals emitted by sensors attached to the smarts objects refer to actual movement event (AME) or undesired awakening event (UAE). For this classification task, it presents two different unsupervised algorithms. The first algorithm relies on the principle of DBSCAN clustering method. For each event, it observes the position information of the past $k$ events and determines the
maximal distance which exists to one of those. An event is then classified as AME if the maximum distanced from past $k$ events is higher than a predefined threshold $r$. The second algorithm considers a time-based criterion. For each event, the passed time between its occurrence and the occurrence of the event which was the $k^\text{th}$-last observation is calculated. If this time difference is below a certain threshold $b$, the point is labeled as AME. The parameters ($k$ and $r$ for first algorithm, and $k$ and $b$ for second one) are determined based on visual analysis of their distributions. The effectiveness of both algorithms has been evaluated using a real data set consists of $(x,y,z)$ coordinates of 401 sensors attached with manufacturing objects for two months, with 3.5 million positions in total. Although the results show that both algorithms achieve similar classification accuracy, the former one performs better in the presence of noise in data. 

\subsection{Non-Trajectory Based Data}

The applications of IoT have led to a data-rich manufacturing environment. This is having a positive impact on decision making and monitoring. However, the data generated by IoT enabled smart objects is unstructured, and expected to grow exponentially. Additionally, manufacturing data obtained from such smart objects or sensors does not characterize movement or trajectories in many cases (e.g. ~\cite{syafrudin_performance_2018, tao_data-driven_2018},~\cite{zhang_framework_2017}). Below we review the prominent studies utilizing non-trajectory based data. 

Big data analytics have great potential for processing massive volumes of manufacturing
data and developing applications for (but not limited to) fault detection, quality
prediction and defect classification. \citet{syafrudin_performance_2018} present a system
for monitoring the production line of automotive manufacturing by combining IoT enabled
sensors, big data processing, and a ML based fault detection model. The system first
utilizes IoT based sensors to collect the real-time temperature, humidity, accelerometer,
and gyroscope data from an automotive production line. The unstructured and large volume
of data relating to manufacturing process is then stored and processed using big data
technologies that include Apache Kafka, Apache Storm, and MondoDB. Subsequently, it
applies a hybrid ML model utilizing the concept of both supervised and unsupervised
learning. Specifically, the hybrid ML model employs DBSCAN clustering to identity noise in
the data and a random forest
algorithm for identifying anomalous activity or fault detection in the manufacturing
process given the current sensor data from the production line. The proposed system has
been evaluated using a real data set from an automotive company in Korea. Experimental
results indicate that the presented system is scalable and efficient to process the large
volume of sensor data, and reduces both CPU and memory utilization.The fault detection
system is evaluated using 342 instances, each consists of eight features. Results showed
that random forest
can detect fault/anomalous activities  with better accuracy compared to the other models
tested that include na\"ive Bayes, logistic regression, and neural networks (NNs). 

In a similar study, \citet{tao_data-driven_2018} discuss the role of big data analytics at different stages of the data life cycle such as data collection, transmission, storage, pre-processing, filtering, analysis, mining, visualization, and applications in supporting smart manufacturing. They also presented a case study focusing on fault diagnosis and prediction by applying NNs utilizing vibration data of machines as inputs. \citet{zhang_framework_2017} also proposed a big data analytics framework for optimization and management of product life cycle. Their framework includes four components: data sensing and acquisition, data processing and storage, data mining, and applications for product life cycle management. The most important data mining module is designed to discover the hidden patterns and knowledge from historical and real-time data by utilizing clustering techniques and association rule mining. 

Moreover, real-time scheduling and revised-scheduling of the shop floor is one of the key
factors for quality control and fast delivery of products in modern manufacturing
industry. Flexible and adaptable scheduling enables rearrangement of tasks should there be
any unexpected events or faults.  \citet{ji_big_2017} propose a big data analytics based
approach for predicting errors or potential faults of planned tasks or WIP to support shop
floor scheduling. They represent the planned tasks and WIP using a set of data attributes
and compare them with the fault patterns mined from shop floor database. Based on this,
they compute the similarity or difference relative to the mined fault patterns and provide
a reference for potential faults that include machining errors, machine faults, and
maintenance states ahead of machining task and before actual faults during machining. To
minimize database query time, they used RapidMiner
\footnote{\url{https://rapidminer.com/}}, an integrated open source software platform
which supports data processing, ML, deep learning, text mining, and predictive
analytics. \citet{zhong_mining_2014} consider computing standard operation times 
and discovering unknown dispatching rules from shop floor data for advanced production
planning and scheduling under different operational conditions. They develop a data mining
method which involves clustering of shop floor data using support vector machines.
The clustered data is then used to estimate standard operation times and mine job dispatching rules
based on a decision tree
model.

\subsection{Non-Position Based Movement Data}
\label{sec:applcations-non-positioning-data}
In the sections above, we emphasized the usage of positioning data. However, movement can be captured by \emph{non-}positioning data as well, e.g., by accelerometers. 
Miniaturized versions of accelerometers, gyroscopes and magnetometers are commonly
packaged as wearable sensors to capture movement patterns in human and animals. These
sensors generate signals that characterize different movement patterns. Note that these
sensors do not generate location data. In some outdoor scenarios, GPS devices are included
as part of the wearable sensor pack to provide location data, but accelerometers,
gyroscopes and magnetometers do not themselves generate any position information.

\citet{stetter_wearable_2021} summaries a good number of applications of such sensors in human space for strategic decision making. These sensors are are commonly used for:
\begin{itemize}
    \item Common daily activity, exercise levels
    \item Biomechanical parameters (e.g., bends)
    \item Understanding injury risks \cite{soter_analytics_soter_2020}
    \item Sport performance diagnosis
    \item Clinical human movement analytics
    \item Movement abnormalities or identifying changes due to orthopedic or physiotherapeutic interventions
    \item Gait analysis
    \item Patient healing progress
    \item Ambulatory monitoring methods for applications to neurological disorders
    \item Measuring worker productivity: (e.g., fruit picker efficiency)~\cite{dabrowski_sensor_2019}
\end{itemize}

The accelerometer sensors are also used to understand movement patterns in animals. Following are some common applications of such sensors in the livestock industry:
\begin{itemize}
    \item Behavior analysis (e.g. walking, lying, standing etc. in animals)~\cite{rahman_cattle_2018}
    \item Health monitoring (e.g. birthing events, estrus etc. in animals)~\cite{shahriar_detecting_2016}\cite{smith_automatic_2020}
    \item Understanding group behavior (e.g. animal group foraging)
\end{itemize}

Notice that positioning-based and non-positioning based analytics are not disjoint
dimensions.  For instance, one can consider combining position data with first derivative
(velocity) and second derivative (acceleration) to characterize or enhance position based
trajectories. 

\begin{summary}{Analytics on position based movement data}

The above review suggests that several approaches have been investigated in the literature
for efficiency monitoring, production control, safety and collaboration. Most of the
reviewed studies consider using standard ML or data mining methods with only a few
exceptions: for example, \cite{arkan_evaluating_2013, gyulai_analysis_2020} compute KPIs
for evaluating workflow efficiency from operational data without ML involvement, and
\cite{liu_mining_2012} presents an analytical method for processing raw trajectory
data. However, the literature supporting the utilization of RTLS data based on ML is too
shallow -- there are few studies available and the applied ML techniques are
simplistic. The applied traditional ML models have limitations to fit the specific
characteristics of position-based data such as the presence of dependencies among
measurements induced by the spatial and temporal dimensions. For example, standard
classification models (e.g. random forests, neural networks, decision trees) cannot characterize the non-uniform and sequential patterns commonly associated with trajectory data, and classical k-means algorithm does not consider the spatio-temporal relation while grouping the trajectories into different clusters. Thus, advanced and specialized models can be appropriate choice to develop applications utilizing RTLS data. These include sequence-to-sequence models that require to predict an output sequence (e.g. complete future trajectories of objects moving in a dynamic environment), and TCNN  for classification of trajectories by considering similarities among them both in temporal and spatial senses.       

Different application possibilities exist in manufacturing using RTLS data. Taking the knowledge from extensive studies in other domains, the literature can be extended by utilizing RTLS data in manufacturing by following ways:
\begin{enumerate}[label=\roman*]
     \item \textbf{Anomaly detection:} Anomaly (also known as outlier) identification approaches are an active area of research across domains and involve the usage of data in various formats including sequence or trajectory data. Anomaly identification also can be studied either in the context of individual outliers or collective outliers. Most of the existing approaches solely focus on identification of simple basic outliers~\cite{belhadi_hybrid_2021}. However, outliers in manufacturing data are likely to exist in a group when there is a group of objects (e.g., workers, robots, or other smart objects) that deviates from the anticipated and usual trajectory in a given time due bottlenecks in the production systems. Moreover, defining the abnormality of movement behavior and detecting anomalies from complex and large trajectories is an inefficient approach since the models can be overloaded with the dramatic increase of trajectory streams generated by multiple interacting objects. Hence, features reflecting the spatial, sequential, and behavioral characteristics of the objects can be identified from the long trajectory streams and used with the ML models as an efficient alternative (e.g. ~\cite{lei_framework_2016, yu_feature-oriented_2021}). 
     
    \item \textbf{Trajectory prediction:} In the context of manufacturing, trajectory prediction has considered predicting only next position of workers or objects~\cite{locklin_trajectory_2020}. However, prediction of a complete route or trajectory from a set of past position sequences could be more useful to plan and monitor the WIP, collision avoidance, and enhance collaborations among workers and/or smart objects. 
    
    \item \textbf{Trajectory clustering:} Clustering is a widely used method for discovering interesting or unexpected patterns in trajectory data. Previous studies on trajectory clustering apply different algorithms that compare and group trajectories as a whole~\cite{lee_trajectory_2007}. However, in reality each trajectory may have a long and complicated path and moving objects move rarely together for entire path. Besides, common sub-trajectories is also useful in many manufacturing applications, especially when there are regions of special interest for analysis~\cite{lee_trajectory_2007, wang_big_2020}. Hence, approaches that partition each trajectory into characteristic segments and then group the segments of all trajectories can help to better discover common patterns from a trajectory data set.

    
    \item \textbf{Trajectory representation:} Trajectory data is expected to be large since RTLS record the positions in very short intervals of time and real-time processing of such data sets is quite challenging. Hence, existing studies consider dimensionality reduction of trajectories by mapping raw trajectory data onto a layout (which is know beforehand) based representation (e.g.,~\cite{arkan_evaluating_2013, gyulai_analysis_2020}), or using grid based indexing of trajectory data (e.g.~\cite{ding_anomaly_2018}. Although this mapping process is easy to implement, it is not feasible to apply if the exact layout is not available. This problem can be addressed in two ways. Firstly, by representing and extracting features from trajectories by the use of generative models where the behavior of the each trajectory has been approximated by a parametric model~\cite{atluri_spatio-temporal_2018}. The learned parameters or features can then be used as succinct representations of the trajectories. Secondly, mapping trajectories based on the distance to a set of landmarks points chosen arbitrarily or placed randomly to cover a domain of focus ~\cite{phillips_simple_2019}. New distance measures that easily and interpretably map objects can be computed based on how they interact with the set of landmarks. These distance measures subsequently can be used effortlessly with well established ML models for trajectory classification, anomaly detection, clustering, etc. Alternative methods can also be explored for trajectory representation that include semantic mapping based on classical logic and knowledge representation techniques.

\end{enumerate}


\end{summary}

\section{Industrial Applications, Commercial Systems and Digital Twins}
\label{sec:industrial-applications-and-commercial-systems}
In the previous section we took a broad view on movement analytics applications for collision avoidance, finding trajectory patterns, event detection, workflow evaluation, and health, among others. We viewed these from a technology angle, in particular the ML approaches underlying most of them. 

In this section, we take a different viewpoint from the end-user perspective: what are the industrial real-world applications that are currently supported by movement data analytics, and to what benefit (Section~\ref{sec:industrial-applications}), and what are the available commercial systems (Section~\ref{sec:commercial-systems})?
Finally, we investigate on the opportunity to employ digital twin technology (Section~\ref{sec:digital-twins}).

\subsection{Industrial Applications}
\label{sec:industrial-applications}
Table \ref{tab:ref_appliction_and_use_of_ma} lists industrial applications in maritime industries~\cite{yang_how_2019, du_optimized_2021}, transportation~\cite{al_nuaimi_survey_2011, arp_dynamic_2020, andrienko_visual_2007-1, mo_calibrating_2021, prato_route_2009, asakura_tracking_2004}, autonomous vehicles~\cite{yang_obstacle_2022, song_intelligent_2021}, health~\cite{al_nuaimi_survey_2011, nguyen_improved_2006, andrienko_visual_2007-1}, behavior analysis~\cite{prato_route_2009, asakura_tracking_2004}, manufacturing~\cite{schabus_geographic_2015-1, andrienko_visual_2007-1, dobler_supporting_2020} and indoor positioning~\cite{schabus_geographic_2015-1, al_nuaimi_survey_2011}.

\newcommand{\tcomment}[1]{\multicolumn{3}{@{}>{\raggedright}p{30em}}{\small\emph{Comment: }#1}}
\newcommand{\tspace}{\rule{0pt}{12pt}}

\begin{xltabular}{\linewidth}{@{}l@{~~}>{\raggedright}p{8em}@{~~}>{\raggedright}p{8em}>{\raggedright}X@{}}
  \textbf{\small Paper} &\textbf{\small Category}&\textbf{\small Technique} &
  \textbf{\small  Applications}  \\\hline
\tspace  \cite{schabus_geographic_2015-1} 
  & Indoor movement analysis
  & GIS
  & Manufacturing
  \\
  \hline
\tspace  \cite{al_nuaimi_survey_2011}
  & Survey on indoor positioning
  &
  & Transportation, manufacturing, logistics, safety, health industry
\\  \hline
\tspace  \cite{yang_how_2019} 
& Review of analytics and big data applications
& AIS
& Maritime
\\
  & \tcomment{The survey covers techniques that are relevant to manufacturing as well. For example: Behavior analysis, environmental impact, performance, trading.} \\
  \hline
\tspace\cite{nguyen_improved_2006} 
& Health and Safety
& Bayesian neural network  
& Wheel chair movement
\\
& \tcomment{The technique mentioned can be used for vehicles moving in manufacturing space} \\
\hline
\tspace\cite{arp_dynamic_2020}
& Route estimation, collision avoidance
& Parametric optimization
& Transportation
\\
& \tcomment{The technique mentioned can be used for automated vehicles, robots and worker movements in manufacturing space} \\
\hline
\tspace\cite{andrienko_visual_2007-1}
& Visual movement analytics
& Visual analytic tool (VAT)
\\
& \tcomment{Transportation, behavior analysis, logistic optimization, manufacturing} \\
\hline
\tspace\cite{mo_calibrating_2021}
& Transportation, capacity management
& Transit network flow simulation
& Transportation
\\
& \tcomment{The paper covers problems that also exists in manufacturing at different level} \\
\hline
\tspace\cite{yang_obstacle_2022}
& Collision avoidance
& Optimization and RRT
& Autonomous flight
\\
& \tcomment{Methods can be used in Manufacturing for collision avoidance. There is a scope to improve the solution of the other methods such as logic inferences and Bayesian networks or ML for more accurate position estimation.  } \\
\hline
\tspace\cite{du_optimized_2021}
& Maritime, collision avoidance
& DDPG and DP
& Transportation
\\
& \tcomment{Collision avoidance and path planning with minimum turns is also a challenge in manufacturing space.  } \\
\hline
\tspace\cite{prato_route_2009} 
& Review of route choice methods and impact of behavior
& 
& Transportation
\\
& \tcomment{Routing is a problem in manufacturing as well} \\
\hline
\tspace\cite{asakura_tracking_2004}
& Behavioral analysis
& Threshold based labeling algorithm
& Transportation management and demand planning
\\
& \tcomment{Tracking worker movements is important in manufacturing as well} \\
\hline
\tspace\cite{song_intelligent_2021}
& Collision avoidance of mobile robot
& ML+ random tree + heuristics
& Manufacturing, transportation\\
\hline
\tspace\cite{dobler_supporting_2020}
& Manufacturing
& Optimization (static) + situational awareness and rescheduling
& Manufacturing
\\\hline
\caption{\tspace Big data and movement analytics across various industries.}
   \label{tab:ref_appliction_and_use_of_ma}\\
\end{xltabular}

The papers~\cite{yang_how_2019, al_nuaimi_survey_2011, asakura_tracking_2004} give a good review of how movement and data analytics are used to assist real-time decision making in the areas of maritime industries, indoor positioning, behavior analysis, and transportation, respectively.

\citet{al_nuaimi_survey_2011} presents a survey of systems and methods used for tracking
people and objects in indoor environments such as factories, hospitals, nursing homes, and
train terminals. The authors have compared different indoor positioning systems such as ``Fixed indoor positioning systems'' (that have a fixed number
of Base Stations (BS) installed at fixed locations within the
building); and ``Indoor pedestrian positioning''(people carrying localization sensors). The comparison was done with respect to the challenges of providing the best indoor position systems. The key factor in deciding tracking system efficiency depends on how accurately it can track the people movement and which methods are available to quantify the movement. With this perspective, the authors have extended the survey on different methods used for estimating the position of people. The two main methods outlined in the survey are:  ``Bayesian filtering'' to estimate the steps of the pedestrian at a certain time when knowing the previous steps of the same pedestrian at number of times before it and ``Kalman filter-based algorithms'': a mathematical model which is used to accurately estimate the position with the existence of noise. Their findings suggest that the fixed indoor position system provides a good accuracy, however there is good scope to enhance the performance of  the ``indoor pedestrian positioning'' system.

Later  \citet{schabus_geographic_2015-1}, also focused on tracking objects and people in
indoor environments. However, rather than deciding on which system is better as in \cite{al_nuaimi_survey_2011}, the authors focused on the objective of finding the best paths from one point to another
and identifying the bottlenecks. For this purpose they analyzed the movement behavior in an
indoor environment using Geographical Information Services (GISs). The movements behavior
is visualized as a network with paths created from one point to another using routing
algorithms to get shortest paths. These paths are compared with historical paths (actually
visited path by an asset) to gain insight about detailed movement behavior and deviations
from the optimal path. The bottlenecks are identified by summing up the number of times an
edge is visited by an asset.

While the focus of \citet{schabus_geographic_2015-1} was to identify the deviation from
shortest paths and bottleneck paths with using movement behavior analysis,
\citet{song_intelligent_2021} worked with the objective of finding safe paths in an indoor
environment. They proposed mobile robot path planning  to produce the optimal safe path. A
real-time obstacle avoidance decision model based on ML algorithms is
designed to improve the accuracy and speed of real-time obstacle avoidance prediction for
mobile robots in local path planning. The Rapid-exploration Random Tree algorithm (RRT)
algorithm is extended by greedy algorithm approach to smooth the global path and shorten
its total length by removing the redundant nodes. The method is called Smooth Rapidly
exploring Random Tree (S-RRT) method. The authors have also looked at optimizing the path
planning time together with finding the shortest path. For optimizing the path planning
time and generate a more stable collision-free optimization path, an improved hybrid
genetic algorithm-ant colony optimization 
algorithm is proposed. It is based on the idea of hybrid
algorithm that combines the advantages of genetic algorithm and ant colony optimization, and
can generate better paths in global path planning and local path planning. 

All three papers mentioned so far used movement analytics to find best and/or safe paths in indoor
environment. However, there are studies demonstrating the benefits of using the movement analytics with respect to other aspects such as capturing real time disruptions in manufacturing space. For example,  \citet{dobler_supporting_2020} studied how big data from interwoven, autonomous and intelligent supply chains can be integrated and used in optimizing the manufacturing systems against various real-time disruptions such as machine failure, resource become unavailable etc. They propose and compare two different approaches for the optimization of manufacturing lines. The first approach is based on static optimization of production demand. In the second approach, real-time situational awareness---implemented as digital avatar---is used to assign local intelligence to jobs and raw materials. The real-time situational awareness will inform about events such as disruptions and machine failures which will be fed into the optimization framework to make better decisions. The results are generated using event-discrete simulation and are compared to common (heuristic) job scheduling algorithms. 

The application of movement analysis is studied in literature from
various point of view in manufacturing as well as other industries.
Another industry making use of movement analytics is the transport industry where there are many applications such as identifying consumer behavior that in turn determines the
demand of a specific route; collision free paths; route congestion; demand of different
paths at different times; and estimating alternate routes   \cite{asakura_tracking_2004,
  prato_route_2009, arp_dynamic_2020, mo_calibrating_2021}. To identify consumer
behavior in using specific routes, correct labeling of events plays an important role.
\citet{asakura_tracking_2004} worked on this problem and propose a labeling algorithm for
tracking travelers' behavior. They used mobile communication systems such as GPS (global
positioning systems), cellular phone and RFID  for the
same purpose. Two of the important events to label that helps in identifying the demand of a
specific path/position of route and hence help in identifying the possible congestion are:
move or stay. The authors have proposed a simple labeling algorithm based on the approach
of varying thresholds or discarding some points from the analysis, for example points that
have very short duration at a place are discarded while labeling stay. The threshold variation approach helped them in predicting the areas of congestion or demand at a time with more confidence.

\citet{andrienko_visual_2007-1} worked on the similar problem; however rather than just
labeling the data efficiently, they suggest a complete framework for movement analysis
combining interactive visual displays with database operations and computational
methods. Their
movement analysis framework has three main components: 1). Data cleaning and filtering by
including additional fields such as speed or time interval between two movements that can
help with logically filtering the data; 2). Extraction of significant places where an object
stops frequently or for more duration using a SQL query or user defined criteria;
3). Extraction and examination of trips  where a trip may be application-and
goal-dependent using a SQL based query or thresholds. To avoid the false
flagging/categorizing of a place as significant place, for example, removing occasional
places from regular ones, the clustering algorithm (OPTICS) is used where parameters of a
cluster are user configurable.

While \citet{asakura_tracking_2004}  and \citet{andrienko_visual_2007-1} worked on data
handling and labeling part, \citet{mo_calibrating_2021} studied the use of movement data
to predict the demand of flow through the rail network. They propose a simulation-based
optimization (SBO) framework to simultaneously calibrate origin-destination (OD)  flow,
passenger path choices and train capacity for urban rail systems using automated fare
collection and automated vehicle location data to analyze performance and
conduct performance retrospectives of urban rail systems. The SBO model has the objective
of minimizing the square error between model-derived OD exit flows and the corresponding
observations and the difference between model-derived and observed  journey time
distribution (JTD) where observation data is obtained from autoamted fare collection data. The model calculates
the OD exit flow, and JTD using a black-box function that corresponds to the transit
network loading (TNL) model (a forecasting model), which  assigns passengers over a
transit network given the (dynamic) OD entry demand and path choices. TNL can output the
model-derived OD exit flows and JTD for a given set of path choices and train
capacity. TNL has no analytical form therefore SBO is used. The proposed optimization
method is tested on different scenarios representing different degrees of path choice
randomness and crowding sensitivity. Data from the Hong Kong Mass Transit Railway 
system is used as a case study for generating synthetic observations used as ``ground
truth". The results show that the response surface methods (particularly constrained
optimization using response surfaces) have consistently good performance under all
scenarios.

Later on \citet{arp_dynamic_2020} worked on the problem of determining best
alternative routes in case of congestion or diversions using movement data. They use
parametric optimization and network based real-time forecasting for traffic flow en route on a
network.

\citet{prato_route_2009} present a good review of the state of the art in the analysis of route choice behavior within a discrete choice modeling framework. This review focuses on drivers' route choice behavior in transportation networks, but the same modeling framework is applicable to the multi-modal context. The review examines both major challenges in route choice modeling, namely the generation of a choice set of alternative routes, and the estimation of discrete choice models. This is difficult as the semantics identifying the different routes can be a challenge based on the length and topology of area and noise present in the data. The  choice set generation methods are classified as deterministic shortest path-based methods,  stochastic shortest path-based techniques, constrained enumeration algorithms that rely on the behavioral assumption that travelers choose routes according to behavioral rules other than the minimum cost path, and probabilistic approaches that attach a generation probability to each route.

The maritime industry is another industry also benefited from the use of movement analytics
~\cite{yang_how_2019, du_optimized_2021}. \citet{yang_how_2019} focus on the comprehensive
review of the literature regarding Automatic Information Systems (AIS) applications in
maritime industries.  They categorized the AIS applications into seven application fields
of data analysis for maritime industries: AIS data mining, navigation safety, ship
behavior analysis, environmental evaluation, trade analysis, ship and port performance,
and Arctic shipping. The methodologies in the literature are categorized into four
categories: data processing and mining, index measurement, causality analysis, and
operational research. One of the most recent and relevant works in the maritime industry
using movement analytics is by \citet{du_optimized_2021}. They worked on the problem of
identifying the obstacles and optimize against those obstacles to determine the
collision-free path for coastal ships with minimum turning points. They  present an
optimized path planning method based on improved Deep Deterministic Policy Gradient (DDPG)
and Douglas Peucker (DP) algorithms. The DDPG method works for continuous space and partly
depends on the grid environment and grid partition strategy. The aim of the study is  to
avoid known obstacles and shore-based information obstacles (such as ship-wreck area,
restricted navigation area, and military exercise area) by making better predictions and
finding an obstacle-free path with minimum turning points. To make better predictions,
authors have used Long Short Term Memory (LSTM)  as the first layer of DDPG to enable the uses of historical state information to approximate the current environmental state information. The DDPG is further improved by using a two stage reward function,  mainline reward function and auxiliary reward function, to overcome the low learning efficiency and convergence speed of the traditional DDPG method. The mainline reward function is used to guide the ship to reach the target point and complete the path planning task. Meanwhile, the auxiliary function gives reasonable punishment in the process of path planning, so as to avoid obstacles. The problem that too many turning points may exist in the above-planned path, which may increase the navigation risk, an improved DP algorithm is proposed to further optimize the planned path to make the final path more safe and economical. The proposed DP algorithm helps in removing the excess turning points.


The other important industry which benefited from movement analytics is health. \citet{nguyen_improved_2006, andrienko_visual_2007-1} study the use of Bayesian neural networks in developing a hands-free wheelchair control system. The experimental results show that with the optimized architecture, classification Bayesian neural networks can detect head commands of wheelchair users accurately irrespective to their level of injury


\begin{summary}{Industrial applications}
Big data and movement analytics played a crucial role in different industries to assist real-time data-informed decision making to mitigate the effects of possible disruptions. The application areas in the papers reviewed in this section cover path optimization to avoid collisions with minimum turns and distance travel; layout optimization to increase the efficiency of resources and throughput of the system; collision avoidance for systems working with automated vehicles and robots; demand and inventory management through consumer behavior; and flow management to avoid congestion (especially in the transportation industry). For tracking data, various techniques are used based on the application area and suitability. For example, GIS;  infrared, ultrasonic, radio frequency systems; and indoor pedestrian positioning are used to track data in an indoor environment. In contrast, GPS, cellular phone, and RFID systems are used for tracking objects in an outdoor environment. Once the data is obtained, mainly ML, random tree, neural network, Bayesian probability, and situational awareness (simulation) based techniques are used to process the data and make accurate predictions about the obstacles and issues. To act against the informed obstacles and issues, heuristics and simulation-based optimization techniques are used for real-time decision-making. 

In summary, different industries are benefiting from the use of movement and data analysis in predicting relevant events and making efficient decisions based on those predicted events.  There are good review papers for the application of movement and data analysis in other industries~\cite{yang_how_2019, al_nuaimi_survey_2011, asakura_tracking_2004}, but there is no existing review paper for movement analytic and manufacturing.  Methods ranging from ML, neural networks, Bayesian probability, and logical inference have been used in the literature for data  and movement analytics, but none of the mentioned papers have leveraged the advantage by hybridizing different techniques. There is a clear scope in combining the different techniques used across various industries and leveraging their benefit to inform decision making in manufacturing. 
\end{summary}

\subsection{Commercial Systems}
\label{sec:commercial-systems}

There are a number of products on the market that provide movement analytics in manufacturing. In this section, we describe some of the commercial off-the-shelf products available. Details of individual products can be found in Table \ref{tab:cots-summary}.
\begin{table}[htbp]
    \centering
    \begin{tabular}{ p{0.15\linewidth}  >{\raggedright}p{0.22\linewidth}  >{\raggedright}p{0.22\linewidth}  >{\raggedright}p{0.27\linewidth} }
    \hline
    \textbf{Company} & \textbf{Hardware} & \textbf{Software} & \textbf{What can it do?} \\ \hline
    Data Dog~\cite{data_dog_monitor_2022} & None & Software for monitoring IoT devices. & Monitor performance of a system of IoT devices. \\ \hline
    iMonitor~\cite{imonitor_smart_2022} & None (QR codes which can be scanned by tablets) & Organizes lists of tasks and parts/inventory & Reduces paperwork by making data collection easy through scanning of parts using QR codes and tablets. \\ \hline
    Machine Metrics~\cite{machinemetrics_elminate_2022} & Box that connects to manufacturing machine via Ethernet cable. & Real-time visualization of equipment, system or worker on factory floor. & Visualization, bottleneck analysis and optimization of workflow by sending instructions to factory workers via app. \\ \hline
    Motion Analysis~\cite{motion_analysis_motion_2022} & Sensors for human motion tracking. & Motion capture software. & Ford used this to conduct an ergonomic analysis of their assembly line. \\ \hline
    SmartX HUB~\cite{smartx_hub_industrial_2022} & RFID tags to track parts, inventory and equipment. & IoT asset tracking software & Keeps track of parts and inventory, and monitors where and for how long items are being used. \\ \hline
    Worximity Technology~\cite{worximity_technology_oee_2022} & TileConnect, a wifi-enabled smart sensor. & Collects and sends data to the company for calculating the Overall Equipment Effectiveness (OEE) and displays data in real time.
        & Calculates OEE, displays key performance indicators in real-time on dashboards in factory, and tracks downtime to see when and for how long stoppages occur. \\
    \hline
    \end{tabular}
    \caption{Summary of commercial off-the-shelf products for movement analysis in manufacturing.}
    \label{tab:cots-summary}
\end{table}

Existing systems focus on collecting data to improve organization, real-time monitoring and improving factory performance and output. Tagging and scanning parts reduces paperwork makes record keeping easier~\cite{imonitor_smart_2022}. It also allows staff to know where inventory is located to prevent losing it~\cite{smartx_hub_industrial_2022}. Real-time data visualization~\cite{worximity_technology_oee_2022,machinemetrics_elminate_2022,data_dog_monitor_2022} allows companies to keep track of key performance indicators from dashboards in the factory. Finally, some products offer services for improving factory performance and output~\cite{machinemetrics_elminate_2022}, though the details of what optimization techniques they use are not clear from their web sites. The service provided by \citet{worximity_technology_oee_2022} includes a consultation with a Six Sigma expert who calculates the Overall Equipment Effectiveness (OEE) and will provide suggestions on how to improve it. 

An interesting project which took a different approach was that of \citet{motion_analysis_motion_2022}, who used cameras and sensors on workers to assess the ergonomics of their work stations. Rather than tracking the movement of parts through a factory, this approach was about understanding the body positions of humans to avoid repetitive strain injuries. 

The off-the-shelf products put an emphasis on ease of use and incorporation into the factory. This could indicate that when creating a product for industry, a focus needs to be put on ease of implementation, if the technology is to be taken up by industry. 

\begin{summary}{Commercial Systems}
In summary, there exist some products on the market to do movement analytics in manufacturing, and these products focus on improving organization, monitoring and factory performance. As these products are sold by private companies, the underlying analysis methods used in these systems are not clear. This is understandable, as companies need to protect their intellectual property, but it leaves a gap in the market for more transparent and explainable approaches. 
\end{summary}

\subsection{Digital Twins}
\label{sec:digital-twins}
A Digital Twin (DT) is a virtual representation of a physical system and its associated
environment and processes that is updated through the exchange of information between the
physical and virtual system. The concept of DTs was initially introduced in 2003 by
Michael Grieves, in an industry presentation concerning product lifecycle management.
In his white paper from 2014, \citet{grieves_digital_twin_2014} describes a DT as a three-dimension digital model consisting of: 1) Physical model: used for defining, describing, and extracting information about the physical entities present in the underline physical space; 2) Virtual model: which should be the exact mirror image of the physical model in virtual space; and 3) Connection model: ties physical and virtual space through transferring the historical and real-time information between the two. 

In DTs, online IoT sensors are used to collect data from the physical space, such as the
physical entities' status data and trajectories. The collected data
is then processed and filtered (to handle scalability issues) using AI/ML and other big data
techniques. The processed data is then sent to the virtual layer of the DT for an update.
Then, AI/ML, optimization, and simulation techniques are employed to identify potential
problems and to propose an on-time solution.
Clearly, the identification of the problem and its on-time solutions depends on
the frequency and correctness of the interaction between the physical and virtual space
through IoT sensors, and hence plays a crucial role in the success of DTs.

DTs have captured the attention of many researchers. The focus of research 
includes reviewing the definition and key parameters of DTs, and the evolution of their structure over time~\cite{tao_digital_2019,vanderhorn_digital_2021,glaessgen_digital_2012,kahlen_digital_2017,dahmen_what_2021};
how to improve DT implementations for better usability~\cite{damjanovic-behrendt_open_2019,dolgui_advances_2021,ruppert_integration_2020}; and
application of DTs to specific industries and objectives
\cite{damjanovic-behrendt_open_2019,dolgui_advances_2021,ruppert_integration_2020,
  macchi_exploring_2018,li_dynamic_2017,
  liu_role_2018,karve_digital_2020,guo_modular_2019,hauge_employing_2020}.



\subsubsection{Digital Twins Definitions, Structure, and Key Parameters}
With the development of industry 4.0 many authors have redefined DTs and extended their scope~\cite{vanderhorn_digital_2021, glaessgen_digital_2012, kahlen_digital_2017, tao_digital_2019, dahmen_what_2021}. They also reviewed the state-of-the-art DT research concerning the key components of DTs, the development of DTs, and the major DT applications in the industry. For example, \citet{glaessgen_digital_2012} redefined DT as ``the DT consists of a virtual representation of a production system that can run on different simulation disciplines that are characterized by the synchronization between the virtual and real system, thanks to sensed data and connected smart devices, mathematical models and real-time data elaboration. The topical role within Industry 4.0 manufacturing systems is to exploit these features to forecast and optimize the behavior of the production system at each life cycle phase in real-time.'' \citet{kritzinger_digital_2018}, gave the definition of DTs for manufacturing space, writing that ``a manufacturing DT offers an opportunity to simulate and optimize the production system, including its logistical aspects, and enables detailed visualization of the manufacturing process from single components up to the whole assembly.'' 

In 2019, \citet{tao_digital_2019} further contributed in definition and structure of the DT concept by extending the three-dimensional model for the DT and proposing that a complete DT should include five models: 1) Physical modeling: for extracting, defining, and describing the key features of a physical entity; 2) Virtual modeling: representing a mirror image of the physical world; 3) Data modeling: data definition, transmission, conversion, and storage; and 4) Service modeling: for identification, analysis, and upgrade services 5) Connection model: for maintaining a constant connection between the physical model, virtual model, data model, and service model. Based on the review, the author has identified the production, prognostics, and health management industries as the primary application industries for DTs. 

While \citet{tao_digital_2019} extended the model for DT, \citet{vanderhorn_digital_2021} has proposed a generalized definition of DT. They highlighted the key parameters required, such as: data update frequency; and level of abstraction of data, for the implementation of DT and how they can impact the performance of DT. According to them, these key parameters should be chosen based on the use case. For example, updating the data every minute may not be required. Similarly, it may not be required to consider all dimensions and details of the physical world while creating its virtual replica. However, the high level of data abstraction gives rise to the high accuracy of the physical model but may be pretty expensive and not practical. The impact of highlighted key components of the implementation of DT is demonstrated through a case study where a DT is developed to support the ongoing asset integrity management of a naval vessel. 

Recently many authors have reviewed the concept of DT \cite{kritzinger_digital_2018, cimino_review_2019, uhlenkamp_digital_2019, liu_review_2021, pronost_towards_2021, assad_neto_digital_2021, leng_digital_2021} from different perspectives. For example, \citet{cimino_review_2019} analyzed the status of DT research and the key technologies needed to apply DTs. \citet{uhlenkamp_digital_2019}, reviewed the concept and different areas of application of DTs to categorize the literature by identifying major distinguishing characteristics of the different approaches in the different application areas. \citet{liu_review_2021} conducts a comprehensive and in-depth review of the literature to analyze DT from the perspective of concepts, technologies, and industrial applications. \citet{pronost_towards_2021} did a literature review aiming at categorizing the objects defined under the term "Digital Twins" in the literature. \citet{assad_neto_digital_2021} summarizes a variety of features (such as digital, analytical, and timeliness) proposed by recent models in the literature to implement DTs.

\subsection{DT Implementation}



From the reviews mentioned above, it is clear that the implementation of DTs is a complex
task and is, therefore, getting the attention of many researchers
\cite{damjanovic-behrendt_open_2019, dolgui_advances_2021,
  ruppert_integration_2020}. \citet{damjanovic-behrendt_open_2019}  discuss an open-source
approach for implementing a DT demonstrator for Smart Manufacturing.
The DT demonstrator supports the supervision activity of the operator in monitoring how
the manufacturing system responds to production and environmental changes. As described by
the authors, one of the main potentials of using open source technology in Smart Manufacturing is to enhance inter-operation and reduce the capital costs of designing and implementing new manufacturing solutions. 
The authors described the major implementation requirements of DTs and Smart Cyber-Physical Systems (CPSs) for intelligent manufacturing systems where CPS is the integration of a virtual world that interacts with a physical world. CPS helps manufacturers to accelerate the design and improve inter-operation across actual life-cycle processes. 

Given the importance of CPS, \citet{dolgui_advances_2021} reviewed and identified the gaps in the architecture of CPS. The authors' main focus was to find a method for implementing a robust process control DT in a small or medium enterprise (SME) that ensures the needed functionality while being easy to understand, maintain and adjust. They proposed a specific implementation to fulfill a set of defined requirements. The proposed setup is tested on an industrial use case focusing on controlling two magnetic induction ovens that preheat aluminum extrusion billets. Their results demonstrate that a process's DT must be as specialized and customized as the system controlling it.  CBS integrates the simulation of DT models with real-time sensory and manufacturing data. Therefore real-time development and maintenance of simulation models play a crucial role. \citet{ruppert_integration_2020} focused on the problem of real-time development and maintenance of simulation models. The authors proposed a method that continuously updates the simulation models based on information provided by RTLS. 

\subsubsection{Digital Twins Application in Manufacturing}
The application of DTs in the manufacturing space and its benefits have been studied extensively. \citet{kritzinger_digital_2018} has reviewed the application and contribution of DTs in the manufacturing space. The authors have highlighted the following three main disciplines of production systems that can benefit from DTs to increase competitiveness, productivity, and efficiency. 
\begin{itemize}
    \item Production planning and control \cite{rosen_about_2015}: for example, orders planning based on statistical assumptions; improved decision support through detailed diagnosis; and automatic planning and execution of orders by the production units.
    \item Maintenance \cite{lee_recent_2013, susto_machine_2015, macchi_exploring_2018, liu_role_2018, karve_digital_2020}: for example: identify the impact of state changes on a production system; identification and evaluation of anticipatory maintenance measures; evaluation of machine conditions to achieve better predictions of the machine’s health condition.
    \item Layout planning \cite{uhlemann_digital_2017, guo_modular_2019, hauge_employing_2020}: for example: continuous production system evaluation and planning; identify hidden design flaws and proposing solutions within time.
\end{itemize}
\citet{leng_digital_2021} also reviewed the literature in the manufacturing space. They cover the DT review in manufacturing with the perspective of covering the available definitions, frameworks, major design steps, new blueprint models, key enabling technologies, design cases, and research directions of DT-based smart manufacturing system (SMS) design in this survey.

While \citet{kritzinger_digital_2018} and \citet{leng_digital_2021} present the literature review of DTs in the manufacturing space, there are studies covering other aspects of DTs such as model structure, implementation steps, requirements, and the use of available digital software to reduce cost and time of implementation of DTs in the manufacturing space.  For example, \citet{banica_stepping_2019} present a DT model for manufacturing based on the 5 dimension model \cite{tao_digital_2019}. Based on the importance and application of DTs, the authors have described three types of DTs:
\begin{itemize}
    \item Product Twin – A virtual prototype of a product used before starting its production line to analyze its behavior and make adjustments if needed. Thus, product twin decreases the costs of control and validation phases and can be used to improve the physical product's functional performance and quality.
	\item Process Twin – the next level is represented by the model for a virtual manufacturing process, which allows the company's management to make the best decision in terms of manufactured products, and operations to accomplish and test them. Process Twin could use Product Twin for each component of the manufacturing line, establishing its opportunity and efficiency.
	\item System Twin – this is the higher level, representing the virtual model of an entire system, based on Product Twin for each device and Process Twin to optimize the manufacturing processes of these components. 
\end{itemize}
The authors used the supply chain model as a case study to show how the available digital software can be used to implement DT. The three main elements of the DT model for the supply chain specified by the author are the real-time transmission of manufacturing updates, tracking and updating warehouse inventory, and controlling the distribution networks.
\citet{vachalek_digital_2017, zhuang_connotation_2021}  gave a detailed description of requirement, implementation and role of each step in DT for a production line in manufacturing space.


\subsubsection{Objective-Specific Implementation of DT}
While in the last section we outlined reviews that focused on area specific (supply chain, production line)  DT requirements and steps~\cite{banica_stepping_2019, vachalek_digital_2017, zhuang_connotation_2021}, here we summarise those that focused on objective specific (objective such as: maintenance activities, life cycle management, factory layout) implementation of DTs~\cite{macchi_exploring_2018, li_dynamic_2017, liu_role_2018, karve_digital_2020, guo_modular_2019, hauge_employing_2020}.

For example, \citet{macchi_exploring_2018} highlighted the role of DTs in supporting decision-making in asset lifecycle management. The authors have highlighted the use cases where a DT can help in asset management. \citet{li_dynamic_2017} also studied the use of DTs in life-cycle management; however, their work is focused on the maintenance of aircraft wing health. They proposed a DT model for the maintenance of aircraft wing health. To capture the various aleatory (random) and epistemic (lack of knowledge) uncertainty sources in crack growth prediction in an aircraft wing, the author proposed a  probabilistic model based on the concept of a dynamic Bayesian network (DBN) (see Section \ref{DBN}) for diagnosis and prognosis (a forecast of the likely outcome of a situation) to realize the DT vision. The DBN integrates physics models and uncertainty sources in crack growth predictions. In diagnosis, the DBN is utilized to track the evolution of the time-dependent variables and calibrate the time-independent variables; in prognosis, the DBN is used for probabilistic prediction of crack growth in the future. The author further enhances the DBN structure to make it economical in terms of time by avoiding Bayesian updating with load data. The proposed approach uses filters as the Bayesian inference algorithm for the DBN that enables the handling of discrete and continuous variables of various distribution types and non-linear relationships between nodes. The author has also addressed the challenge of implementing the particle filter in the DBN where 1) both dynamic and static nodes exist, and 2) a state variable may have parent nodes across two adjacent networks.

\citet{karve_digital_2020} also focused on the application of DTs in maintenance using Bayesian methods for quantifying uncertainty in diagnosis and damage prognosis. The authors considered the problem to predict, diagnose and optimize the repair/ maintenance planning, ensuring system safety for fatigue cracking under uncertainty using DT. The uncertainty can arise from system properties, operational parameters, loading and environment, noise in sensor data, and prediction models. The system safety against fatigue cracking is ensured by designing mission load profiles for the mechanical component such that the damage growth in the component is minimized while the component performs the desired work.

The use of DTs with the objective of optimizing and analyzing layout design is studied by \citet{guo_modular_2019}. They propose a layout design scheme and modular-based DT model for a factory design. The modular approach brings the flexibility to accommodate changes that may happen in each design stage of DT, hence saving workload and time for developing a new DT. The authors have explained the three main design stages as 1) Conceptual design: the first stage focused on designing the new factory's concept, including plant layout, capital investment, and throughput prediction; 2) Elaborate design: extension of conceptual design by including machine configuration, process design, production line or production unit configuration, material handling system configuration, and work shift configuration; 3) Finalized design: the final stage where the machine and logistics unit control strategy will be designed and the whole manufacturing system needs to be integrated. Because the virtual factory corresponds to the finalized design and is the most similar to the future physical factory, the DT has the most fidelity to the physical world in this stage. 

To demonstrate the proposed DT design, \citet{guo_modular_2019} used a paper cup factory in China as a case study. The company needs to expand its production and factory area due to the increasing demand, with its main products as single-layer, double-layer, and corrugated cups.  
The authors have demonstrated that by using the proposed DT model, not only are the flaws in factory design identified timely, the solutions to overcome the flaws can also be identified. Moreover, the author shows that the modular approach duration for building a DT model is significantly reduced, improving the feasibility of applying DTs to changeable factory design.

\citet{hauge_employing_2020} studied the application of DT to support decision-making processes in two different areas: workstation design and logistics operation analysis. As explained by the authors, the main task of logistics is to handle and provide material at the right time, quantity, quality, and place. Logistics operations often comprise of manual effort and may not be fully automated.  The authors investigate how DTs can contribute to supporting the decision-making process of selecting the right components for a specific company. The authors emphasize that the granularity of the DT model depends on the intended use. For example, the DT model for the interaction of the AGV and the picking robot in a picking process is more detailed compared to the DT model that looks at the material flow from a warehouse to an assembly area.

\begin{summary}{Digital Twins}
In summary, a DT is a simulation-based planning and optimization technology that makes use of the real-time data transformation between the physical and its virtual counterpart, enabling the virtual system's dynamic update, leading to a reliable simulation, and hence better prediction and decisions can be made for the physical system. The online data is collected using IoT platforms, which are then stored and processed using the big  data applications. The DT layer then receives the selected data to identify the current and future (forecasted) possible ambiguities, if there are any, between the physical and virtual worlds using AI technologies. This helps provide recommendations for on-the-fly adjustments and optimize its functionality towards the desired goal. DTs are widely applied in manufacturing space for effective maintenance planning, life cycle management, supply chain, layout and factory design. However, as highlighted in the literature, a DT is a complex system consisting of many layers, and its success relies on the use-specific selection of the critical components that include granularity of data and system entities to avoid an over-complicated virtual model; the integration of data at the right frequency/interval (not necessarily continuous); and the correct format and accuracy of physical data passed to the digital model to ensure that the virtual model is aware of the correct status of its physical counterpart at any time and the right decisions can be recommended. 
The movement analytics can help make the virtual model of a DT more reliable because of its ability to identify and predict the more accurate status of different system entities involved in a manufacturing system. 
Furthermore, it is mentioned in the literature that various AI/ML and Bayesian network-based techniques are used to draw the correct inference from data/information regarding the status of the different entities; however, there is a lack of research about increasing the efficiency of DTs by hybridizing these techniques.
\end{summary}

\section{Conclusions}
\label{sec:conclusions}
In the introduction, we described \emph{movement analytics} as the process of gleaning
knowledge from tracking data so that it aids meaningful decision making. 
The main part of this review was then devoted to better understand the state of the
art on movement analytics in the context of (indoor) manufacturing and other relevant areas.

We kept the scope of this review wide. Methods, applications and problems that qualified
as potentially relevant were considered in scope. We included fundamental methods for
spatio-temporal analysis from diverse areas such as logic-based knowledge representation,
ML, constraint processing, and combinations thereof. We looked into
applications of movement analytics across various industries, and we reviewed what commercial systems
are on the market that provide movement analytics in manufacturing.
To our knowledge, this review is unique with respect to such diversity. 


We conclude this review with a summary of each of the sections and propose ideas for future research.

\littleparagraph{Classical Logic and Knowledge
  Representation~(Section~\ref{sec:classical-logic-KR}).} Classical logics, such as first-order logic, and variations, such as, e.g., description logics
support fundamental needs
for movement analysis in terms of representation of internal structure of objects, time and
space. While pure classical logic-based applications for real-world movement
analytics are rare, spatial and temporal logics can be of value in supporting roles. For
example, temporal logic is an obvious choice for specifying constraints in object
movement planning and monitoring.
It would be interesting to see how advanced generic reasoning schemes like SMT solving and
description logics can be
instantiated with (spatial) theories and utilized in a
temporalized environment for movement analytics.

Knowledge representation logics can be built on a closed-world semantics
of first-order logic. This makes reasoning non-monotonic but enables drawing strong
conclusions by way of default reasoning -- a most useful concept for domain modeling in general.
Prominent realizations are (answer set) logic programming and,
implicitly, relational databases. With respect to movement analytics, the logic programming based event calculus, some stream
processing and other symbolic trajectories techniques fall into this category. Probabilistic variants can be of particular interest as they add capabilities for modeling probabilistic transition systems. We have the impression that more research in this direction could be done. 

\littleparagraph{Probabilistic Transition
  Systems~(Section~\ref{sec:probabilistic-transition-systems}).}
Probabilistic state space models enable dynamic systems, such as those with applications to movement data to be modeled. They can be formulated for pre-processing, prediction, classification and anomaly detection tasks. Whilst state space models offer some advantages relative to neural based sequence models, such as the ability to represent domain knowledge and uncertainty, unlike neural models, they cannot represent long term sequential dependencies. An interesting area of current research is developing hybrid models that parameterize state space models with neural based sequence models to provide the best of both worlds: interpretability and uncertainty in conjunction with long term sequential memory.

  
   
\littleparagraph{Trajectory pre-processing techniques (Section~\ref{sec:trajectories-preprocessing}).}
Pre-processing is a fundamental component of trajectory modeling, independent of whether ML, probabilistic or logic-based methods are employed. Pre-processing methods often reduce the complexity required to model raw trajectory data downstream. Common methods include noise reduction of erroneous position estimates, harmonizing trajectories of non-uniform length and/or sampling interval, segmentation of long trajectories into shorter, more homogeneous subsections, and mapping raw trajectories into semantically labeled sequences. 

\littleparagraph{Neural Networks Based Sequence Models (Section~\ref{sec:machine-learning}).}
Neural network based sequence models have potential value to movement based
manufacturing applications with respect to pre-processing, classification, prediction and
anomaly detection tasks. As mentioned previously, unlike  i.i.d. (independent and identically distributed ) ML or state space models, neural network
sequence models can represent long term spatio-temporal dependencies within motion data. Current research focuses on how to effectively model long sequences and how to do so in a computationally efficient manner. 

The manner in which domain knowledge and context can be included into ML
models is limited. For instance, there remain many open questions about how
symbolic knowledge can be effectively represented, aligned and fused within ML models. 

\littleparagraph{Integration of Logic-Based Methods with
  Probabilistic/ML Methods (Section~\ref{sec:ML-non-ML-integration}).}
A weakness of today's neural network architectures is their
  lack of capability to incorporate domain knowledge, specifically symbolic knowledge
  \cite{singer_rise_2021,marcus_next_2020}.  Integrating logic (for symbolic knowledge) with 
  ML is a still an emerging topic~\cite{davis_symbolic_2020}.
We expect that progress in this area will help advancing
movement analytics by offering ways for a more holistic comprehension of
trajectory data and the participating objects.

Some established paths are available already today.
For instance, one could build heterogeneous architectures comprised of different 
methods for (sensor) data acquisition, filtering, aggregation and evaluating  in a more global context. 
One way of approaching this is by transfer of results in the \emph{data integration} area.\footnote{
The data integration problem, in general, is to provide uniform access to multiple
heterogeneous information sources.}
A key element in data integration is a language capable of describing and querying data sources over a
uniform, mediated schema. Logical languages have been shown to be very useful for that~\cite{levy_logic-based_2000}, 
including logic programming~\cite{genesereth_data_2010,bravo_logic_2003} and description logics~\cite{calvanese_description_2002,caruccio_visual_2014}.

Statistical Relational AI (SRAI) is another relevant research area.  It uses first-order logic for 
 relational structure within and between objects of discourse and probabilities 
 for uncertainty. SRAI methods support probabilistic inference and learning. A prominent representative is
 probabilistic logic programming (but there are many more) which has applications to
 trajectory analysis.
 A rather recent development is (probabilistic)  
 logic programming coupled with deep neural networks.
 There are already applications to movement analysis, but, generally speaking, the area seems to offer unexplored potential. This could be a promising direction for movement analytic applications
 via integrating  generative modeling, state transitions systems, probabilistic
 inference, learning and deep models. 
 
Other proposals set out from deep learning architectures and add ``logic'' components to it.   
According to \citet{dash_review_2022}, this can be done in a variety of ways, (a) by transforming data; (b) by transforming the loss function informed by a domain model; and (c) by transforming the model, e.g., by modeling logic operators within the network itself.  All these methods appear generic enough to be applicable to movement analytics.
However, questions remain if the expressive power of the supported logics are strong enough for all purposes (need support for space and time).
 
As an overall observation, it would be important to understand when combinations of ML and logic are not working. This is being addressed in areas where knowing confidence in conclusions is paramount~\cite{tensorflow_core_uncertainty-aware_2022}, e.g., autonomous driving, but becomes even more relevant when more (logic) components with their own set of shortcomings come into play.

\littleparagraph{Constraint Optimization (Section~\ref{sec:constraint-optimisation}).}
Constraint optimization is a consumer of curated data from big data and movement analytics to make better 
decisions in production scheduling and factory layout problems.
The curated data can provide more realistic value estimations, real-time information, and
discovery of work patterns as an input for optimization methods.
Since these methods are the consumer of such information, standard optimization techniques can
be applied.

A possible direction for future research
lies in improving optimization capabilities for deriving accurate information about the (state of)
a manufacturing process, e.g., by adapting movement analytics techniques from path planning 
(see~\citet{yang_obstacle_2022, lutz_temporal_2008}). 
Another possible direction is to explore the question how to combine constraint optimization 
techniques to with ML and logic techniques in order to bridge a semantic gap between ML methods.

\littleparagraph{Applications of Movement Data
  Analytics (Section~\ref{sec:applications-movement-data-analytics}).}
Our review indicates that movement analytics applications for 
manufacturing
revolve around efficiency monitoring, production control, safety and collaboration.
By and large, they fall under categories of  trajectory anomaly detection,
prediction, clustering and representation (approximation, dimensionality reduction).
The use of generic i.i.d.\ ML and data mining methods is prevalent within these applications. However, such models have limitations in fitting 
trajectory data sets, given the long term spatio-temporal dependencies present within the object motion.
Consequently, the uptake of sequence based neural architectures, as outlined in Section \ref{sec:machine-learning}, would be beneficial for movement based manufacturing applications. Furthermore, it could be beneficial to include motion properties extracted from trajectories, such as velocity and acceleration, as the inputs to ML models.




\littleparagraph{Industrial Applications (Section~\ref{sec:industrial-applications}).}
Movement analytics plays an important role in food, transportation, supply chain and health industries, among others. Specific applications target (real-time) data-informed decision-making informed by current system state, e.g., path optimization to avoid collisions with minimum turns and distance travel; layout optimization to increase efficient use of resources and system throughput; collision avoidance in systems with automated vehicles and robots; demand and inventory management through consumer behavior; and flow management to avoid congestion (especially in the transportation industry). Several ML/AI, optimization, and Bayesian network-based methods have been proposed for making inferences from movement data. We expect an opportunity in combining these and other different techniques, as summarized above, and put them into actual industrial (manufacturing) use.

\littleparagraph{Commercial Systems (Section~\ref{sec:commercial-systems}).}
The focus of most commercial products is to accurately produce indoor tracking data. There are some products on movement analytics in manufacturing with a focus on improving organization, monitoring, and improving factory performance. These analytics services are limited in nature, and it is not clear how underlying analytics are done (most likely to protect IP). This leaves an opportunity for more transparent and
explainable analytics approaches. By including analysis from other types of movement based analytics approaches (based on kinetic e.g. accelerator), a wider range of other decision making is possible from movement data. This is also a gap in this space that can be researched further.

\littleparagraph{Digital Twins (Section~\ref{sec:digital-twins}).}
We have found wide applications of DTs in industries specifically in manufacturing, such
as for effective maintenance planning, life cycle management, supply chain, layout, and
factory design. The success of DTs largely depends on how accurately the virtual model
reflects its physical counterpart and the size of the virtual model.
This by itself is a non-trivial problem that can be addressed by paying attention to basic implementation,
architecture aspects, and sensor methods such as RTLS for updating simulations.

We speculate that movement analytics can also contribute to addressing this problem through
capabilities to identify an entities state when it can potentially impact the system's
performance. However, we have not found any paper
explicitly mentioning the use of movement analytics for such purposes.

Furthermore, it is mentioned in the literature that various AI/ML and Bayesian
network-based techniques are used in making the correct inferences from data/information
regarding the status of the different entities; however, there is a lack of research about
increasing the efficiency of DTs by hybridizing these techniques.

{\small
\bibliography{FDMFPROJ2}

\begin{thebibliography}{279}
\providecommand{\natexlab}[1]{#1}
\providecommand{\url}[1]{\texttt{#1}}
\expandafter\ifx\csname urlstyle\endcsname\relax
  \providecommand{\doi}[1]{doi: #1}\else
  \providecommand{\doi}{doi: \begingroup \urlstyle{rm}\Url}\fi

\bibitem[A.~de Freitas et~al.(2021)A.~de Freitas, Coelho~da Silva, Fernandes~de
  Macêdo, Melo~Junior, and Cordeiro]{a_de_freitas_using_2021}
Nicksson A.~de Freitas, Ticiana Coelho~da Silva, José Fernandes~de Macêdo,
  Leopoldo Melo~Junior, and Matheus Cordeiro.
\newblock Using {Deep} {Learning} for {Trajectory} {Classification}:.
\newblock In \emph{Proceedings of the 13th {International} {Conference} on
  {Agents} and {Artificial} {Intelligence}}, pages 664--671, Online Streaming,
  --- Select a Country ---, 2021. SCITEPRESS - Science and Technology
  Publications.
\newblock ISBN 978-989-758-484-8.
\newblock \doi{10.5220/0010227906640671}.
\newblock URL
  \url{https://www.scitepress.org/DigitalLibrary/Link.aspx?doi=10.5220/0010227906640671}.

\bibitem[Al~Nuaimi and Kamel(2011)]{al_nuaimi_survey_2011}
Klaithem Al~Nuaimi and Hesham Kamel.
\newblock A survey of indoor positioning systems and algorithms.
\newblock In \emph{2011 {International} {Conference} on {Innovations} in
  {Information} {Technology}}, pages 185--190, Abu Dhabi, United Arab Emirates,
  April 2011. IEEE.
\newblock ISBN 978-1-4577-0311-9.
\newblock \doi{10.1109/INNOVATIONS.2011.5893813}.
\newblock URL \url{http://ieeexplore.ieee.org/document/5893813/}.

\bibitem[Al-Shedivat et~al.(2020)Al-Shedivat, Dubey, and
  Xing]{al-shedivat_contextual_2020}
Maruan Al-Shedivat, Avinava Dubey, and Eric Xing.
\newblock Contextual {Explanation} {Networks}.
\newblock \emph{Journal of Machine Learning Research}, 21:\penalty0 1--44,
  2020.

\bibitem[Alahi et~al.(2016)Alahi, Goel, Ramanathan, Robicquet, Fei-Fei, and
  Savarese]{alahi_social_2016}
Alexandre Alahi, Kratarth Goel, Vignesh Ramanathan, Alexandre Robicquet,
  Li~Fei-Fei, and Silvio Savarese.
\newblock Social {LSTM}: {Human} {Trajectory} {Prediction} in {Crowded}
  {Spaces}.
\newblock In \emph{2016 {IEEE} {Conference} on {Computer} {Vision} and
  {Pattern} {Recognition} ({CVPR})}, pages 961--971, Las Vegas, NV, USA, June
  2016. IEEE.
\newblock ISBN 978-1-4673-8851-1.
\newblock \doi{10.1109/CVPR.2016.110}.
\newblock URL \url{http://ieeexplore.ieee.org/document/7780479/}.

\bibitem[Alevizos et~al.(2017)Alevizos, Skarlatidis, Artikis, and
  Paliouras]{alevizos_probabilistic_2017}
Elias Alevizos, Anastasios Skarlatidis, Alexander Artikis, and Georgios
  Paliouras.
\newblock Probabilistic {Complex} {Event} {Recognition}: {A} {Survey}.
\newblock \emph{ACM Computing Surveys}, 50\penalty0 (5):\penalty0 71:1--71:31,
  September 2017.
\newblock ISSN 0360-0300.
\newblock \doi{10.1145/3117809}.
\newblock URL \url{https://doi.org/10.1145/3117809}.

\bibitem[Allen(1983)]{allen_maintaining_1983}
James~F. Allen.
\newblock Maintaining knowledge about temporal intervals.
\newblock \emph{Communications of the ACM}, 26\penalty0 (11):\penalty0
  832--843, November 1983.
\newblock ISSN 0001-0782, 1557-7317.
\newblock \doi{10.1145/182.358434}.
\newblock URL \url{https://dl.acm.org/doi/10.1145/182.358434}.

\bibitem[Altiok(2012)]{altiok_performance_2012}
Tayfur Altiok.
\newblock \emph{Performance {Analysis} of {Manufacturing} {Systems}}.
\newblock Springer Science \& Business Media, December 2012.
\newblock ISBN 978-1-4612-1924-8.
\newblock Google-Books-ID: hdfiBwAAQBAJ.

\bibitem[Alvares et~al.(2007)Alvares, Bogorny, Kuijpers, de~Macedo, Moelans,
  and Vaisman]{alvares_model_2007}
Luis~Otavio Alvares, Vania Bogorny, Bart Kuijpers, Jose Antonio~Fernandes
  de~Macedo, Bart Moelans, and Alejandro Vaisman.
\newblock A model for enriching trajectories with semantic geographical
  information.
\newblock In \emph{Proceedings of the 15th annual {ACM} international symposium
  on {Advances} in geographic information systems - {GIS} '07}, page~1,
  Seattle, Washington, 2007. ACM Press.
\newblock ISBN 978-1-59593-914-2.
\newblock \doi{10.1145/1341012.1341041}.
\newblock URL \url{http://portal.acm.org/citation.cfm?doid=1341012.1341041}.

\bibitem[Analytics(2020)]{soter_analytics_soter_2020}
Soter Analytics.
\newblock Soter {Analytics} - {Reduce} back \& shoulder injuries by up to 55\%,
  April 2020.
\newblock URL \url{https://soteranalytics.com/}.

\bibitem[Andrienko et~al.(2007)Andrienko, Andrienko, and
  Wrobel]{andrienko_visual_2007-1}
Gennady Andrienko, Natalia Andrienko, and Stefan Wrobel.
\newblock Visual analytics tools for analysis of movement data.
\newblock \emph{ACM SIGKDD Explorations Newsletter}, 9\penalty0 (2):\penalty0
  38--46, December 2007.
\newblock ISSN 1931-0145, 1931-0153.
\newblock \doi{10.1145/1345448.1345455}.
\newblock URL \url{https://dl.acm.org/doi/10.1145/1345448.1345455}.

\bibitem[Apt and Bol(1994)]{apt_logic_1994}
Krzysztof~R. Apt and Roland~N. Bol.
\newblock Logic programming and negation: {A} survey.
\newblock \emph{The Journal of Logic Programming}, 19-20:\penalty0 9--71, May
  1994.
\newblock ISSN 0743-1066.
\newblock \doi{10.1016/0743-1066(94)90024-8}.
\newblock URL
  \url{https://www.sciencedirect.com/science/article/pii/0743106694900248}.

\bibitem[Arkan and Van~Landeghem(2013)]{arkan_evaluating_2013}
Ihsan Arkan and Hendrik Van~Landeghem.
\newblock Evaluating the performance of a discrete manufacturing process using
  {RFID}: {A} case study.
\newblock \emph{Robotics and Computer-Integrated Manufacturing}, 29\penalty0
  (6):\penalty0 502--512, December 2013.
\newblock ISSN 0736-5845.
\newblock \doi{10.1016/j.rcim.2013.06.003}.
\newblock URL
  \url{https://www.sciencedirect.com/science/article/pii/S0736584513000471}.

\bibitem[Arp et~al.(2020)Arp, van Vreumingen, Gawehns, and
  Baratchi]{arp_dynamic_2020}
Laurens Arp, Dyon van Vreumingen, Daniela Gawehns, and Mitra Baratchi.
\newblock Dynamic macro scale traffic flow optimisation using crowd-sourced
  urban movement data.
\newblock In \emph{2020 21st {IEEE} {International} {Conference} on {Mobile}
  {Data} {Management} ({MDM})}, pages 168--177, Versailles, France, June 2020.
  IEEE.
\newblock ISBN 978-1-72814-663-8.
\newblock \doi{10.1109/MDM48529.2020.00039}.
\newblock URL \url{https://ieeexplore.ieee.org/document/9162242/}.

\bibitem[Artikis et~al.(2012)Artikis, Skarlatidis, Portet, and
  Paliouras]{artikis_logic-based_2012}
Alexander Artikis, Anastasios Skarlatidis, François Portet, and Georgios
  Paliouras.
\newblock Logic-based event recognition.
\newblock \emph{The Knowledge Engineering Review}, 27\penalty0 (4):\penalty0
  469--506, December 2012.
\newblock ISSN 0269-8889, 1469-8005.
\newblock \doi{10.1017/S0269888912000264}.
\newblock URL
  \url{https://www.cambridge.org/core/product/identifier/S0269888912000264/type/journal_article}.

\bibitem[Arulkumaran et~al.(2017)Arulkumaran, Deisenroth, Brundage, and
  Bharath]{arulkumaran_deep_2017}
Kai Arulkumaran, Marc~Peter Deisenroth, Miles Brundage, and Anil~Anthony
  Bharath.
\newblock Deep {Reinforcement} {Learning}: {A} {Brief} {Survey}.
\newblock \emph{IEEE Signal Processing Magazine}, 34\penalty0 (6):\penalty0
  26--38, November 2017.
\newblock ISSN 1053-5888.
\newblock \doi{10.1109/MSP.2017.2743240}.
\newblock URL \url{http://ieeexplore.ieee.org/document/8103164/}.

\bibitem[Asakura and Hato(2004)]{asakura_tracking_2004}
Yasuo Asakura and Eiji Hato.
\newblock Tracking survey for individual travel behaviour using mobile
  communication instruments.
\newblock \emph{Transportation Research Part C: Emerging Technologies},
  12\penalty0 (3-4):\penalty0 273--291, June 2004.
\newblock ISSN 0968090X.
\newblock \doi{10.1016/j.trc.2004.07.010}.
\newblock URL
  \url{https://linkinghub.elsevier.com/retrieve/pii/S0968090X04000130}.

\bibitem[Assad~Neto et~al.(2021)Assad~Neto, Ribeiro~da Silva, Deschamps, and
  Pinheiro~de Lima]{assad_neto_digital_2021}
Anis Assad~Neto, Elias Ribeiro~da Silva, Fernando Deschamps, and Edson
  Pinheiro~de Lima.
\newblock Digital twins in manufacturing: {An} assessment of key features.
\newblock \emph{Procedia CIRP}, 97:\penalty0 178--183, 2021.
\newblock ISSN 22128271.
\newblock \doi{10.1016/j.procir.2020.05.222}.
\newblock URL
  \url{https://linkinghub.elsevier.com/retrieve/pii/S2212827120314438}.

\bibitem[Atluri et~al.(2018)Atluri, Karpatne, and
  Kumar]{atluri_spatio-temporal_2018}
Gowtham Atluri, Anuj Karpatne, and Vipin Kumar.
\newblock Spatio-{Temporal} {Data} {Mining}: {A} {Survey} of {Problems} and
  {Methods}.
\newblock \emph{ACM Computing Surveys}, 51\penalty0 (4):\penalty0 83:1--83:41,
  August 2018.
\newblock ISSN 0360-0300.
\newblock \doi{10.1145/3161602}.
\newblock URL \url{https://doi.org/10.1145/3161602}.

\bibitem[Baader et~al.(2008)Baader, Horrocks, and
  Sattler]{baader_description_2008}
Franz Baader, Ian Horrocks, and Ulrike Sattler.
\newblock Description {Logics}.
\newblock In \emph{Foundations of {Artificial} {Intelligence}}, volume~3, pages
  135--179. Elsevier, 2008.
\newblock ISBN 978-0-444-52211-5.
\newblock \doi{10.1016/S1574-6526(07)03003-9}.
\newblock URL
  \url{https://linkinghub.elsevier.com/retrieve/pii/S1574652607030039}.

\bibitem[Backes et~al.(2019)Backes, Bayless, Cook, Dodge, Gacek, Hu, Kahsai,
  Kocik, Kotelnikov, Kukovec, McLaughlin, Reed, Rungta, Sizemore, Stalzer,
  Srinivasan, Subotić, Varming, and Whaley]{backes_reachability_2019}
John Backes, Sam Bayless, Byron Cook, Catherine Dodge, Andrew Gacek, Alan~J.
  Hu, Temesghen Kahsai, Bill Kocik, Evgenii Kotelnikov, Jure Kukovec, Sean
  McLaughlin, Jason Reed, Neha Rungta, John Sizemore, Mark Stalzer, Preethi
  Srinivasan, Pavle Subotić, Carsten Varming, and Blake Whaley.
\newblock Reachability {Analysis} for {AWS}-{Based} {Networks}.
\newblock In Isil Dillig and Serdar Tasiran, editors, \emph{Computer {Aided}
  {Verification}}, Lecture {Notes} in {Computer} {Science}, pages 231--241,
  Cham, 2019. Springer International Publishing.
\newblock ISBN 978-3-030-25543-5.
\newblock \doi{10.1007/978-3-030-25543-5_14}.

\bibitem[Ball et~al.(2004)Ball, Cook, Levin, and Rajamani]{ball_slam_2004}
Thomas Ball, Byron Cook, Vladimir Levin, and Sriram~K. Rajamani.
\newblock {SLAM} and {Static} {Driver} {Verifier}: {Technology} {Transfer} of
  {Formal} {Methods} inside {Microsoft}.
\newblock In Eerke~A. Boiten, John Derrick, and Graeme Smith, editors,
  \emph{Integrated {Formal} {Methods}}, Lecture {Notes} in {Computer}
  {Science}, pages 1--20, Berlin, Heidelberg, 2004. Springer.
\newblock ISBN 978-3-540-24756-2.
\newblock \doi{10.1007/978-3-540-24756-2_1}.

\bibitem[Banica and Stefan(2019)]{banica_stepping_2019}
Logica Banica and Cristian Stefan.
\newblock Stepping into the {Industry} 4.0: {The} {Digital} {Twin} {Approach}.
\newblock \emph{Annals of Dunarea de Jos University of Galati. Fascicle I.
  Economics and Applied Informatics}, 25\penalty0 (3):\penalty0 107--113,
  December 2019.
\newblock ISSN 15840409, 2344441X.
\newblock \doi{10.35219/eai1584040962}.
\newblock URL
  \url{http://eia.feaa.ugal.ro/images/eia/2019_3/Banica_Stefan.pdf}.

\bibitem[Baptiste et~al.(2001)Baptiste, Le~Pape, and
  Nuijten]{baptiste_constraint-based_2001}
Philippe Baptiste, Claude Le~Pape, and Wim Nuijten.
\newblock \emph{Constraint-{Based} {Scheduling}}, volume~39 of
  \emph{International {Series} in {Operations} {Research} \& {Management}
  {Science}}.
\newblock Springer US, Boston, MA, 2001.
\newblock ISBN 978-1-4613-5574-8 978-1-4615-1479-4.
\newblock \doi{10.1007/978-1-4615-1479-4}.
\newblock URL \url{http://link.springer.com/10.1007/978-1-4615-1479-4}.

\bibitem[Baral and Gelfond(1994)]{baral_logic_1994}
Chitta Baral and Michael Gelfond.
\newblock Logic programming and knowledge representation.
\newblock \emph{The Journal of Logic Programming}, 19-20:\penalty0 73--148, May
  1994.
\newblock ISSN 0743-1066.
\newblock \doi{10.1016/0743-1066(94)90025-6}.
\newblock URL
  \url{https://www.sciencedirect.com/science/article/pii/0743106694900256}.

\bibitem[Barrett and Tinelli(2018)]{barrett_satisfiability_2018}
Clark Barrett and Cesare Tinelli.
\newblock Satisfiability {Modulo} {Theories}.
\newblock In Edmund~M. Clarke, Thomas~A. Henzinger, Helmut Veith, and Roderick
  Bloem, editors, \emph{Handbook of {Model} {Checking}}, pages 305--343.
  Springer International Publishing, Cham, 2018.
\newblock ISBN 978-3-319-10575-8.
\newblock \doi{10.1007/978-3-319-10575-8_11}.
\newblock URL \url{https://doi.org/10.1007/978-3-319-10575-8_11}.

\bibitem[Bauer et~al.(2011)Bauer, Leucker, and Schallhart]{bauer_runtime_2011}
Andreas Bauer, Martin Leucker, and Christian Schallhart.
\newblock Runtime {Verification} for {LTL} and {TLTL}.
\newblock \emph{ACM Transactions on Software Engineering and Methodology},
  20\penalty0 (4):\penalty0 14:1--14:64, September 2011.
\newblock ISSN 1049-331X.
\newblock \doi{10.1145/2000799.2000800}.
\newblock URL \url{https://doi.org/10.1145/2000799.2000800}.

\bibitem[{Baum, L.E.} and {Petrie, T.}(1966)]{baum_le_statistical_1966}
{Baum, L.E.} and {Petrie, T.}
\newblock Statistical inference for probabilistic functions of finite-state
  {Markov} chains.
\newblock \emph{Annals of MAthematical Statistics}, 37\penalty0 (6):\penalty0
  1554--1563, 1966.

\bibitem[Baumgartner(2021)]{konev_combining_2021}
Peter Baumgartner.
\newblock Combining {Event} {Calculus} and {Description} {Logic} {Reasoning}
  via {Logic} {Programming}.
\newblock In Boris Konev and Giles Reger, editors, \emph{Frontiers of
  {Combining} {Systems}}, volume 12941, pages 98--117. Springer International
  Publishing, Cham, 2021.
\newblock ISBN 978-3-030-86204-6 978-3-030-86205-3.
\newblock \doi{10.1007/978-3-030-86205-3_6}.
\newblock URL \url{https://link.springer.com/10.1007/978-3-030-86205-3_6}.
\newblock Series Title: Lecture Notes in Computer Science.

\bibitem[Baumgartner and Krumpholz(2021)]{baumgartner_anomaly_2021}
Peter Baumgartner and Alexander Krumpholz.
\newblock Anomaly {Detection} in a {Boxed} {Beef} {Supply} {Chain}.
\newblock In \emph{{ICCMS} '21}, pages 1--7. Association for Computing
  Machinery, June 2021.
\newblock ISBN 978-1-4503-8979-2.
\newblock \doi{10.1145/3474963.3474964}.

\bibitem[Bechhofer(2009)]{bechhofer_owl_2009}
Sean Bechhofer.
\newblock {OWL}: {Web} {Ontology} {Language}.
\newblock In Ling Liu and M.~Tamer Özsu, editors, \emph{Encyclopedia of
  {Database} {Systems}}, pages 2008--2009. Springer US, Boston, MA, 2009.
\newblock ISBN 978-0-387-39940-9.
\newblock \doi{10.1007/978-0-387-39940-9_1073}.
\newblock URL \url{https://doi.org/10.1007/978-0-387-39940-9_1073}.

\bibitem[Beck et~al.(2018)Beck, Dao-Tran, and Eiter]{beck_lars_2018}
Harald Beck, Minh Dao-Tran, and Thomas Eiter.
\newblock {LARS}: {A} {Logic}-based framework for {Analytic} {Reasoning} over
  {Streams}.
\newblock \emph{Artificial Intelligence}, 261:\penalty0 16--70, August 2018.
\newblock ISSN 00043702.
\newblock \doi{10.1016/j.artint.2018.04.003}.
\newblock URL
  \url{https://linkinghub.elsevier.com/retrieve/pii/S0004370218301929}.

\bibitem[Belhadi et~al.(2021)Belhadi, Djenouri, Srivastava, Cano, and
  Lin]{belhadi_hybrid_2021}
Asma Belhadi, Youcef Djenouri, Gautam Srivastava, Alberto Cano, and Jerry
  Chun-Wei Lin.
\newblock Hybrid {Group} {Anomaly} {Detection} for {Sequence} {Data}:
  {Application} to {Trajectory} {Data} {Analytics}.
\newblock \emph{IEEE Transactions on Intelligent Transportation Systems}, pages
  1--12, 2021.
\newblock ISSN 1558-0016.
\newblock \doi{10.1109/TITS.2021.3114064}.

\bibitem[Belle(2020)]{davis_symbolic_2020}
Vaishak Belle.
\newblock Symbolic {Logic} {Meets} {Machine} {Learning}: {A} {Brief} {Survey}
  in {Infinite} {Domains}.
\newblock In Jesse Davis and Karim Tabia, editors, \emph{Scalable {Uncertainty}
  {Management}}, volume 12322, pages 3--16. Springer International Publishing,
  Cham, 2020.
\newblock ISBN 978-3-030-58448-1 978-3-030-58449-8.
\newblock \doi{10.1007/978-3-030-58449-8_1}.
\newblock URL \url{http://link.springer.com/10.1007/978-3-030-58449-8_1}.
\newblock Series Title: Lecture Notes in Computer Science.

\bibitem[Bellomarini et~al.(2020)Bellomarini, Laurenza, Sallinger, and
  Sherkhonov]{gutierrez-basulto_reasoning_2020}
Luigi Bellomarini, Eleonora Laurenza, Emanuel Sallinger, and Evgeny Sherkhonov.
\newblock Reasoning {Under} {Uncertainty} in {Knowledge} {Graphs}.
\newblock In Víctor Gutiérrez-Basulto, Tomáš Kliegr, Ahmet Soylu, Martin
  Giese, and Dumitru Roman, editors, \emph{Rules and {Reasoning}}, volume
  12173, pages 131--139. Springer International Publishing, Cham, 2020.
\newblock ISBN 978-3-030-57976-0 978-3-030-57977-7.
\newblock URL \url{https://link.springer.com/10.1007/978-3-030-57977-7_9}.

\bibitem[Beltagy et~al.(2020)Beltagy, Peters, and
  Cohan]{beltagy_longformer_2020}
Iz~Beltagy, Matthew~E. Peters, and Arman Cohan.
\newblock Longformer: {The} {Long}-{Document} {Transformer}.
\newblock \emph{arXiv:2004.05150 [cs]}, December 2020.
\newblock URL \url{http://arxiv.org/abs/2004.05150}.
\newblock arXiv: 2004.05150.

\bibitem[Bian et~al.(2019)Bian, Tian, Tang, and Tao]{bian_trajectory_2019-1}
Jiang Bian, Dayong Tian, Yuanyan Tang, and Dacheng Tao.
\newblock Trajectory {Data} {Classification}: {A} {Review}.
\newblock \emph{ACM Transactions on Intelligent Systems and Technology},
  10\penalty0 (4):\penalty0 33:1--33:34, August 2019.
\newblock ISSN 2157-6904.
\newblock \doi{10.1145/3330138}.
\newblock URL \url{https://doi.org/10.1145/3330138}.

\bibitem[Bienvenu and Ortiz(2015)]{faber_ontology-mediated_2015}
Meghyn Bienvenu and Magdalena Ortiz.
\newblock Ontology-{Mediated} {Query} {Answering} with {Data}-{Tractable}
  {Description} {Logics}.
\newblock In Wolfgang Faber and Adrian Paschke, editors, \emph{Reasoning {Web}.
  {Web} {Logic} {Rules}}, volume 9203, pages 218--307. Springer International
  Publishing, Cham, 2015.
\newblock ISBN 978-3-319-21767-3 978-3-319-21768-0.
\newblock URL \url{http://link.springer.com/10.1007/978-3-319-21768-0_9}.

\bibitem[Blazquez and Vonderohe(2005)]{blazquez_simple_2005}
Carola~A. Blazquez and Alan~P. Vonderohe.
\newblock Simple {Map}-{Matching} {Algorithm} {Applied} to {Intelligent}
  {Winter} {Maintenance} {Vehicle} {Data}.
\newblock \emph{Transportation Research Record}, 1935\penalty0 (1):\penalty0
  68--76, January 2005.
\newblock ISSN 0361-1981.
\newblock \doi{10.1177/0361198105193500108}.
\newblock URL \url{https://doi.org/10.1177/0361198105193500108}.
\newblock Publisher: SAGE Publications Inc.

\bibitem[Borgwardt and Thost(2015)]{borgwardt_temporal_2015}
Stefan Borgwardt and Veronika Thost.
\newblock Temporal query answering in the description logic {EL}.
\newblock In \emph{Proceedings of the 24th {International} {Conference} on
  {Artificial} {Intelligence}}, {IJCAI}'15, pages 2819--2825, Buenos Aires,
  Argentina, July 2015. AAAI Press.
\newblock ISBN 978-1-57735-738-4.

\bibitem[{Box, George} et~al.(2008){Box, George}, {Jenkins, Gwilym}, and
  {Reinsel, Gregory}]{box_george_time_2008}
{Box, George}, {Jenkins, Gwilym}, and {Reinsel, Gregory}.
\newblock \emph{Time {Series} {Analysis}: {Forecasting} and {Control}}.
\newblock Wiley, 4 edition, 2008.

\bibitem[Bragaglia et~al.(2012)Bragaglia, Chesani, Mello, Montali, and
  Torroni]{hutchison_reactive_2012}
Stefano Bragaglia, Federico Chesani, Paola Mello, Marco Montali, and Paolo
  Torroni.
\newblock Reactive {Event} {Calculus} for {Monitoring} {Global} {Computing}
  {Applications}.
\newblock In David Hutchison, Takeo Kanade, Josef Kittler, Jon~M. Kleinberg,
  Friedemann Mattern, John~C. Mitchell, Moni Naor, Oscar Nierstrasz,
  C.~Pandu~Rangan, Bernhard Steffen, Madhu Sudan, Demetri Terzopoulos, Doug
  Tygar, Moshe~Y. Vardi, Gerhard Weikum, Alexander Artikis, Robert Craven,
  Nihan Kesim~Çiçekli, Babak Sadighi, and Kostas Stathis, editors,
  \emph{Logic {Programs}, {Norms} and {Action}}, volume 7360, pages 123--146.
  Springer Berlin Heidelberg, Berlin, Heidelberg, 2012.
\newblock ISBN 978-3-642-29413-6 978-3-642-29414-3.
\newblock \doi{10.1007/978-3-642-29414-3_8}.
\newblock URL \url{http://link.springer.com/10.1007/978-3-642-29414-3_8}.
\newblock Series Title: Lecture Notes in Computer Science.

\bibitem[Bravo and Bertossi(2003)]{bravo_logic_2003}
Loreto Bravo and Leopoldo Bertossi.
\newblock Logic programs for consistently querying data integration systems.
\newblock In \emph{Proceedings of the 18th international joint conference on
  {Artificial} intelligence}, {IJCAI}'03, pages 10--15, San Francisco, CA, USA,
  August 2003. Morgan Kaufmann Publishers Inc.

\bibitem[Bu(2018)]{bu_data_2018}
Fan Bu.
\newblock A {Data} {Mining} {Framework} for {Massive} {RFID} {Data} {Based} on
  {Apriori} {Algorithm}.
\newblock \emph{Journal of Physics: Conference Series}, 1087:\penalty0 022020,
  September 2018.
\newblock ISSN 1742-6596.
\newblock \doi{10.1088/1742-6596/1087/2/022020}.
\newblock URL \url{https://doi.org/10.1088/1742-6596/1087/2/022020}.

\bibitem[Bu et~al.(2012)Bu, Borkar, Carey, Rosen, Polyzotis, Condie, Weimer,
  and Ramakrishnan]{bu_scaling_2012}
Yingyi Bu, Vinayak Borkar, Michael~J. Carey, Joshua Rosen, Neoklis Polyzotis,
  Tyson Condie, Markus Weimer, and Raghu Ramakrishnan.
\newblock Scaling {Datalog} for {Machine} {Learning} on {Big} {Data}.
\newblock \emph{arXiv:1203.0160 [cs]}, March 2012.
\newblock URL \url{http://arxiv.org/abs/1203.0160}.
\newblock arXiv: 1203.0160.

\bibitem[Cai et~al.(2017)Cai, Guo, Yang, and Lu]{cai_mining_2017}
Haoshu Cai, Yu~Guo, Wen-An Yang, and Kun Lu.
\newblock Mining frequent trajectory patterns of {WIP} in {Internet} of
  {Things}-based spatial-temporal database.
\newblock \emph{International Journal of Computer Integrated Manufacturing},
  30\penalty0 (12):\penalty0 1253--1271, December 2017.
\newblock ISSN 0951-192X.
\newblock \doi{10.1080/0951192X.2017.1307522}.
\newblock URL \url{https://doi.org/10.1080/0951192X.2017.1307522}.
\newblock Publisher: Taylor \& Francis \_eprint:
  https://doi.org/10.1080/0951192X.2017.1307522.

\bibitem[Calvanese et~al.(2002)Calvanese, De~Giacomo, and
  Lenzerini]{calvanese_description_2002}
Diego Calvanese, Giuseppe De~Giacomo, and Maurizio Lenzerini.
\newblock Description {Logics} for {Information} {Integration}.
\newblock In Antonis~C. Kakas and Fariba Sadri, editors, \emph{Computational
  {Logic}: {Logic} {Programming} and {Beyond}: {Essays} in {Honour} of {Robert}
  {A}. {Kowalski} {Part} {II}}, Lecture {Notes} in {Computer} {Science}, pages
  41--60. Springer, Berlin, Heidelberg, 2002.
\newblock ISBN 978-3-540-45632-2.
\newblock \doi{10.1007/3-540-45632-5_2}.
\newblock URL \url{https://doi.org/10.1007/3-540-45632-5_2}.

\bibitem[Camacho et~al.(2019)Camacho, Toro~Icarte, Klassen, Valenzano, and
  McIlraith]{camacho_ltl_2019}
Alberto Camacho, Rodrigo Toro~Icarte, Toryn~Q. Klassen, Richard Valenzano, and
  Sheila~A. McIlraith.
\newblock {LTL} and {Beyond}: {Formal} {Languages} for {Reward} {Function}
  {Specification} in {Reinforcement} {Learning}.
\newblock In \emph{Proceedings of the {Twenty}-{Eighth} {International} {Joint}
  {Conference} on {Artificial} {Intelligence}}, pages 6065--6073, Macao, China,
  August 2019. International Joint Conferences on Artificial Intelligence
  Organization.
\newblock ISBN 978-0-9992411-4-1.
\newblock \doi{10.24963/ijcai.2019/840}.
\newblock URL \url{https://www.ijcai.org/proceedings/2019/840}.

\bibitem[Caruccio et~al.(2014)Caruccio, Deufemia, and
  Polese]{caruccio_visual_2014}
Loredana Caruccio, Vincenzo Deufemia, and Giuseppe Polese.
\newblock Visual data integration based on description logic reasoning.
\newblock In \emph{Proceedings of the 18th {International} {Database}
  {Engineering} \& {Applications} {Symposium}}, {IDEAS} '14, pages 19--28, New
  York, NY, USA, July 2014. Association for Computing Machinery.
\newblock ISBN 978-1-4503-2627-8.
\newblock \doi{10.1145/2628194.2628215}.
\newblock URL \url{https://doi.org/10.1145/2628194.2628215}.

\bibitem[Centobelli et~al.(2016)Centobelli, Cerchione, and
  Murino]{centobelli_layout_2016-1}
P.~Centobelli, R.~Cerchione, and T.~Murino.
\newblock Layout and {Material} {Flow} {Optimization} in {Digital} {Factory}.
\newblock \emph{International Journal of Simulation Modelling}, 15\penalty0
  (2):\penalty0 223--235, June 2016.
\newblock ISSN 17264529.
\newblock \doi{10.2507/IJSIMM15(2)3.327}.
\newblock URL
  \url{http://www.ijsimm.com/Full_Papers/Fulltext2016/text15-2_223-235.pdf}.

\bibitem[Ceri et~al.(1990)Ceri, Gottlob, and Tanca]{ceri_logic_1990}
Stefano Ceri, Georg Gottlob, and Letizia Tanca.
\newblock \emph{Logic {Programming} and {Databases}}.
\newblock Surveys in {Computer} {Science}. Springer Berlin Heidelberg, Berlin,
  Heidelberg, 1990.
\newblock ISBN 978-3-642-83954-2 978-3-642-83952-8.
\newblock \doi{10.1007/978-3-642-83952-8}.
\newblock URL \url{http://link.springer.com/10.1007/978-3-642-83952-8}.

\bibitem[Chekol and Stuckenschmidt(2019)]{chekol_time-aware_2019}
Melisachew~Wudage Chekol and Heiner Stuckenschmidt.
\newblock Time-{Aware} {Probabilistic} {Knowledge} {Graphs}.
\newblock In Johann Gamper, Sophie Pinchinat, and Guido Sciavicco, editors,
  \emph{26th {International} {Symposium} on {Temporal} {Representation} and
  {Reasoning}, {TIME} 2019, {October} 16-19, 2019, {Málaga}, {Spain}}, volume
  147 of \emph{{LIPIcs}}, pages 8:1--8:17. Schloss Dagstuhl - Leibniz-Zentrum
  für Informatik, 2019.
\newblock \doi{10.4230/LIPIcs.TIME.2019.8}.

\bibitem[Chen et~al.(2020)Chen, Liu, Achuthan, and Zhang]{chen_ship_2020}
Xiang Chen, Yuanchang Liu, Kamalasudhan Achuthan, and Xinyu Zhang.
\newblock A ship movement classification based on {Automatic} {Identification}
  {System} ({AIS}) data using {Convolutional} {Neural} {Network}.
\newblock \emph{Ocean Engineering}, 218:\penalty0 108182, December 2020.
\newblock ISSN 00298018.
\newblock \doi{10.1016/j.oceaneng.2020.108182}.
\newblock URL
  \url{https://linkinghub.elsevier.com/retrieve/pii/S0029801820311124}.

\bibitem[Chen et~al.(2021)Chen, Xu, Zhou, Chen, Fang, and
  Liu]{chen_trajvae_2021}
Xinyu Chen, Jiajie Xu, Rui Zhou, Wei Chen, Junhua Fang, and Chengfei Liu.
\newblock {TrajVAE}: {A} {Variational} {AutoEncoder} model for trajectory
  generation.
\newblock \emph{Neurocomputing}, 428:\penalty0 332--339, March 2021.
\newblock ISSN 0925-2312.
\newblock \doi{10.1016/j.neucom.2020.03.120}.
\newblock URL
  \url{https://www.sciencedirect.com/science/article/pii/S0925231220312017}.

\bibitem[Cheng et~al.(2020)Cheng, Sun, Liu, and Tomizuka]{cheng_towards_2020}
Yujiao Cheng, Liting Sun, Changliu Liu, and Masayoshi Tomizuka.
\newblock Towards {Efficient} {Human}-{Robot} {Collaboration} {With} {Robust}
  {Plan} {Recognition} and {Trajectory} {Prediction}.
\newblock \emph{IEEE Robotics and Automation Letters}, 5\penalty0 (2):\penalty0
  2602--2609, April 2020.
\newblock ISSN 2377-3766.
\newblock \doi{10.1109/LRA.2020.2972874}.

\bibitem[Choromanski et~al.(2021)Choromanski, Likhosherstov, Dohan, Song, Gane,
  Sarlos, Hawkins, Davis, Mohiuddin, Kaiser, Belanger, Colwell, and
  Weller]{choromanski_rethinking_2021}
Krzysztof Choromanski, Valerii Likhosherstov, David Dohan, Xingyou Song,
  Andreea Gane, Tamas Sarlos, Peter Hawkins, Jared Davis, Afroz Mohiuddin,
  Lukasz Kaiser, David Belanger, Lucy Colwell, and Adrian Weller.
\newblock Rethinking {Attention} with {Performers}.
\newblock \emph{arXiv:2009.14794 [cs, stat]}, March 2021.
\newblock URL \url{http://arxiv.org/abs/2009.14794}.
\newblock arXiv: 2009.14794.

\bibitem[Chung et~al.(2014)Chung, Gulcehre, Cho, and
  Bengio]{chung_empirical_2014}
Junyoung Chung, Caglar Gulcehre, KyungHyun Cho, and Yoshua Bengio.
\newblock Empirical {Evaluation} of {Gated} {Recurrent} {Neural} {Networks} on
  {Sequence} {Modeling}.
\newblock \emph{arXiv:1412.3555 [cs]}, December 2014.
\newblock URL \url{http://arxiv.org/abs/1412.3555}.
\newblock arXiv: 1412.3555.

\bibitem[Cimino et~al.(2019)Cimino, Negri, and Fumagalli]{cimino_review_2019}
Chiara Cimino, Elisa Negri, and Luca Fumagalli.
\newblock Review of digital twin applications in manufacturing.
\newblock \emph{Computers in Industry}, 113:\penalty0 103130, December 2019.
\newblock ISSN 01663615.
\newblock \doi{10.1016/j.compind.2019.103130}.
\newblock URL
  \url{https://linkinghub.elsevier.com/retrieve/pii/S0166361519304385}.

\bibitem[Cinar et~al.(2017)Cinar, Mirisaee, Goswami, Gaussier, Aït-Bachir, and
  Strijov]{cinar_position-based_2017}
Yagmur~Gizem Cinar, Hamid Mirisaee, Parantapa Goswami, Eric Gaussier, Ali
  Aït-Bachir, and Vadim Strijov.
\newblock Position-{Based} {Content} {Attention} for {Time} {Series}
  {Forecasting} with {Sequence}-to-{Sequence} {RNNs}.
\newblock In Derong Liu, Shengli Xie, Yuanqing Li, Dongbin Zhao, and
  El-Sayed~M. El-Alfy, editors, \emph{Neural {Information} {Processing}},
  Lecture {Notes} in {Computer} {Science}, pages 533--544, Cham, 2017. Springer
  International Publishing.
\newblock ISBN 978-3-319-70139-4.
\newblock \doi{10.1007/978-3-319-70139-4_54}.

\bibitem[Cohn and Renz(2008)]{cohn_qualitative_2008}
Anthony~G. Cohn and Jochen Renz.
\newblock Qualitative {Spatial} {Representation} and {Reasoning}.
\newblock In \emph{Foundations of {Artificial} {Intelligence}}, volume~3, pages
  551--596. Elsevier, 2008.
\newblock ISBN 978-0-444-52211-5.
\newblock \doi{10.1016/S1574-6526(07)03013-1}.
\newblock URL
  \url{https://linkinghub.elsevier.com/retrieve/pii/S1574652607030131}.

\bibitem[Collins and Cockburn(2020)]{collins_beyond_2020}
Anne G.~E. Collins and Jeffrey Cockburn.
\newblock Beyond dichotomies in reinforcement learning.
\newblock \emph{Nature Reviews Neuroscience}, 21\penalty0 (10):\penalty0
  576--586, October 2020.
\newblock ISSN 1471-0048.
\newblock \doi{10.1038/s41583-020-0355-6}.
\newblock URL \url{https://www.nature.com/articles/s41583-020-0355-6}.
\newblock Number: 10 Publisher: Nature Publishing Group.

\bibitem[Cook(2018)]{cook_formal_2018}
Byron Cook.
\newblock Formal {Reasoning} {About} the {Security} of {Amazon} {Web}
  {Services}.
\newblock In Hana Chockler and Georg Weissenbacher, editors, \emph{Computer
  {Aided} {Verification}}, Lecture {Notes} in {Computer} {Science}, pages
  38--47, Cham, 2018. Springer International Publishing.
\newblock ISBN 978-3-319-96145-3.
\newblock \doi{10.1007/978-3-319-96145-3_3}.

\bibitem[Core(2022)]{tensorflow_core_uncertainty-aware_2022}
TensorFlow Core.
\newblock Uncertainty-aware {Deep} {Learning} with {SNGP}, 2022.
\newblock URL \url{https://www.tensorflow.org/tutorials/understanding/sngp}.

\bibitem[Cowling and Johansson(2002)]{cowling_using_2002}
Peter Cowling and Marcus Johansson.
\newblock Using real time information for effective dynamic scheduling.
\newblock \emph{European Journal of Operational Research}, 139\penalty0
  (2):\penalty0 230--244, June 2002.
\newblock ISSN 03772217.
\newblock \doi{10.1016/S0377-2217(01)00355-1}.
\newblock URL
  \url{https://linkinghub.elsevier.com/retrieve/pii/S0377221701003551}.

\bibitem[CSIRO()]{csiro_secure_nodate}
CSIRO.
\newblock Secure {Intelligent} {IoT} for {Digital} {Manufacturing}.
\newblock URL
  \url{https://www.csiro.au/en/work-with-us/industries/manufacturing/future-digital-manufacturing-fund/secure-intelligent-iot-for-digital-manufacturing}.
\newblock Publisher: CSIRO.

\bibitem[Dabrowski and Rahman(2019)]{dabrowski_sensor_2019}
Joel Dabrowski and Ashfaqur Rahman.
\newblock Sensor {Data} {Analytics} for {Fruit} {Picker} {Bag} {Drop}
  {Detection}: {A} {Feasibility} {Study}.
\newblock Technical Report EP191416, Data61, CSIRO, 2019.
\newblock URL
  \url{https://epublish.csiro.au/v7y18/sensor-data-analytics-for-fruit-}.

\bibitem[Dahmen and Rossmann(2021)]{dahmen_what_2021}
Ulrich Dahmen and Juergen Rossmann.
\newblock What is a {Digital} {Twin} – {A} {Mediation} {Approach}.
\newblock In \emph{2021 {IEEE} {International} {Conference} on {Electro}
  {Information} {Technology} ({EIT})}, pages 165--172, Mt. Pleasant, MI, USA,
  May 2021. IEEE.
\newblock ISBN 978-1-66541-846-1.
\newblock \doi{10.1109/EIT51626.2021.9491883}.
\newblock URL \url{https://ieeexplore.ieee.org/document/9491883/}.

\bibitem[Damjanovic-Behrendt and Behrendt(2019)]{damjanovic-behrendt_open_2019}
Violeta Damjanovic-Behrendt and Wernher Behrendt.
\newblock An open source approach to the design and implementation of {Digital}
  {Twins} for {Smart} {Manufacturing}.
\newblock \emph{International Journal of Computer Integrated Manufacturing},
  32\penalty0 (4-5):\penalty0 366--384, May 2019.
\newblock ISSN 0951-192X, 1362-3052.
\newblock \doi{10.1080/0951192X.2019.1599436}.
\newblock URL
  \url{https://www.tandfonline.com/doi/full/10.1080/0951192X.2019.1599436}.

\bibitem[Dash et~al.(2022)Dash, Chitlangia, Ahuja, and
  Srinivasan]{dash_review_2022}
Tirtharaj Dash, Sharad Chitlangia, Aditya Ahuja, and Ashwin Srinivasan.
\newblock A review of some techniques for inclusion of domain-knowledge into
  deep neural networks.
\newblock \emph{Scientific Reports 2022 12:1}, 12\penalty0 (1):\penalty0 1--15,
  January 2022.
\newblock ISSN 2045-2322.
\newblock \doi{10.1038/s41598-021-04590-0}.
\newblock URL \url{https://www.nature.com/articles/s41598-021-04590-0}.
\newblock Publisher: Nature Publishing Group ISBN: 0123456789.

\bibitem[{Data Dog}(2022)]{data_dog_monitor_2022}
{Data Dog}.
\newblock Monitor your {IoT} devices and backend services in a single unified
  platform, 2022.
\newblock URL \url{https://www.datadoghq.com/dg/monitor/iot/}.

\bibitem[De~Raedt and Kersting(2008)]{de_raedt_probabilistic_2008}
Luc De~Raedt and Kristian Kersting.
\newblock Probabilistic {Inductive} {Logic} {Programming}.
\newblock In Luc De~Raedt, Paolo Frasconi, Kristian Kersting, and Stephen
  Muggleton, editors, \emph{Probabilistic {Inductive} {Logic} {Programming}:
  {Theory} and {Applications}}, Lecture {Notes} in {Computer} {Science}, pages
  1--27. Springer, Berlin, Heidelberg, 2008.
\newblock ISBN 978-3-540-78652-8.
\newblock \doi{10.1007/978-3-540-78652-8_1}.
\newblock URL \url{https://doi.org/10.1007/978-3-540-78652-8_1}.

\bibitem[De~Raedt et~al.(2007)De~Raedt, Kimmig, and
  Toivonen]{de_raedt_problog_2007}
Luc De~Raedt, Angelika Kimmig, and Hannu Toivonen.
\newblock {ProbLog}: a probabilistic prolog and its application in link
  discovery.
\newblock In \emph{Proceedings of the 20th international joint conference on
  {Artifical} intelligence}, {IJCAI}'07, pages 2468--2473, San Francisco, CA,
  USA, January 2007. Morgan Kaufmann Publishers Inc.

\bibitem[de~Vries and van Someren(2012)]{de_vries_machine_2012}
Gerben Klaas~Dirk de~Vries and Maarten van Someren.
\newblock Machine learning for vessel trajectories using compression,
  alignments and domain knowledge.
\newblock \emph{Expert Systems with Applications}, 39\penalty0 (18):\penalty0
  13426--13439, December 2012.
\newblock ISSN 0957-4174.
\newblock \doi{10.1016/j.eswa.2012.05.060}.
\newblock URL
  \url{https://www.sciencedirect.com/science/article/pii/S0957417412007762}.

\bibitem[Dean and Kanazawa(1989)]{dean_model_1989}
Thomas Dean and Keiji Kanazawa.
\newblock A model for reasoning about persistence and causation.
\newblock \emph{Computational Intelligence}, 5\penalty0 (2):\penalty0 142--150,
  1989.
\newblock ISSN 1467-8640.
\newblock \doi{10.1111/j.1467-8640.1989.tb00324.x}.

\bibitem[{Del Moral, Pierre}(1996)]{del_moral_pierre_non_1996}
{Del Moral, Pierre}.
\newblock Non {Linear} {Filtering}: {Interacting} {Particle} {Solution}.
\newblock \emph{Markov Processes and Related Fields}, 2\penalty0 (4):\penalty0
  555--580, 1996.

\bibitem[Ding et~al.(2018)Ding, Wang, Ge, and Li]{ding_anomaly_2018}
Feng Ding, Jian Wang, Jiaqi Ge, and Wenfeng Li.
\newblock Anomaly {Detection} in {Large}-{Scale} {Trajectories} {Using}
  {Hybrid} {Grid}-{Based} {Hierarchical} {Clustering}.
\newblock \emph{International Journal of Robotics and Automation}, 33\penalty0
  (5), 2018.
\newblock ISSN 1925-7090.
\newblock \doi{10.2316/Journal.206.2018.5.206-0061}.
\newblock URL \url{http://www.actapress.com/PaperInfo.aspx?paperId=45822}.

\bibitem[Ding et~al.(2011)Ding, Smith, Belta, and Rus]{ding_mdp_2011}
Xu~Chu Ding, Stephen~L. Smith, Calin Belta, and Daniela Rus.
\newblock {MDP} optimal control under temporal logic constraints.
\newblock In \emph{{IEEE} {Conference} on {Decision} and {Control} and
  {European} {Control} {Conference}}, pages 532--538, Orlando, FL, USA,
  December 2011. IEEE.
\newblock ISBN 978-1-61284-801-3 978-1-61284-800-6 978-1-4673-0457-3
  978-1-61284-799-3.
\newblock \doi{10.1109/CDC.2011.6161122}.
\newblock URL \url{http://ieeexplore.ieee.org/document/6161122/}.

\bibitem[Dobler et~al.(2020)Dobler, Schumacher, Busel, and
  Hartmann]{dobler_supporting_2020}
Martin Dobler, Jens Schumacher, Philipp Busel, and Christian Hartmann.
\newblock Supporting {SMEs} in the {Lake} {Constance} {Region} in the
  {Implementation} of {Cyber}-{Physical}-{Systems}: {Framework} and
  {Demonstrator}.
\newblock In \emph{2020 {IEEE} {International} {Conference} on {Engineering},
  {Technology} and {Innovation} ({ICE}/{ITMC})}, pages 1--8, Cardiff, United
  Kingdom, June 2020. IEEE.
\newblock ISBN 978-1-72817-037-4.
\newblock \doi{10.1109/ICE/ITMC49519.2020.9198430}.
\newblock URL \url{https://ieeexplore.ieee.org/document/9198430/}.

\bibitem[Dolgui et~al.(2021)Dolgui, Bernard, Lemoine, von Cieminski, and
  Romero]{dolgui_advances_2021}
Alexandre Dolgui, Alain Bernard, David Lemoine, Gregor von Cieminski, and David
  Romero, editors.
\newblock \emph{Advances in {Production} {Management} {Systems}. {Artificial}
  {Intelligence} for {Sustainable} and {Resilient} {Production} {Systems}:
  {IFIP} {WG} 5.7 {International} {Conference}, {APMS} 2021, {Nantes},
  {France}, {September} 5–9, 2021, {Proceedings}, {Part} {IV}}, volume 633 of
  \emph{{IFIP} {Advances} in {Information} and {Communication} {Technology}}.
\newblock Springer International Publishing, Cham, 2021.
\newblock ISBN 978-3-030-85909-1 978-3-030-85910-7.
\newblock \doi{10.1007/978-3-030-85910-7}.
\newblock URL \url{https://link.springer.com/10.1007/978-3-030-85910-7}.

\bibitem[Domingos and Lowd(2019)]{domingos_unifying_2019}
Pedro Domingos and Daniel Lowd.
\newblock Unifying logical and statistical {AI} with {Markov} logic.
\newblock \emph{Communications of the ACM}, 62\penalty0 (7):\penalty0 74--83,
  June 2019.
\newblock ISSN 0001-0782.
\newblock \doi{10.1145/3241978}.
\newblock URL \url{https://doi.org/10.1145/3241978}.

\bibitem[Dries et~al.(2015)Dries, Kimmig, Meert, Renkens, Van~den Broeck,
  Vlasselaer, and De~Raedt]{dries_problog2_2015}
Anton Dries, Angelika Kimmig, Wannes Meert, Joris Renkens, Guy Van~den Broeck,
  Jonas Vlasselaer, and Luc De~Raedt.
\newblock {ProbLog2}: {Probabilistic} {Logic} {Programming}.
\newblock In Albert Bifet, Michael May, Bianca Zadrozny, Ricard Gavalda, Dino
  Pedreschi, Francesco Bonchi, Jaime Cardoso, and Myra Spiliopoulou, editors,
  \emph{Machine {Learning} and {Knowledge} {Discovery} in {Databases}}, Lecture
  {Notes} in {Computer} {Science}, pages 312--315, Cham, 2015. Springer
  International Publishing.
\newblock ISBN 978-3-319-23461-8.
\newblock \doi{10.1007/978-3-319-23461-8_37}.

\bibitem[Du et~al.(2020)Du, Turrin, Jansen, van~den Dobbelsteen, and
  Fang]{du_gaps_2020}
Tiantian Du, Michela Turrin, Sabine Jansen, Andy van~den Dobbelsteen, and Jian
  Fang.
\newblock Gaps and requirements for automatic generation of space layouts with
  optimised energy performance.
\newblock \emph{Automation in Construction}, 116:\penalty0 103132, August 2020.
\newblock ISSN 09265805.
\newblock \doi{10.1016/j.autcon.2020.103132}.
\newblock URL
  \url{https://linkinghub.elsevier.com/retrieve/pii/S0926580519307496}.

\bibitem[Du et~al.(2021)Du, Zhang, Cao, Wang, Liang, Zhang, and
  Tang]{du_optimized_2021}
Yiquan Du, Xiuguo Zhang, Zhiying Cao, Shaobo Wang, Jiacheng Liang, Fengge
  Zhang, and Jiawei Tang.
\newblock An {Optimized} {Path} {Planning} {Method} for {Coastal} {Ships}
  {Based} on {Improved} {DDPG} and {DP}.
\newblock \emph{Journal of Advanced Transportation}, 2021:\penalty0 1--23,
  October 2021.
\newblock ISSN 2042-3195, 0197-6729.
\newblock \doi{10.1155/2021/7765130}.
\newblock URL \url{https://www.hindawi.com/journals/jat/2021/7765130/}.

\bibitem[Duan and Cao(2020)]{duan_emerging_2020}
Kang-Kang Duan and Shuang-Yin Cao.
\newblock Emerging {RFID} technology in structural engineering – {A} review.
\newblock \emph{Structures}, 28:\penalty0 2404--2414, December 2020.
\newblock ISSN 23520124.
\newblock \doi{10.1016/j.istruc.2020.10.036}.
\newblock URL
  \url{https://linkinghub.elsevier.com/retrieve/pii/S2352012420305968}.

\bibitem[Dzeng et~al.(2014)Dzeng, Lin, and Hsiao]{dzeng_application_2014}
Ren-Jye Dzeng, Chong-Wey Lin, and Fan-Yi Hsiao.
\newblock Application of {RFID} tracking to the optimization of function-space
  assignment in buildings.
\newblock \emph{Automation in Construction}, 40:\penalty0 68--83, April 2014.
\newblock ISSN 09265805.
\newblock \doi{10.1016/j.autcon.2013.12.011}.
\newblock URL
  \url{https://linkinghub.elsevier.com/retrieve/pii/S092658051300232X}.

\bibitem[Eiter and Kiesel(2020)]{eiter_weighted_2020}
Thomas Eiter and Rafael Kiesel.
\newblock Weighted {LARS} for {Quantitative} {Stream} {Reasoning}.
\newblock In \emph{{ECAI} 2020 - 24th {European} {Conference} on {Artificial}
  {Intelligence}}, pages 729--736, 2020.

\bibitem[Feinberg et~al.(2002)Feinberg, Shwartz, and
  Hillier]{feinberg_handbook_2002}
Eugene~A. Feinberg, Adam Shwartz, and Frederick~S. Hillier, editors.
\newblock \emph{Handbook of {Markov} {Decision} {Processes}}, volume~40 of
  \emph{International {Series} in {Operations} {Research} \& {Management}
  {Science}}.
\newblock Springer US, Boston, MA, 2002.
\newblock ISBN 978-1-4613-5248-8 978-1-4615-0805-2.
\newblock \doi{10.1007/978-1-4615-0805-2}.
\newblock URL \url{http://link.springer.com/10.1007/978-1-4615-0805-2}.

\bibitem[Ferreira et~al.(2021)Ferreira, Lopes, Gonçalves, Knorr, Krippahl, and
  Leite]{ferreira_deep_2021}
Ricardo Ferreira, Carolina Lopes, Ricardo Gonçalves, Matthias Knorr, Ludwig
  Krippahl, and João Leite.
\newblock Deep {Neural} {Networks} for {Approximating} {Stream} {Reasoning}
  with {C}-{SPARQL}.
\newblock In \emph{Progress in {Artificial} {Intelligence} - 20th {EPIA}
  {Conference} on {Artificial} {Intelligence}}, July 2021.
\newblock URL \url{http://arxiv.org/abs/2106.08452}.

\bibitem[Flossdorf et~al.(2021)Flossdorf, Meyer, Artjuch, Schneider, and
  Jentsch]{flossdorf_unsupervised_2021}
Jonathan Flossdorf, Anne Meyer, Dmitri Artjuch, Jaques Schneider, and Carsten
  Jentsch.
\newblock Unsupervised {Movement} {Detection} in {Indoor} {Positioning}
  {Systems}.
\newblock \emph{arXiv:2109.10757 [cs, stat]}, August 2021.
\newblock URL \url{http://arxiv.org/abs/2109.10757}.
\newblock arXiv: 2109.10757.

\bibitem[Galton(2009)]{galton_spatial_2009}
Antony Galton.
\newblock Spatial and temporal knowledge representation.
\newblock \emph{Earth Science Informatics}, 2\penalty0 (3):\penalty0 169--187,
  September 2009.
\newblock ISSN 18650473.
\newblock \doi{10.1007/s12145-009-0027-6}.

\bibitem[Gao et~al.(2017)Gao, Zhou, Zhang, Trajcevski, Luo, and
  Zhang]{gao_identifying_2017}
Qiang Gao, Fan Zhou, Kunpeng Zhang, Goce Trajcevski, Xucheng Luo, and Fengli
  Zhang.
\newblock Identifying {Human} {Mobility} via {Trajectory} {Embeddings}.
\newblock In \emph{Proceedings of the {Twenty}-{Sixth} {International} {Joint}
  {Conference} on {Artificial} {Intelligence}}, pages 1689--1695, Melbourne,
  Australia, August 2017. International Joint Conferences on Artificial
  Intelligence Organization.
\newblock ISBN 978-0-9992411-0-3.
\newblock \doi{10.24963/ijcai.2017/234}.
\newblock URL \url{https://www.ijcai.org/proceedings/2017/234}.

\bibitem[Ge et~al.(2018)Ge, Renz, and Hua]{ge_towards_2018}
Xiaoyu Ge, Jochen Renz, and Hua Hua.
\newblock Towards {Explainable} {Inference} about {Object} {Motion} using
  {Qualitative} {Reasoning}.
\newblock In Michael Thielscher, Francesca Toni, and Frank Wolter, editors,
  \emph{Principles of {Knowledge} {Representation} and {Reasoning}:
  {Proceedings} of the {Sixteenth} {International} {Conference}, {KR} 2018,
  {Tempe}, {Arizona}, 30 {October} - 2 {November} 2018}, pages 641--642. AAAI
  Press, 2018.
\newblock URL \url{https://aaai.org/ocs/index.php/KR/KR18/paper/view/18044}.

\bibitem[Genesereth(2010)]{genesereth_data_2010}
Michael Genesereth.
\newblock Data {Integration}: {The} {Relational} {Logic} {Approach}.
\newblock \emph{Synthesis Lectures on Artificial Intelligence and Machine
  Learning}, 4\penalty0 (1):\penalty0 1--97, January 2010.
\newblock ISSN 1939-4608.
\newblock \doi{10.2200/S00226ED1V01Y200911AIM008}.
\newblock URL
  \url{https://www.morganclaypool.com/doi/abs/10.2200/S00226ED1V01Y200911AIM008}.
\newblock Publisher: Morgan \& Claypool Publishers.

\bibitem[Ghahramani(1998)]{carbonell_learning_1998}
Zoubin Ghahramani.
\newblock Learning dynamic {Bayesian} networks.
\newblock In Jaime~G. Carbonell, Jörg Siekmann, G.~Goos, J.~Hartmanis, J.~van
  Leeuwen, C.~Lee Giles, and Marco Gori, editors, \emph{Adaptive {Processing}
  of {Sequences} and {Data} {Structures}}, volume 1387, pages 168--197.
  Springer Berlin Heidelberg, Berlin, Heidelberg, 1998.
\newblock ISBN 978-3-540-64341-8 978-3-540-69752-7.
\newblock URL \url{http://link.springer.com/10.1007/BFb0053999}.

\bibitem[Giatrakos et~al.(2020)Giatrakos, Alevizos, Artikis, Deligiannakis, and
  Garofalakis]{giatrakos_complex_2020}
Nikos Giatrakos, Elias Alevizos, Alexander Artikis, Antonios Deligiannakis, and
  Minos Garofalakis.
\newblock Complex event recognition in the {Big} {Data} era: a survey.
\newblock \emph{The VLDB Journal}, 29\penalty0 (1):\penalty0 313--352, January
  2020.
\newblock ISSN 1066-8888, 0949-877X.
\newblock \doi{10.1007/s00778-019-00557-w}.
\newblock URL \url{http://link.springer.com/10.1007/s00778-019-00557-w}.

\bibitem[Giuliari et~al.(2021)Giuliari, Hasan, Cristani, and
  Galasso]{giuliari_transformer_2021}
Francesco Giuliari, Irtiza Hasan, Marco Cristani, and Fabio Galasso.
\newblock Transformer {Networks} for {Trajectory} {Forecasting}.
\newblock In \emph{2020 25th {International} {Conference} on {Pattern}
  {Recognition} ({ICPR})}, pages 10335--10342, January 2021.
\newblock \doi{10.1109/ICPR48806.2021.9412190}.
\newblock ISSN: 1051-4651.

\bibitem[Glaessgen and Stargel(2012)]{glaessgen_digital_2012}
Edward Glaessgen and David Stargel.
\newblock The {Digital} {Twin} {Paradigm} for {Future} {NASA} and {U}.{S}.
  {Air} {Force} {Vehicles}.
\newblock In \emph{53rd {AIAA}/{ASME}/{ASCE}/{AHS}/{ASC} {Structures},
  {Structural} {Dynamics} and {Materials} {Conference}}, Honolulu, Hawaii,
  April 2012. American Institute of Aeronautics and Astronautics.
\newblock ISBN 978-1-60086-937-2.
\newblock \doi{10.2514/6.2012-1818}.
\newblock URL \url{http://arc.aiaa.org/doi/abs/10.2514/6.2012-1818}.

\bibitem[Gogate and Domingos(2016)]{gogate_probabilistic_2016}
Vibhav Gogate and Pedro Domingos.
\newblock Probabilistic theorem proving.
\newblock \emph{Communications of the ACM}, 59\penalty0 (7):\penalty0 107--115,
  June 2016.
\newblock ISSN 0001-0782, 1557-7317.
\newblock \doi{10.1145/2936726}.
\newblock URL \url{https://dl.acm.org/doi/10.1145/2936726}.

\bibitem[Goodfellow et~al.(2014)Goodfellow, Pouget-Abadie, Mirza, Xu,
  Warde-Farley, Ozair, Courville, and Bengio]{goodfellow_generative_2014}
Ian Goodfellow, Jean Pouget-Abadie, Mehdi Mirza, Bing Xu, David Warde-Farley,
  Sherjil Ozair, Aaron Courville, and Yoshua Bengio.
\newblock Generative {Adversarial} {Nets}.
\newblock In \emph{Advances in {Neural} {Information} {Processing} {Systems}},
  volume~27. Curran Associates, Inc., 2014.
\newblock URL
  \url{https://papers.nips.cc/paper/2014/hash/5ca3e9b122f61f8f06494c97b1afccf3-Abstract.html}.

\bibitem[Goranko and Galton(1999)]{goranko_temporal_1999}
Valentin Goranko and Antony Galton.
\newblock Temporal {Logic}, November 1999.
\newblock URL
  \url{https://plato.stanford.edu/archives/win2015/entries/logic-temporal/}.
\newblock Last Modified: 2015-05-20.

\bibitem[Greco and Molinaro(2015)]{greco_datalog_2015}
Sergio Greco and Cristian Molinaro.
\newblock Datalog and {Logic} {Databases}.
\newblock \emph{Synthesis Lectures on Data Management}, 7\penalty0
  (2):\penalty0 1--169, November 2015.
\newblock ISSN 2153-5418, 2153-5426.
\newblock \doi{10.2200/S00648ED1V01Y201505DTM041}.
\newblock URL
  \url{http://www.morganclaypool.com/doi/10.2200/S00648ED1V01Y201505DTM041}.

\bibitem[Grieves(2014)]{grieves_digital_twin_2014}
Michael Grieves.
\newblock Digital {Twin}: {Manufacturing} {Excellence} through {Virtual}
  {Factory} {Replication}, 2014.
\newblock URL
  \url{https://www.3ds.com/fileadmin/PRODUCTS-SERVICES/DELMIA/PDF/Whitepaper/DELMIA-APRISO-Digital-Twin-Whitepaper.pdf}.

\bibitem[Grieves and Vickers(2017)]{kahlen_digital_2017}
Michael Grieves and John Vickers.
\newblock Digital {Twin}: {Mitigating} {Unpredictable}, {Undesirable}
  {Emergent} {Behavior} in {Complex} {Systems}.
\newblock In Franz-Josef Kahlen, Shannon Flumerfelt, and Anabela Alves,
  editors, \emph{Transdisciplinary {Perspectives} on {Complex} {Systems}},
  pages 85--113. Springer International Publishing, Cham, 2017.
\newblock ISBN 978-3-319-38754-3 978-3-319-38756-7.
\newblock \doi{10.1007/978-3-319-38756-7_4}.
\newblock URL \url{http://link.springer.com/10.1007/978-3-319-38756-7_4}.

\bibitem[Gu et~al.(2020)Gu, Guan, and Missier]{gu_towards_2020}
Yulong Gu, Yu~Guan, and Paolo Missier.
\newblock Towards {Learning} {Instantiated} {Logical} {Rules} from {Knowledge}
  {Graphs}.
\newblock \emph{arXiv:2003.06071 [cs]}, May 2020.
\newblock URL \url{http://arxiv.org/abs/2003.06071}.
\newblock arXiv: 2003.06071.

\bibitem[Guo et~al.(2019)Guo, Zhao, Sun, and Zhang]{guo_modular_2019}
Jiapeng Guo, Ning Zhao, Lin Sun, and Saipeng Zhang.
\newblock Modular based flexible digital twin for factory design.
\newblock \emph{Journal of Ambient Intelligence and Humanized Computing},
  10\penalty0 (3):\penalty0 1189--1200, March 2019.
\newblock ISSN 1868-5137, 1868-5145.
\newblock \doi{10.1007/s12652-018-0953-6}.
\newblock URL \url{http://link.springer.com/10.1007/s12652-018-0953-6}.

\bibitem[Gupta et~al.(2018)Gupta, Johnson, Fei-Fei, Savarese, and
  Alahi]{gupta_social_2018}
Agrim Gupta, Justin Johnson, Li~Fei-Fei, Silvio Savarese, and Alexandre Alahi.
\newblock Social {GAN}: {Socially} {Acceptable} {Trajectories} with
  {Generative} {Adversarial} {Networks}.
\newblock In \emph{2018 {IEEE}/{CVF} {Conference} on {Computer} {Vision} and
  {Pattern} {Recognition}}, pages 2255--2264, Salt Lake City, UT, June 2018.
  IEEE.
\newblock ISBN 978-1-5386-6420-9.
\newblock \doi{10.1109/CVPR.2018.00240}.
\newblock URL \url{https://ieeexplore.ieee.org/document/8578338/}.

\bibitem[Gustafsson et~al.(2002)Gustafsson, Gunnarsson, Bergman, Forssell,
  Jansson, Karlsson, and Nordlund]{gustafsson_particle_2002}
F.~Gustafsson, F.~Gunnarsson, N.~Bergman, U.~Forssell, J.~Jansson, R.~Karlsson,
  and P.-J. Nordlund.
\newblock Particle filters for positioning, navigation, and tracking.
\newblock \emph{IEEE Transactions on Signal Processing}, 50\penalty0
  (2):\penalty0 425--437, February 2002.
\newblock ISSN 1941-0476.
\newblock \doi{10.1109/78.978396}.

\bibitem[Gyulai et~al.(2020)Gyulai, Pfeiffer, and
  Bergmann]{gyulai_analysis_2020}
Dávid Gyulai, András Pfeiffer, and Júlia Bergmann.
\newblock Analysis of asset location data to support decisions in production
  management and control.
\newblock \emph{Procedia CIRP}, 88:\penalty0 197--202, January 2020.
\newblock ISSN 2212-8271.
\newblock \doi{10.1016/j.procir.2020.05.035}.
\newblock URL
  \url{https://www.sciencedirect.com/science/article/pii/S2212827120303504}.

\bibitem[Güting et~al.(2006)Güting, De~Almeida, and
  Ding]{guting_modeling_2006}
Ralf~Hartmut Güting, Victor~Teixeira De~Almeida, and Zhiming Ding.
\newblock Modeling and querying moving objects in networks.
\newblock \emph{VLDB Journal}, 15\penalty0 (2):\penalty0 165--190, 2006.
\newblock ISSN 10668888.
\newblock \doi{10.1007/s00778-005-0152-x}.
\newblock Publisher: Springer New York.

\bibitem[Güting et~al.(2015)Güting, Valdés, and
  Damiani]{guting_symbolic_2015}
Ralf~Hartmut Güting, Fabio Valdés, and Maria~Luisa Damiani.
\newblock Symbolic trajectories.
\newblock \emph{ACM Transactions on Spatial Algorithms and Systems}, 1\penalty0
  (2), July 2015.
\newblock ISSN 23740361.
\newblock \doi{10.1145/2786756}.

\bibitem[Halpern et~al.(2001)Halpern, Harper, Immerman, Kolaitis, Vardi, and
  Vianu]{halpern_unusual_2001}
Joseph~Y. Halpern, Robert Harper, Neil Immerman, Phokion~G. Kolaitis, Moshe~Y.
  Vardi, and Victor Vianu.
\newblock On the {Unusual} {Effectiveness} of {Logic} in {Computer} {Science}.
\newblock \emph{Bulletin of Symbolic Logic}, 7\penalty0 (2):\penalty0 213--236,
  March 2001.
\newblock ISSN 1079-8986, 1943-5894.
\newblock \doi{10.2307/2687775}.
\newblock URL
  \url{https://www.cambridge.org/core/journals/bulletin-of-symbolic-logic/article/abs/on-the-unusual-effectiveness-of-logic-in-computer-science/64C8A4DE3D8E95FF54C970310A1F0A8E}.

\bibitem[Han et~al.(2014)Han, Tucker, Simpson, and Davidson]{han_data_2014}
Yixiang Han, Conrad~S. Tucker, Timothy~W. Simpson, and Erik Davidson.
\newblock A {Data} {Mining} {Trajectory} {Clustering} {Methodology} for
  {Modeling} {Indoor} {Design} {Space} {Utilization}.
\newblock In \emph{{ASME} 2013 {International} {Design} {Engineering}
  {Technical} {Conferences} and {Computers} and {Information} in {Engineering}
  {Conference}}. American Society of Mechanical Engineers Digital Collection,
  February 2014.
\newblock \doi{10.1115/DETC2013-12690}.
\newblock URL
  \url{https://biomechanical.asmedigitalcollection.asme.org/IDETC-CIE/proceedings/IDETC-CIE2013/55898/V03BT03A017/253862}.

\bibitem[Harding et~al.(2005)Harding, Shahbaz, {Srinivas}, and
  Kusiak]{harding_data_2005}
J.~A. Harding, M.~Shahbaz, {Srinivas}, and A.~Kusiak.
\newblock Data {Mining} in {Manufacturing}: {A} {Review}.
\newblock \emph{Journal of Manufacturing Science and Engineering}, 128\penalty0
  (4):\penalty0 969--976, December 2005.
\newblock ISSN 1087-1357.
\newblock \doi{10.1115/1.2194554}.
\newblock URL \url{https://doi.org/10.1115/1.2194554}.

\bibitem[Hauge et~al.(2020)Hauge, Zafarzadeh, Jeong, Li, Khilji, and
  Wiktorsson]{hauge_employing_2020}
Jannicke~Baalsrud Hauge, Masoud Zafarzadeh, Yongkuk Jeong, Yi~Li, Wajid~Ali
  Khilji, and Magnus Wiktorsson.
\newblock Employing digital twins within production logistics.
\newblock In \emph{2020 {IEEE} {International} {Conference} on {Engineering},
  {Technology} and {Innovation} ({ICE}/{ITMC})}, pages 1--8, Cardiff, UK, June
  2020. IEEE.
\newblock ISBN 978-1-72817-037-4.
\newblock \doi{10.1109/ICE/ITMC49519.2020.9198540}.
\newblock URL \url{https://ieeexplore.ieee.org/document/9198540/}.

\bibitem[Herr et~al.(2019)Herr, Grund, and Ertl]{herr_bluecollar_2019}
Dominik Herr, S.~Grund, and T.~Ertl.
\newblock {BlueCollar}: {Optimizing} {Worker} {Paths} on {Factory} {Shop}
  {Floors} with {Visual} {Analytics}.
\newblock In \emph{{HICSS}}, 2019.
\newblock \doi{10.24251/HICSS.2019.191}.

\bibitem[Hu et~al.(2020)Hu, Gao, Li, Wu, Du, and Maybank]{hu_anomaly_2020}
Weiming Hu, Jun Gao, Bing Li, Ou~Wu, Junping Du, and Stephen Maybank.
\newblock Anomaly {Detection} {Using} {Local} {Kernel} {Density} {Estimation}
  and {Context}-{Based} {Regression}.
\newblock \emph{IEEE Transactions on Knowledge and Data Engineering},
  32\penalty0 (2):\penalty0 218--233, February 2020.
\newblock ISSN 1041-4347, 1558-2191, 2326-3865.
\newblock \doi{10.1109/TKDE.2018.2882404}.
\newblock URL \url{https://ieeexplore.ieee.org/document/8540843/}.

\bibitem[Hu et~al.(2016)Hu, Ma, Liu, Hovy, and Xing]{hu_harnessing_2016}
Zhiting Hu, Xuezhe Ma, Zhengzhong Liu, Eduard Hovy, and Eric Xing.
\newblock Harnessing {Deep} {Neural} {Networks} with {Logic} {Rules}.
\newblock In \emph{Proceedings of the 54th {Annual} {Meeting} of the
  {Association} for {Computational} {Linguistics} ({Volume} 1: {Long}
  {Papers})}, pages 2410--2420, Berlin, Germany, 2016. Association for
  Computational Linguistics.
\newblock \doi{10.18653/v1/P16-1228}.
\newblock URL \url{http://aclweb.org/anthology/P16-1228}.

\bibitem[{iMonitor}(2022)]{imonitor_smart_2022}
{iMonitor}.
\newblock Smart manufacturing platform, 2022.
\newblock URL
  \url{https://www.imonitor.net/?gclid=EAIaIQobChMI46iV8pnd9QIVlQsrCh2kbQDsEAAYBCAAEgLhu_D_BwE}.

\bibitem[Ji and Wang(2017)]{ji_big_2017}
Wei Ji and Lihui Wang.
\newblock Big data analytics based fault prediction for shop floor scheduling.
\newblock \emph{Journal of Manufacturing Systems}, 43:\penalty0 187--194, April
  2017.
\newblock ISSN 0278-6125.
\newblock \doi{10.1016/j.jmsy.2017.03.008}.
\newblock URL
  \url{https://www.sciencedirect.com/science/article/pii/S0278612517300389}.

\bibitem[{Jing Yuan} et~al.(2013){Jing Yuan}, {Yu Zheng}, {Xing Xie}, and
  {Guangzhong Sun}]{jing_yuan_t-drive_2013}
{Jing Yuan}, {Yu Zheng}, {Xing Xie}, and {Guangzhong Sun}.
\newblock T-{Drive}: {Enhancing} {Driving} {Directions} with {Taxi} {Drivers}'
  {Intelligence}.
\newblock \emph{IEEE Transactions on Knowledge and Data Engineering},
  25\penalty0 (1):\penalty0 220--232, January 2013.
\newblock ISSN 1041-4347.
\newblock \doi{10.1109/TKDE.2011.200}.
\newblock URL \url{http://ieeexplore.ieee.org/document/6025355/}.

\bibitem[{Kalman R.E}(1960)]{kalman_re_new_1960}
{Kalman R.E}.
\newblock A {New} {Approach} to {Linear} {Filtering} and {Prediction}
  {Problems}.
\newblock \emph{Journal of Basic Engineering}, 82:\penalty0 35--45, 1960.

\bibitem[Kanduč and Rodič(2015)]{kanduc_optimization_2015}
Tadej Kanduč and Blaž Rodič.
\newblock Optimization of a furniture factory layout.
\newblock \emph{Croatian Operational Research Review}, 6\penalty0 (1):\penalty0
  121--130, March 2015.
\newblock ISSN 18480225, 18489931.
\newblock \doi{10.17535/crorr.2015.0010}.
\newblock URL
  \url{http://hrcak.srce.hr/index.php?show=clanak&id_clanak_jezik=204318&lang=en}.

\bibitem[Karve et~al.(2020)Karve, Guo, Kapusuzoglu, Mahadevan, and
  Haile]{karve_digital_2020}
Pranav~M. Karve, Yulin Guo, Berkcan Kapusuzoglu, Sankaran Mahadevan, and
  Mulugeta~A. Haile.
\newblock Digital twin approach for damage-tolerant mission planning under
  uncertainty.
\newblock \emph{Engineering Fracture Mechanics}, 225:\penalty0 106766, February
  2020.
\newblock ISSN 00137944.
\newblock \doi{10.1016/j.engfracmech.2019.106766}.
\newblock URL
  \url{https://linkinghub.elsevier.com/retrieve/pii/S0013794419306496}.

\bibitem[Katzouris et~al.(2021)Katzouris, Paliouras, and
  Artikis]{katzouris_online_2021}
Nikos Katzouris, Georgios Paliouras, and Alexander Artikis.
\newblock Online {Learning} {Probabilistic} {Event} {Calculus} {Theories} in
  {Answer} {Set} {Programming}.
\newblock \emph{Theory and Practice of Logic Programming}, pages 1--25, August
  2021.
\newblock ISSN 1471-0684, 1475-3081.
\newblock \doi{10.1017/S1471068421000107}.
\newblock URL
  \url{https://www.cambridge.org/core/journals/theory-and-practice-of-logic-programming/article/abs/online-learning-probabilistic-event-calculus-theories-in-answer-set-programming/57E24EBFE1CFBBD5CF10CD7EBDD5F848}.

\bibitem[Kingma and Welling(2019)]{kingma_introduction_2019}
Diederik~P. Kingma and Max Welling.
\newblock An {Introduction} to {Variational} {Autoencoders}.
\newblock \emph{Foundations and Trends® in Machine Learning}, 12\penalty0
  (4):\penalty0 307--392, 2019.
\newblock ISSN 1935-8237, 1935-8245.
\newblock \doi{10.1561/2200000056}.
\newblock URL \url{http://arxiv.org/abs/1906.02691}.
\newblock arXiv: 1906.02691.

\bibitem[Kitaev et~al.(2020)Kitaev, Kaiser, and Levskaya]{kitaev_reformer_2020}
Nikita Kitaev, Łukasz Kaiser, and Anselm Levskaya.
\newblock Reformer: {The} {Efficient} {Transformer}.
\newblock \emph{arXiv:2001.04451 [cs, stat]}, February 2020.
\newblock URL \url{http://arxiv.org/abs/2001.04451}.
\newblock arXiv: 2001.04451.

\bibitem[Koller and Friedman(2009)]{koller_probabilistic_2009}
Daphne Koller and Nir Friedman.
\newblock \emph{Probabilistic graphical models: principles and techniques}.
\newblock Adaptive computation and machine learning. MIT Press, Cambridge, MA,
  2009.
\newblock ISBN 978-0-262-01319-2.

\bibitem[Kosaraju et~al.(2019)Kosaraju, Sadeghian, Martín-Martín, Reid,
  Rezatofighi, and Savarese]{kosaraju_social-bigat_2019}
Vineet Kosaraju, Amir Sadeghian, Roberto Martín-Martín, Ian Reid, Hamid
  Rezatofighi, and Silvio Savarese.
\newblock Social-{BiGAT}: {Multimodal} {Trajectory} {Forecasting} using
  {Bicycle}-{GAN} and {Graph} {Attention} {Networks}.
\newblock In \emph{Advances in {Neural} {Information} {Processing} {Systems}},
  volume~32. Curran Associates, Inc., 2019.
\newblock URL
  \url{https://proceedings.neurips.cc/paper/2019/hash/d09bf41544a3365a46c9077ebb5e35c3-Abstract.html}.

\bibitem[Kowalski(1979)]{kowalski_algorithm_1979}
Robert Kowalski.
\newblock Algorithm = logic + control.
\newblock \emph{Communications of the ACM}, 22\penalty0 (7):\penalty0 424--436,
  July 1979.
\newblock ISSN 0001-0782.
\newblock \doi{10.1145/359131.359136}.
\newblock URL \url{https://doi.org/10.1145/359131.359136}.

\bibitem[Kowalski and Sergot(1986)]{kowalski_logic-based_1986}
Robert Kowalski and Marek Sergot.
\newblock A logic-based calculus of events.
\newblock \emph{New Generation Computing}, 4\penalty0 (1):\penalty0 67--95,
  March 1986.
\newblock ISSN 0288-3635, 1882-7055.
\newblock \doi{10.1007/BF03037383}.
\newblock URL \url{http://link.springer.com/10.1007/BF03037383}.

\bibitem[Kritzinger et~al.(2018)Kritzinger, Karner, Traar, Henjes, and
  Sihn]{kritzinger_digital_2018}
Werner Kritzinger, Matthias Karner, Georg Traar, Jan Henjes, and Wilfried Sihn.
\newblock Digital {Twin} in manufacturing: {A} categorical literature review
  and classification.
\newblock \emph{IFAC-PapersOnLine}, 51\penalty0 (11):\penalty0 1016--1022,
  2018.
\newblock ISSN 24058963.
\newblock \doi{10.1016/j.ifacol.2018.08.474}.
\newblock URL
  \url{https://linkinghub.elsevier.com/retrieve/pii/S2405896318316021}.

\bibitem[Launchbury(2017)]{launchbury_darpa_2017}
John Launchbury.
\newblock A {DARPA} {Perspective} on {Artificial} {Intelligence}, 2017.
\newblock URL \url{https://www.youtube.com/watch?v=-O01G3tSYpU}.

\bibitem[Lecun et~al.(1998)Lecun, Bottou, Bengio, and
  Haffner]{lecun_gradient-based_1998}
Y.~Lecun, L.~Bottou, Y.~Bengio, and P.~Haffner.
\newblock Gradient-based learning applied to document recognition.
\newblock \emph{Proceedings of the IEEE}, 86\penalty0 (11):\penalty0
  2278--2324, November 1998.
\newblock ISSN 1558-2256.
\newblock \doi{10.1109/5.726791}.
\newblock Conference Name: Proceedings of the IEEE.

\bibitem[Lee et~al.(2007)Lee, Han, and Whang]{lee_trajectory_2007}
Jae-Gil Lee, Jiawei Han, and Kyu-Young Whang.
\newblock Trajectory clustering: a partition-and-group framework.
\newblock In \emph{Proceedings of the 2007 {ACM} {SIGMOD} international
  conference on {Management} of data}, {SIGMOD} '07, pages 593--604, New York,
  NY, USA, June 2007. Association for Computing Machinery.
\newblock ISBN 978-1-59593-686-8.
\newblock \doi{10.1145/1247480.1247546}.
\newblock URL \url{https://doi.org/10.1145/1247480.1247546}.

\bibitem[Lee et~al.(2013)Lee, Lapira, Bagheri, and Kao]{lee_recent_2013}
Jay Lee, Edzel Lapira, Behrad Bagheri, and Hung-an Kao.
\newblock Recent advances and trends in predictive manufacturing systems in big
  data environment.
\newblock \emph{Manufacturing Letters}, 1\penalty0 (1):\penalty0 38--41,
  October 2013.
\newblock ISSN 22138463.
\newblock \doi{10.1016/j.mfglet.2013.09.005}.
\newblock URL
  \url{https://linkinghub.elsevier.com/retrieve/pii/S2213846313000114}.

\bibitem[Lee et~al.(2011)Lee, Han, and Yang]{lee_construction_2011-1}
Jonghwan Lee, Soonhung Han, and Jeongsam Yang.
\newblock Construction of a computer-simulated mixed reality environment for
  virtual factory layout planning.
\newblock \emph{Computers in Industry}, 62\penalty0 (1):\penalty0 86--98,
  January 2011.
\newblock ISSN 01663615.
\newblock \doi{10.1016/j.compind.2010.07.001}.
\newblock URL
  \url{https://linkinghub.elsevier.com/retrieve/pii/S016636151000093X}.

\bibitem[Lee and Kim(2022)]{lee_reinforcement_2022}
Jun-Ho Lee and Hyun-Jung Kim.
\newblock Reinforcement learning for robotic flow shop scheduling with
  processing time variations.
\newblock \emph{International Journal of Production Research}, 60\penalty0
  (7):\penalty0 2346--2368, April 2022.
\newblock ISSN 0020-7543.
\newblock \doi{10.1080/00207543.2021.1887533}.
\newblock URL \url{https://doi.org/10.1080/00207543.2021.1887533}.
\newblock Publisher: Taylor \& Francis \_eprint:
  https://doi.org/10.1080/00207543.2021.1887533.

\bibitem[Lee and Krumm(2011)]{lee_trajectory_2011}
Wang-Chien Lee and John Krumm.
\newblock Trajectory {Preprocessing}.
\newblock In Yu~Zheng and Xiaofang Zhou, editors, \emph{Computing with
  {Spatial} {Trajectories}}, pages 3--33. Springer, New York, NY, 2011.
\newblock ISBN 978-1-4614-1629-6.
\newblock \doi{10.1007/978-1-4614-1629-6_1}.
\newblock URL \url{https://doi.org/10.1007/978-1-4614-1629-6_1}.

\bibitem[Lei(2016)]{lei_framework_2016}
Po-Ruey Lei.
\newblock A framework for anomaly detection in maritime trajectory behavior.
\newblock \emph{Knowledge and Information Systems}, 47\penalty0 (1):\penalty0
  189--214, April 2016.
\newblock ISSN 0219-1377, 0219-3116.
\newblock \doi{10.1007/s10115-015-0845-4}.
\newblock URL \url{http://link.springer.com/10.1007/s10115-015-0845-4}.

\bibitem[Leng et~al.(2021)Leng, Wang, Shen, Li, Liu, and
  Chen]{leng_digital_2021}
Jiewu Leng, Dewen Wang, Weiming Shen, Xinyu Li, Qiang Liu, and Xin Chen.
\newblock Digital twins-based smart manufacturing system design in {Industry}
  4.0: {A} review.
\newblock \emph{Journal of Manufacturing Systems}, 60:\penalty0 119--137, July
  2021.
\newblock ISSN 02786125.
\newblock \doi{10.1016/j.jmsy.2021.05.011}.
\newblock URL
  \url{https://linkinghub.elsevier.com/retrieve/pii/S0278612521001151}.

\bibitem[Levy(2000)]{levy_logic-based_2000}
Alon~Y. Levy.
\newblock Logic-{Based} {Techniques} in {Data} {Integration}.
\newblock In Jack Minker, editor, \emph{Logic-{Based} {Artificial}
  {Intelligence}}, The {Springer} {International} {Series} in {Engineering} and
  {Computer} {Science}, pages 575--595. Springer US, Boston, MA, 2000.
\newblock ISBN 978-1-4615-1567-8.
\newblock \doi{10.1007/978-1-4615-1567-8_24}.
\newblock URL \url{https://doi.org/10.1007/978-1-4615-1567-8_24}.

\bibitem[Li et~al.(2017{\natexlab{a}})Li, Mahadevan, Ling, Choze, and
  Wang]{li_dynamic_2017}
Chenzhao Li, Sankaran Mahadevan, You Ling, Sergio Choze, and Liping Wang.
\newblock Dynamic {Bayesian} {Network} for {Aircraft} {Wing} {Health}
  {Monitoring} {Digital} {Twin}.
\newblock \emph{AIAA Journal}, 55\penalty0 (3):\penalty0 930--941, March
  2017{\natexlab{a}}.
\newblock ISSN 0001-1452, 1533-385X.
\newblock \doi{10.2514/1.J055201}.
\newblock URL \url{https://arc.aiaa.org/doi/10.2514/1.J055201}.

\bibitem[Li et~al.(2016)Li, Chan, Wong, and Skitmore]{li_real-time_2016-1}
Heng Li, Greg Chan, Johnny Kwok~Wai Wong, and Martin Skitmore.
\newblock Real-time locating systems applications in construction.
\newblock \emph{Automation in Construction}, 63:\penalty0 37--47, March 2016.
\newblock ISSN 09265805.
\newblock \doi{10.1016/j.autcon.2015.12.001}.
\newblock URL
  \url{https://linkinghub.elsevier.com/retrieve/pii/S0926580515002411}.

\bibitem[Li et~al.(2020{\natexlab{a}})Li, Lu, Chen, Chen, Chen, and
  Shou]{li_toward_2020}
Huan Li, Hua Lu, Gang Chen, Ke~Chen, Qinkuang Chen, and Lidan Shou.
\newblock Toward {Translating} {Raw} {Indoor} {Positioning} {Data} into
  {Mobility} {Semantics}.
\newblock \emph{ACM/IMS Transactions on Data Science}, 1\penalty0 (4):\penalty0
  1--37, November 2020{\natexlab{a}}.
\newblock ISSN 2691-1922.
\newblock \doi{10.1145/3385190}.
\newblock Publisher: Association for Computing Machinery (ACM).

\bibitem[Li et~al.(2019)Li, Vardi, and Rozier]{li_satisfiability_2019}
Jianwen Li, Moshe~Y. Vardi, and Kristin~Y. Rozier.
\newblock Satisfiability {Checking} for {Mission}-{Time} {LTL}.
\newblock In Isil Dillig and Serdar Tasiran, editors, \emph{Computer {Aided}
  {Verification}}, Lecture {Notes} in {Computer} {Science}, pages 3--22, Cham,
  2019. Springer International Publishing.
\newblock ISBN 978-3-030-25543-5.
\newblock \doi{10.1007/978-3-030-25543-5_1}.

\bibitem[Li and Srikumar(2019)]{li_augmenting_2019}
Tao Li and Vivek Srikumar.
\newblock Augmenting {Neural} {Networks} with {First}-order {Logic}.
\newblock \emph{ACL 2019 - 57th Annual Meeting of the Association for
  Computational Linguistics, Proceedings of the Conference}, pages 292--302,
  June 2019.
\newblock \doi{10.18653/v1/p19-1028}.
\newblock URL \url{https://arxiv.org/abs/1906.06298v3}.
\newblock arXiv: 1906.06298 Publisher: Association for Computational
  Linguistics (ACL) ISBN: 9781950737482.

\bibitem[Li et~al.(2020{\natexlab{b}})Li, Liu, Sun, Chen, Zhang, and
  Sun]{li_spatio-temporal_2020}
Tengfei Li, Jing Liu, Haiying Sun, Xiang Chen, Lipeng Zhang, and Junfeng Sun.
\newblock A spatio-temporal specification language and its completeness \&
  decidability.
\newblock \emph{Journal of Cloud Computing}, 9\penalty0 (1):\penalty0 65,
  December 2020{\natexlab{b}}.
\newblock ISSN 2192-113X.
\newblock \doi{10.1186/s13677-020-00209-3}.
\newblock URL
  \url{https://journalofcloudcomputing.springeropen.com/articles/10.1186/s13677-020-00209-3}.

\bibitem[Li et~al.(2017{\natexlab{b}})Li, Vasile, and
  Belta]{li_reinforcement_2017}
Xiao Li, Cristian-Ioan Vasile, and Calin Belta.
\newblock Reinforcement learning with temporal logic rewards.
\newblock In \emph{2017 {IEEE}/{RSJ} {International} {Conference} on
  {Intelligent} {Robots} and {Systems} ({IROS})}, pages 3834--3839, September
  2017{\natexlab{b}}.
\newblock \doi{10.1109/IROS.2017.8206234}.
\newblock ISSN: 2153-0866.

\bibitem[Li et~al.(2018)Li, Zhao, Cong, Jensen, and Wei]{li_deep_2018}
Xiucheng Li, Kaiqi Zhao, Gao Cong, Christian~S. Jensen, and Wei Wei.
\newblock Deep {Representation} {Learning} for {Trajectory} {Similarity}
  {Computation}.
\newblock In \emph{2018 {IEEE} 34th {International} {Conference} on {Data}
  {Engineering} ({ICDE})}, pages 617--628, April 2018.
\newblock \doi{10.1109/ICDE.2018.00062}.
\newblock ISSN: 2375-026X.

\bibitem[Li et~al.(2017{\natexlab{c}})Li, Cui, Chen, Wu, and
  Zhang]{li_mlog_2017}
Xupeng Li, Bin Cui, Yiru Chen, Wentao Wu, and Ce~Zhang.
\newblock {MLog}: towards declarative in-database machine learning.
\newblock \emph{Proceedings of the VLDB Endowment}, 10\penalty0 (12):\penalty0
  1933--1936, August 2017{\natexlab{c}}.
\newblock ISSN 2150-8097.
\newblock \doi{10.14778/3137765.3137812}.
\newblock URL \url{https://doi.org/10.14778/3137765.3137812}.

\bibitem[Li(2018)]{li_deep_overview_2018}
Yuxi Li.
\newblock Deep {Reinforcement} {Learning}.
\newblock Technical Report arXiv:1810.06339, arXiv, October 2018.
\newblock URL \url{http://arxiv.org/abs/1810.06339}.
\newblock arXiv:1810.06339 [cs, stat] type: article.

\bibitem[Liao(2020)]{liao_survey_2020}
Hsuan-Cheng Liao.
\newblock A {Survey} of {Reinforcement} {Learning} with {Temporal} {Logic}
  {Rewards}.
\newblock TUM, 2020.
\newblock URL \url{https://mediatum.ub.tum.de/doc/1579215/1579215.pdf}.

\bibitem[Limketkai et~al.(2005)Limketkai, Liao, and
  Fox]{limketkai_relational_2005}
Benson Limketkai, Lin Liao, and Dieter Fox.
\newblock Relational object maps for mobile robots.
\newblock In \emph{Proceedings of the 19th international joint conference on
  {Artificial} intelligence}, {IJCAI}'05, pages 1471--1476, San Francisco, CA,
  USA, July 2005. Morgan Kaufmann Publishers Inc.

\bibitem[Liu et~al.(2021)Liu, Fang, Dong, and Xu]{liu_review_2021}
Mengnan Liu, Shuiliang Fang, Huiyue Dong, and Cunzhi Xu.
\newblock Review of digital twin about concepts, technologies, and industrial
  applications.
\newblock \emph{Journal of Manufacturing Systems}, 58:\penalty0 346--361,
  January 2021.
\newblock ISSN 02786125.
\newblock \doi{10.1016/j.jmsy.2020.06.017}.
\newblock URL
  \url{https://linkinghub.elsevier.com/retrieve/pii/S0278612520301072}.

\bibitem[Liu et~al.(2018{\natexlab{a}})Liu, Xie, Yu, and Hu]{liu_survey_2018}
Ran Liu, Xiaolei Xie, Kaiye Yu, and Qiaoyu Hu.
\newblock A survey on simulation optimization for the manufacturing system
  operation.
\newblock \emph{International Journal of Modelling and Simulation}, 38\penalty0
  (2):\penalty0 116--127, April 2018{\natexlab{a}}.
\newblock ISSN 0228-6203, 1925-7082.
\newblock \doi{10.1080/02286203.2017.1401418}.
\newblock URL
  \url{https://www.tandfonline.com/doi/full/10.1080/02286203.2017.1401418}.

\bibitem[Liu and Tang(2021)]{liu_network_2021}
Xueyi Liu and Jie Tang.
\newblock Network representation learning: {A} macro and micro view.
\newblock \emph{AI Open}, 2:\penalty0 43--64, 2021.
\newblock ISSN 26666510.
\newblock \doi{10.1016/j.aiopen.2021.02.001}.
\newblock URL
  \url{https://linkinghub.elsevier.com/retrieve/pii/S2666651021000024}.

\bibitem[Liu et~al.(2012)Liu, Zhao, Chen, Pei, and Han]{liu_mining_2012}
Yunhao Liu, Yiyang Zhao, Lei Chen, Jian Pei, and Jinsong Han.
\newblock Mining {Frequent} {Trajectory} {Patterns} for {Activity} {Monitoring}
  {Using} {Radio} {Frequency} {Tag} {Arrays}.
\newblock \emph{IEEE Transactions on Parallel and Distributed Systems},
  23\penalty0 (11):\penalty0 2138--2149, November 2012.
\newblock ISSN 1558-2183.
\newblock \doi{10.1109/TPDS.2011.307}.
\newblock Conference Name: IEEE Transactions on Parallel and Distributed
  Systems.

\bibitem[Liu et~al.(2018{\natexlab{b}})Liu, Meyendorf, and Mrad]{liu_role_2018}
Zheng Liu, Norbert Meyendorf, and Nezih Mrad.
\newblock The role of data fusion in predictive maintenance using digital twin.
\newblock \emph{AIP Conference Proceedings}, 1949\penalty0 (1):\penalty0
  020023, April 2018{\natexlab{b}}.
\newblock ISSN 0094-243X.
\newblock \doi{10.1063/1.5031520}.
\newblock URL \url{https://aip.scitation.org/doi/abs/10.1063/1.5031520}.
\newblock Publisher: American Institute of Physics.

\bibitem[Ljunggren(2018)]{ljunggren_using_2018}
Henrik Ljunggren.
\newblock Using {Deep} {Learning} for {Classifying} {Ship} {Trajectories}.
\newblock In \emph{2018 21st {International} {Conference} on {Information}
  {Fusion} ({FUSION})}, pages 2158--2164, July 2018.
\newblock \doi{10.23919/ICIF.2018.8455776}.

\bibitem[Luna et~al.(2015)Luna, Lahijanian, Moll, and
  Kavraki]{akin_asymptotically_2015}
Ryan Luna, Morteza Lahijanian, Mark Moll, and Lydia~E. Kavraki.
\newblock Asymptotically {Optimal} {Stochastic} {Motion} {Planning} with
  {Temporal} {Goals}.
\newblock In H.~Levent Akin, Nancy~M. Amato, Volkan Isler, and A.~Frank van~der
  Stappen, editors, \emph{Algorithmic {Foundations} of {Robotics} {XI}}, volume
  107, pages 335--352. Springer International Publishing, Cham, 2015.
\newblock ISBN 978-3-319-16594-3 978-3-319-16595-0.
\newblock \doi{10.1007/978-3-319-16595-0_20}.
\newblock URL \url{http://link.springer.com/10.1007/978-3-319-16595-0_20}.
\newblock Series Title: Springer Tracts in Advanced Robotics.

\bibitem[Lutz and Miličić(2007)]{lutz_tableau_2007}
Carsten Lutz and Maja Miličić.
\newblock A {Tableau} {Algorithm} for {Description} {Logics} with {Concrete}
  {Domains} and {General} {TBoxes}.
\newblock \emph{Journal of Automated Reasoning}, 38\penalty0 (1-3):\penalty0
  227--259, February 2007.
\newblock ISSN 0168-7433, 1573-0670.
\newblock \doi{10.1007/s10817-006-9049-7}.
\newblock URL \url{http://link.springer.com/10.1007/s10817-006-9049-7}.

\bibitem[Lutz et~al.(2008)Lutz, Wolter, and Zakharyaschev]{lutz_temporal_2008}
Carsten Lutz, Frank Wolter, and Michael Zakharyaschev.
\newblock Temporal {Description} {Logics}: {A} {Survey}.
\newblock In \emph{2008 15th {International} {Symposium} on {Temporal}
  {Representation} and {Reasoning}}, pages 3--14, Montreal, QC, June 2008.
  IEEE.
\newblock ISBN 978-0-7695-3181-6.
\newblock \doi{10.1109/TIME.2008.14}.
\newblock URL \url{https://ieeexplore.ieee.org/document/4553284/}.

\bibitem[Lv et~al.(2018)Lv, Li, Sun, and Wang]{lv_t-conv_2018}
Jianming Lv, Qing Li, Qinghui Sun, and Xintong Wang.
\newblock T-{CONV}: {A} {Convolutional} {Neural} {Network} for {Multi}-scale
  {Taxi} {Trajectory} {Prediction}.
\newblock In \emph{2018 {IEEE} {International} {Conference} on {Big} {Data} and
  {Smart} {Computing} ({BigComp})}, pages 82--89, January 2018.
\newblock \doi{10.1109/BigComp.2018.00021}.
\newblock ISSN: 2375-9356.

\bibitem[Löcklin et~al.(2020)Löcklin, Ruppert, Jakab, Libert, Jazdi, and
  Weyrich]{locklin_trajectory_2020}
Andreas Löcklin, Tamás Ruppert, László Jakab, Robert Libert, Nasser Jazdi,
  and Michael Weyrich.
\newblock Trajectory {Prediction} of {Humans} in {Factories} and {Warehouses}
  with {Real}-{Time} {Locating} {Systems}.
\newblock In \emph{2020 25th {IEEE} {International} {Conference} on {Emerging}
  {Technologies} and {Factory} {Automation} ({ETFA})}, volume~1, pages
  1317--1320, September 2020.
\newblock \doi{10.1109/ETFA46521.2020.9211913}.
\newblock ISSN: 1946-0759.

\bibitem[Macchi et~al.(2018)Macchi, Roda, Negri, and
  Fumagalli]{macchi_exploring_2018}
Marco Macchi, Irene Roda, Elisa Negri, and Luca Fumagalli.
\newblock Exploring the role of {Digital} {Twin} for {Asset} {Lifecycle}
  {Management}.
\newblock \emph{IFAC-PapersOnLine}, 51\penalty0 (11):\penalty0 790--795, 2018.
\newblock ISSN 24058963.
\newblock \doi{10.1016/j.ifacol.2018.08.415}.
\newblock URL
  \url{https://linkinghub.elsevier.com/retrieve/pii/S2405896318315416}.

\bibitem[{MachineMetrics}(2022)]{machinemetrics_elminate_2022}
{MachineMetrics}.
\newblock Elminate guesswork with real-time production tracking, 2022.
\newblock URL
  \url{https://www.machinemetrics.com/production-tracking-software}.

\bibitem[Manhaeve et~al.(2021)Manhaeve, Dumančić, Kimmig, Demeester, and
  De~Raedt]{manhaeve_neural_2021}
Robin Manhaeve, Sebastijan Dumančić, Angelika Kimmig, Thomas Demeester, and
  Luc De~Raedt.
\newblock Neural probabilistic logic programming in {DeepProbLog}.
\newblock \emph{Artificial Intelligence}, 298:\penalty0 103504, September 2021.
\newblock ISSN 00043702.
\newblock \doi{10.1016/j.artint.2021.103504}.
\newblock URL
  \url{https://linkinghub.elsevier.com/retrieve/pii/S0004370221000552}.

\bibitem[Mantenoglou et~al.(2020)Mantenoglou, Artikis, and
  Paliouras]{mantenoglou_online_2020}
Periklis Mantenoglou, Alexander Artikis, and Georgios Paliouras.
\newblock Online {Probabilistic} {Interval}-{Based} {Event} {Calculus}.
\newblock In \emph{{ECAI} 2020}, pages 2624--2631. IOS Press, 2020.
\newblock \doi{10.3233/FAIA200399}.
\newblock URL \url{https://ebooks.iospress.nl/doi/10.3233/FAIA200399}.

\bibitem[Marcus(2020)]{marcus_next_2020}
Gary Marcus.
\newblock The {Next} {Decade} in {AI}: {Four} {Steps} {Towards} {Robust}
  {Artificial} {Intelligence}.
\newblock Technical Report arXiv:2002.06177, arXiv, February 2020.
\newblock URL \url{http://arxiv.org/abs/2002.06177}.
\newblock arXiv:2002.06177 [cs] type: article.

\bibitem[Markosian et~al.(2007)Markosian, Mansouri-Samani, Mehlitz, and
  Pressburger]{markosian_program_2007}
Lawrence~Z. Markosian, Masoud Mansouri-Samani, Peter~C. Mehlitz, and Tom
  Pressburger.
\newblock Program {Model} {Checking} {Using} {Design}-for-{Verification}:
  {NASA} {Flight} {Software} {Case} {Study}.
\newblock In \emph{2007 {IEEE} {Aerospace} {Conference}}, pages 1--9, March
  2007.
\newblock \doi{10.1109/AERO.2007.352767}.
\newblock ISSN: 1095-323X.

\bibitem[Mei et~al.(2020)Mei, Qin, Xu, and Eisner]{mei_neural_2020}
Hongyuan Mei, Guanghui Qin, Minjie Xu, and Jason Eisner.
\newblock Neural datalog through time: informed temporal modeling via logical
  specification.
\newblock In \emph{Proceedings of the 37th {International} {Conference} on
  {Machine} {Learning}}, {ICML}'20, pages 6808--6819. JMLR.org, July 2020.

\bibitem[Mo et~al.(2021)Mo, Ma, Koutsopoulos, and Zhao]{mo_calibrating_2021}
Baichuan Mo, Zhenliang Ma, Haris~N. Koutsopoulos, and Jinhua Zhao.
\newblock Calibrating {Path} {Choices} and {Train} {Capacities} for {Urban}
  {Rail} {Transit} {Simulation} {Models} {Using} {Smart} {Card} and {Train}
  {Movement} {Data}.
\newblock \emph{Journal of Advanced Transportation}, 2021:\penalty0 1--15,
  February 2021.
\newblock ISSN 2042-3195, 0197-6729.
\newblock \doi{10.1155/2021/5597130}.
\newblock URL \url{https://www.hindawi.com/journals/jat/2021/5597130/}.

\bibitem[Mohamed et~al.(2017)Mohamed, Aly, and Youssef]{mohamed_accurate_2017}
Reham Mohamed, Heba Aly, and Moustafa Youssef.
\newblock Accurate {Real}-time {Map} {Matching} for {Challenging}
  {Environments}.
\newblock \emph{IEEE Transactions on Intelligent Transportation Systems},
  18\penalty0 (4):\penalty0 847--857, April 2017.
\newblock ISSN 1558-0016.
\newblock \doi{10.1109/TITS.2016.2591958}.
\newblock Conference Name: IEEE Transactions on Intelligent Transportation
  Systems.

\bibitem[{Motion Analysis}(2022)]{motion_analysis_motion_2022}
{Motion Analysis}.
\newblock Motion analysis powered by {Cortex}, 2022.
\newblock URL \url{https://motionanalysis.com/industrial/}.

\bibitem[Muggleton and de~Raedt(1994)]{muggleton_inductive_1994}
Stephen Muggleton and Luc de~Raedt.
\newblock Inductive {Logic} {Programming}: {Theory} and methods.
\newblock \emph{The Journal of Logic Programming}, 19-20:\penalty0 629--679,
  May 1994.
\newblock ISSN 0743-1066.
\newblock \doi{10.1016/0743-1066(94)90035-3}.
\newblock URL
  \url{https://www.sciencedirect.com/science/article/pii/0743106694900353}.

\bibitem[Nagorny et~al.(2017)Nagorny, Lima-Monteiro, Barata, and
  Colombo]{nagorny_big_2017}
Kevin Nagorny, Pedro Lima-Monteiro, Jose Barata, and Armando~Walter Colombo.
\newblock Big {Data} {Analysis} in {Smart} {Manufacturing}: {A} {Review}.
\newblock \emph{International Journal of Communications, Network and System
  Sciences}, 10\penalty0 (3):\penalty0 31--58, March 2017.
\newblock \doi{10.4236/ijcns.2017.103003}.
\newblock URL \url{http://www.scirp.org/Journal/Paperabs.aspx?paperid=75656}.
\newblock Number: 3 Publisher: Scientific Research Publishing.

\bibitem[Natarajan et~al.(2013)Natarajan, Kersting, Ip, Jacobs, and
  Carr]{natarajan_early_2013}
Sriraam Natarajan, Kristian Kersting, Edward Ip, David~R. Jacobs, and Jeffrey
  Carr.
\newblock Early prediction of coronary artery calcification levels using
  machine learning.
\newblock In \emph{Proceedings of the {Twenty}-{Seventh} {AAAI} {Conference} on
  {Artificial} {Intelligence}}, {AAAI}'13, pages 1557--1562, Bellevue,
  Washington, July 2013. AAAI Press.

\bibitem[Nelson and Pecheur(2003)]{nelson_formal_2003}
Stacy~D. Nelson and Charles Pecheur.
\newblock Formal {Verification} for a {Next}-{Generation} {Space} {Shuttle}.
\newblock In Michael~G. Hinchey, James~L. Rash, Walter~F. Truszkowski,
  Christopher Rouff, and Diana Gordon-Spears, editors, \emph{Formal
  {Approaches} to {Agent}-{Based} {Systems}}, Lecture {Notes} in {Computer}
  {Science}, pages 53--67, Berlin, Heidelberg, 2003. Springer.
\newblock ISBN 978-3-540-45133-4.
\newblock \doi{10.1007/978-3-540-45133-4_5}.

\bibitem[Nguyen et~al.(2006)Nguyen, Nguyen, Taylor, and
  Middleton]{nguyen_improved_2006}
Son~T. Nguyen, Hung~T. Nguyen, Philip~B. Taylor, and James Middleton.
\newblock Improved {Head} {Direction} {Command} {Classification} using an
  {Optimised} {Bayesian} {Neural} {Network}.
\newblock In \emph{2006 {International} {Conference} of the {IEEE}
  {Engineering} in {Medicine} and {Biology} {Society}}, pages 5679--5682, New
  York, NY, August 2006. IEEE.
\newblock ISBN 978-1-4244-0032-4.
\newblock \doi{10.1109/IEMBS.2006.260430}.
\newblock URL \url{http://ieeexplore.ieee.org/document/4463095/}.

\bibitem[Nikhil and Morris(2019)]{leal-taixe_convolutional_2019}
Nishant Nikhil and Brendan~Tran Morris.
\newblock Convolutional {Neural} {Network} for {Trajectory} {Prediction}.
\newblock In Laura Leal-Taixé and Stefan Roth, editors, \emph{Computer
  {Vision} – {ECCV} 2018 {Workshops}}, volume 11131, pages 186--196. Springer
  International Publishing, Cham, 2019.
\newblock ISBN 978-3-030-11014-7 978-3-030-11015-4.
\newblock \doi{10.1007/978-3-030-11015-4_16}.
\newblock URL \url{http://link.springer.com/10.1007/978-3-030-11015-4_16}.
\newblock Series Title: Lecture Notes in Computer Science.

\bibitem[Nilsson(1986)]{nilsson_probabilistic_1986}
Nils~J. Nilsson.
\newblock Probabilistic logic.
\newblock \emph{Artificial Intelligence}, 28\penalty0 (1):\penalty0 71--87,
  February 1986.
\newblock ISSN 0004-3702.
\newblock \doi{10.1016/0004-3702(86)90031-7}.
\newblock URL
  \url{https://www.sciencedirect.com/science/article/pii/0004370286900317}.

\bibitem[Nitti et~al.(2014)Nitti, De~Laet, and De~Raedt]{nitti_relational_2014}
Davide Nitti, Tinne De~Laet, and Luc De~Raedt.
\newblock Relational object tracking and learning.
\newblock In \emph{2014 {IEEE} {International} {Conference} on {Robotics} and
  {Automation} ({ICRA})}, pages 935--942, May 2014.
\newblock \doi{10.1109/ICRA.2014.6906966}.
\newblock ISSN: 1050-4729.

\bibitem[Niu et~al.(2014)Niu, Zhu, and Zhang]{niu_deepsense_2014}
Xiaoguang Niu, Ying Zhu, and Xining Zhang.
\newblock {DeepSense}: {A} novel learning mechanism for traffic prediction with
  taxi {GPS} traces.
\newblock In \emph{2014 {IEEE} {Global} {Communications} {Conference}}, pages
  2745--2750, December 2014.
\newblock \doi{10.1109/GLOCOM.2014.7037223}.
\newblock ISSN: 1930-529X.

\bibitem[Oord et~al.(2016)Oord, Dieleman, Zen, Simonyan, Vinyals, Graves,
  Kalchbrenner, Senior, and Kavukcuoglu]{oord_wavenet_2016}
Aäron van~den Oord, S.~Dieleman, H.~Zen, K.~Simonyan, Oriol Vinyals,
  A.~Graves, Nal Kalchbrenner, A.~Senior, and K.~Kavukcuoglu.
\newblock {WaveNet}: {A} {Generative} {Model} for {Raw} {Audio}.
\newblock In \emph{{SSW}}, 2016.

\bibitem[Ouelhadj and Petrovic(2009)]{ouelhadj_survey_2009}
Djamila Ouelhadj and Sanja Petrovic.
\newblock A survey of dynamic scheduling in manufacturing systems.
\newblock \emph{Journal of Scheduling}, 12\penalty0 (4):\penalty0 417--431,
  August 2009.
\newblock ISSN 1094-6136, 1099-1425.
\newblock \doi{10.1007/s10951-008-0090-8}.
\newblock URL \url{http://link.springer.com/10.1007/s10951-008-0090-8}.

\bibitem[Ovacik and Uzsoy(1994)]{ovacik_exploiting_1994}
Irfan~M Ovacik and Reha Uzsoy.
\newblock Exploiting {Shop} {Floor} {Status} {Information} to {Schedule}
  {Complex} {Job} {Shops}.
\newblock \emph{Journal of Manufacturing Systems}, 13\penalty0 (2):\penalty0
  12, 1994.

\bibitem[Ovacik and Uzsoy(1997)]{ovacik_decomposition_1997}
Irfan~M. Ovacik and Reha Uzsoy.
\newblock \emph{Decomposition {Methods} for {Complex} {Factory} {Scheduling}
  {Problems}}.
\newblock Springer US, Boston, MA, 1997.
\newblock ISBN 978-1-4613-7906-5 978-1-4615-6329-7.
\newblock \doi{10.1007/978-1-4615-6329-7}.
\newblock URL \url{http://link.springer.com/10.1007/978-1-4615-6329-7}.

\bibitem[Palmer et~al.(2018)Palmer, Urwin, Niknejad, Petrovic, Popplewell, and
  Young]{palmer_ontology_2018}
Claire Palmer, Esmond~N. Urwin, Ali Niknejad, Dobrila Petrovic, Keith
  Popplewell, and Robert I.~M. Young.
\newblock An ontology supported risk assessment approach for the intelligent
  configuration of supply networks.
\newblock \emph{Journal of Intelligent Manufacturing}, 29\penalty0
  (5):\penalty0 1005--1030, June 2018.
\newblock ISSN 0956-5515, 1572-8145.
\newblock \doi{10.1007/s10845-016-1252-8}.
\newblock URL \url{http://link.springer.com/10.1007/s10845-016-1252-8}.

\bibitem[Panzer and Bender(2021)]{panzer_deep_2021}
Marcel Panzer and Benedict Bender.
\newblock Deep reinforcement learning in production systems: a systematic
  literature review.
\newblock \emph{International Journal of Production Research}, 0\penalty0
  (0):\penalty0 1--26, September 2021.
\newblock ISSN 0020-7543.
\newblock \doi{10.1080/00207543.2021.1973138}.
\newblock URL \url{https://doi.org/10.1080/00207543.2021.1973138}.

\bibitem[Parent et~al.(2013)Parent, Spaccapietra, Renso, Andrienko, Andrienko,
  Bogorny, Damiani, Gkoulalas-Divanis, Macedo, Pelekis, Theodoridis, and
  Yan]{parent_semantic_2013}
Christine Parent, Stefano Spaccapietra, Chiara Renso, Gennady Andrienko,
  Natalia Andrienko, Vania Bogorny, Maria~Luisa Damiani, Aris
  Gkoulalas-Divanis, Jose Macedo, Nikos Pelekis, Yannis Theodoridis, and
  Zhixian Yan.
\newblock Semantic trajectories modeling and analysis.
\newblock \emph{ACM Computing Surveys}, 45\penalty0 (4), August 2013.
\newblock ISSN 03600300.
\newblock \doi{10.1145/2501654.2501656}.

\bibitem[Parisi and Grant(2014)]{parisi_integrity_2014}
Francesco Parisi and John Grant.
\newblock Integrity {Constraints} for {Probabilistic} {Spatio}-{Temporal}
  {Knowledgebases}.
\newblock In Umberto Straccia and Andrea Calì, editors, \emph{Scalable
  {Uncertainty} {Management}}, Lecture {Notes} in {Computer} {Science}, pages
  251--264, Cham, 2014. Springer International Publishing.
\newblock ISBN 978-3-319-11508-5.
\newblock \doi{10.1007/978-3-319-11508-5_21}.

\bibitem[Parisi and Grant(2016)]{parisi_knowledge_2016}
Francesco Parisi and John Grant.
\newblock Knowledge {Representation} in {Probabilistic} {Spatio}-{Temporal}
  {Knowledge} {Bases}.
\newblock \emph{Journal of Artificial Intelligence Research}, 55:\penalty0
  743--798, March 2016.
\newblock ISSN 1076-9757.
\newblock \doi{10.1613/jair.4883}.
\newblock URL \url{https://jair.org/index.php/jair/article/view/10992}.

\bibitem[Park et~al.(2018)Park, Kim, Kang, Chung, and
  Choi]{park_sequence--sequence_2018}
SeongHyeon Park, Byeongdo Kim, C.~Kang, C.~Chung, and J.~Choi.
\newblock Sequence-to-{Sequence} {Prediction} of {Vehicle} {Trajectory} via
  {LSTM} {Encoder}-{Decoder} {Architecture}.
\newblock \emph{2018 IEEE Intelligent Vehicles Symposium (IV)}, 2018.
\newblock \doi{10.1109/IVS.2018.8500658}.

\bibitem[Pearl(1988)]{pearl_probabilistic_1988}
Judea Pearl.
\newblock \emph{Probabilistic {Reasoning} in {Intelligent} {Systems}:
  {Networks} of {Plausible} {Inference}}.
\newblock Morgan Kaufmann Publishers Inc., San Francisco, CA, USA, 1988.
\newblock ISBN 978-1-55860-479-7.

\bibitem[Pellissier~Tanon et~al.(2017)Pellissier~Tanon, Stepanova, Razniewski,
  Mirza, and Weikum]{pellissier_tanon_completeness-aware_2017}
Thomas Pellissier~Tanon, Daria Stepanova, Simon Razniewski, Paramita Mirza, and
  Gerhard Weikum.
\newblock Completeness-{Aware} {Rule} {Learning} from {Knowledge} {Graphs}.
\newblock In Claudia d'Amato, Miriam Fernandez, Valentina Tamma, Freddy Lecue,
  Philippe Cudré-Mauroux, Juan Sequeda, Christoph Lange, and Jeff Heflin,
  editors, \emph{The {Semantic} {Web} – {ISWC} 2017}, Lecture {Notes} in
  {Computer} {Science}, pages 507--525, Cham, 2017. Springer International
  Publishing.
\newblock ISBN 978-3-319-68288-4.
\newblock \doi{10.1007/978-3-319-68288-4_30}.

\bibitem[Pellissier~Tanon et~al.(2020)Pellissier~Tanon, Weikum, and
  Suchanek]{harth_yago_2020}
Thomas Pellissier~Tanon, Gerhard Weikum, and Fabian Suchanek.
\newblock {YAGO} 4: {A} {Reason}-able {Knowledge} {Base}.
\newblock In Andreas Harth, Sabrina Kirrane, Axel-Cyrille Ngonga~Ngomo, Heiko
  Paulheim, Anisa Rula, Anna~Lisa Gentile, Peter Haase, and Michael Cochez,
  editors, \emph{The {Semantic} {Web}}, volume 12123, pages 583--596. Springer
  International Publishing, Cham, 2020.
\newblock ISBN 978-3-030-49460-5 978-3-030-49461-2.
\newblock URL \url{http://link.springer.com/10.1007/978-3-030-49461-2_34}.

\bibitem[Phillips and Tang(2019)]{phillips_simple_2019}
Jeff~M. Phillips and Pingfan Tang.
\newblock Simple {Distances} for {Trajectories} via {Landmarks}.
\newblock In \emph{Proceedings of the 27th {ACM} {SIGSPATIAL} {International}
  {Conference} on {Advances} in {Geographic} {Information} {Systems}}, pages
  468--471, Chicago IL USA, November 2019. ACM.
\newblock ISBN 978-1-4503-6909-1.
\newblock \doi{10.1145/3347146.3359098}.
\newblock URL \url{https://dl.acm.org/doi/10.1145/3347146.3359098}.

\bibitem[Pitsikalis et~al.(2020)Pitsikalis, Bereta, Vodas, Zissis, and
  Artikis]{pitsikalis_event_2020}
Manolis Pitsikalis, Konstantina Bereta, Marios Vodas, Dimitris Zissis, and
  Alexander Artikis.
\newblock Event processing for maritime situational awareness.
\newblock In \emph{Big {Data} {Analytics} for {Time}-{Critical} {Mobility}
  {Forecasting}: {From} {Raw} {Data} to {Trajectory}-{Oriented} {Mobility}
  {Analytics} in the {Aviation} and {Maritime} {Domains}}, pages 255--274.
  Springer International Publishing, January 2020.
\newblock ISBN 978-3-030-45164-6.
\newblock \doi{10.1007/978-3-030-45164-6_9}.

\bibitem[Poole(2003)]{poole_first-order_2003}
David Poole.
\newblock First-order probabilistic inference.
\newblock In \emph{Proceedings of the 18th international joint conference on
  {Artificial} intelligence}, {IJCAI}'03, pages 985--991, San Francisco, CA,
  USA, August 2003. Morgan Kaufmann Publishers Inc.

\bibitem[Prato(2009)]{prato_route_2009}
Carlo~Giacomo Prato.
\newblock Route choice modeling: past, present and future research directions.
\newblock \emph{Journal of Choice Modelling}, 2\penalty0 (1):\penalty0 65--100,
  2009.
\newblock ISSN 17555345.
\newblock \doi{10.1016/S1755-5345(13)70005-8}.
\newblock URL
  \url{https://linkinghub.elsevier.com/retrieve/pii/S1755534513700058}.

\bibitem[Pronost et~al.(2021)Pronost, Mayer, Marche, Camargo, and
  Dupont]{pronost_towards_2021}
Guillaume Pronost, Frederique Mayer, Brunelle Marche, Mauricio Camargo, and
  Laurent Dupont.
\newblock Towards a {Framework} for the {Classification} of {Digital} {Twins}
  and their {Applications}.
\newblock In \emph{2021 {IEEE} {International} {Conference} on {Engineering},
  {Technology} and {Innovation} ({ICE}/{ITMC})}, pages 1--7, Cardiff, United
  Kingdom, June 2021. IEEE.
\newblock ISBN 978-1-66544-963-2.
\newblock \doi{10.1109/ICE/ITMC52061.2021.9570114}.
\newblock URL \url{https://ieeexplore.ieee.org/document/9570114/}.

\bibitem[Quddus et~al.(2007)Quddus, Ochieng, and Noland]{quddus_current_2007}
Mohammed~A. Quddus, Washington~Y. Ochieng, and Robert~B. Noland.
\newblock Current map-matching algorithms for transport applications:
  {State}-of-the art and future research directions.
\newblock \emph{Transportation Research Part C: Emerging Technologies},
  15\penalty0 (5):\penalty0 312--328, October 2007.
\newblock ISSN 0968-090X.
\newblock \doi{10.1016/j.trc.2007.05.002}.
\newblock URL
  \url{https://www.sciencedirect.com/science/article/pii/S0968090X07000265}.

\bibitem[Raedt et~al.(2016)Raedt, Kersting, Natarajan, and
  Poole]{raedt_statistical_2016}
Luc~De Raedt, Kristian Kersting, Sriraam Natarajan, and David Poole.
\newblock \emph{Statistical {Relational} {Artificial} {Intelligence}: {Logic},
  {Probability}, and {Computation}}.
\newblock Morgan and Claypool, March 2016.
\newblock URL
  \url{http://www.morganclaypool.com/doi/10.2200/S00692ED1V01Y201601AIM032}.

\bibitem[Rahman et~al.(2018)Rahman, Smith, Little, Ingham, Greenwood, and
  Bishop-Hurley]{rahman_cattle_2018}
A.~Rahman, D.V. Smith, B.~Little, A.B. Ingham, P.L. Greenwood, and G.J.
  Bishop-Hurley.
\newblock Cattle behaviour classification from collar, halter, and ear tag
  sensors.
\newblock \emph{Information Processing in Agriculture}, 5\penalty0
  (1):\penalty0 124--133, March 2018.
\newblock ISSN 22143173.
\newblock \doi{10.1016/j.inpa.2017.10.001}.
\newblock URL
  \url{https://linkinghub.elsevier.com/retrieve/pii/S2214317317301099}.

\bibitem[Richardson and Domingos(2006)]{richardson_markov_2006}
Matthew Richardson and Pedro Domingos.
\newblock Markov logic networks.
\newblock \emph{Machine Learning}, 62\penalty0 (1-2):\penalty0 107--136,
  February 2006.
\newblock ISSN 0885-6125, 1573-0565.
\newblock \doi{10.1007/s10994-006-5833-1}.
\newblock URL \url{http://link.springer.com/10.1007/s10994-006-5833-1}.

\bibitem[Riguzzi and Swift(2018)]{kifer_survey_2018}
Fabrizio Riguzzi and Theresa Swift.
\newblock A survey of probabilistic logic programming.
\newblock In Michael Kifer and Yanhong~Annie Liu, editors, \emph{Declarative
  {Logic} {Programming}: {Theory}, {Systems}, and {Applications}}, pages
  185--228. ACM, September 2018.
\newblock ISBN 978-1-970001-99-0.
\newblock \doi{10.1145/3191315.3191319}.
\newblock URL \url{https://dl.acm.org/citation.cfm?id=3191319}.

\bibitem[Riguzzi et~al.(2017)Riguzzi, Bellodi, Zese, Cota, and
  Lamma]{riguzzi_survey_2017}
Fabrizio Riguzzi, Elena Bellodi, Riccardo Zese, Giuseppe Cota, and Evelina
  Lamma.
\newblock A survey of lifted inference approaches for probabilistic logic
  programming under the distribution semantics.
\newblock \emph{International Journal of Approximate Reasoning}, 80:\penalty0
  313--333, January 2017.
\newblock ISSN 0888613X.
\newblock \doi{10.1016/j.ijar.2016.10.002}.
\newblock URL
  \url{https://linkinghub.elsevier.com/retrieve/pii/S0888613X16301736}.

\bibitem[Roos et~al.(2017)Roos, Gavin, and Bonnevay]{roos_dynamic_2017}
Jérémy Roos, Gérald Gavin, and Stéphane Bonnevay.
\newblock A dynamic {Bayesian} network approach to forecast short-term urban
  rail passenger flows with incomplete data.
\newblock In \emph{Transportation {Research} {Procedia}}, volume~26, pages
  53--61. Elsevier B.V., 2017.
\newblock \doi{10.1016/j.trpro.2017.07.008}.
\newblock ISSN: 23521465.

\bibitem[Rosen et~al.(2015)Rosen, von Wichert, Lo, and
  Bettenhausen]{rosen_about_2015}
Roland Rosen, Georg von Wichert, George Lo, and Kurt~D. Bettenhausen.
\newblock About {The} {Importance} of {Autonomy} and {Digital} {Twins} for the
  {Future} of {Manufacturing}.
\newblock \emph{IFAC-PapersOnLine}, 48\penalty0 (3):\penalty0 567--572, 2015.
\newblock ISSN 24058963.
\newblock \doi{10.1016/j.ifacol.2015.06.141}.
\newblock URL
  \url{https://linkinghub.elsevier.com/retrieve/pii/S2405896315003808}.

\bibitem[Rumelhart et~al.(1986)Rumelhart, Hinton, and
  Williams]{rumelhart_learning_1986}
David~E. Rumelhart, Geoffrey~E. Hinton, and Ronald~J. Williams.
\newblock Learning representations by back-propagating errors.
\newblock \emph{Nature}, 323\penalty0 (6088):\penalty0 533--536, October 1986.
\newblock ISSN 1476-4687.
\newblock \doi{10.1038/323533a0}.
\newblock URL \url{https://www.nature.com/articles/323533a0}.
\newblock Number: 6088 Publisher: Nature Publishing Group.

\bibitem[Ruppert and Abonyi(2020)]{ruppert_integration_2020}
Tamas Ruppert and Janos Abonyi.
\newblock Integration of real-time locating systems into digital twins.
\newblock \emph{Journal of Industrial Information Integration}, 20:\penalty0
  100174, December 2020.
\newblock ISSN 2452414X.
\newblock \doi{10.1016/j.jii.2020.100174}.
\newblock URL
  \url{https://linkinghub.elsevier.com/retrieve/pii/S2452414X20300492}.

\bibitem[Rácz-Szabó et~al.(2020)Rácz-Szabó, Ruppert, Bántay, Löcklin,
  Jakab, and Abonyi]{racz-szabo_real-time_2020-1}
András Rácz-Szabó, Tamás Ruppert, László Bántay, Andreas Löcklin,
  László Jakab, and János Abonyi.
\newblock Real-{Time} {Locating} {System} in {Production} {Management}.
\newblock \emph{Sensors}, 20\penalty0 (23):\penalty0 6766, November 2020.
\newblock ISSN 1424-8220.
\newblock \doi{10.3390/s20236766}.
\newblock URL \url{https://www.mdpi.com/1424-8220/20/23/6766}.

\bibitem[Sander et~al.(1998)Sander, Ester, Kriegel, and
  Xu]{sander_density-based_1998}
Jörg Sander, Martin Ester, Hans-Peter Kriegel, and Xiaowei Xu.
\newblock Density-{Based} {Clustering} in {Spatial} {Databases}: {The}
  {Algorithm} {GDBSCAN} and {Its} {Applications}.
\newblock \emph{Data Mining and Knowledge Discovery}, 2\penalty0 (2):\penalty0
  169--194, June 1998.
\newblock ISSN 1573-756X.
\newblock \doi{10.1023/A:1009745219419}.
\newblock URL \url{https://doi.org/10.1023/A:1009745219419}.

\bibitem[Sanghai et~al.(2005)Sanghai, Domingos, and
  Weld]{sanghai_relational_2005}
S.~Sanghai, P.~Domingos, and D.~Weld.
\newblock Relational {Dynamic} {Bayesian} {Networks}.
\newblock \emph{Journal of Artificial Intelligence Research}, 24:\penalty0
  759--797, December 2005.
\newblock ISSN 1076-9757.
\newblock \doi{10.1613/jair.1625}.
\newblock URL \url{https://jair.org/index.php/jair/article/view/10431}.

\bibitem[Sato(1995)]{sato_statistical_1995}
Taisuke Sato.
\newblock A {Statistical} {Learning} {Method} for {Logic} {Programs} with
  {Distribution} {Semantics}.
\newblock In \emph{In {Proceedings} of the 12th {International} {Conference} on
  {Logic} {Programming} (iclp’95}, pages 715--729. MIT Press, 1995.

\bibitem[Savitzky and Golay(1964)]{savitzky_smoothing_1964}
Abraham. Savitzky and M.~J.~E. Golay.
\newblock Smoothing and {Differentiation} of {Data} by {Simplified} {Least}
  {Squares} {Procedures}.
\newblock \emph{Analytical Chemistry}, 36\penalty0 (8):\penalty0 1627--1639,
  July 1964.
\newblock ISSN 0003-2700, 1520-6882.
\newblock \doi{10.1021/ac60214a047}.
\newblock URL \url{https://pubs.acs.org/doi/abs/10.1021/ac60214a047}.

\bibitem[Schabus and Scholz(2015)]{schabus_geographic_2015-1}
Stefan Schabus and Johannes Scholz.
\newblock Geographic {Information} {Science} and {Technology} as {Key}
  {Approach} to unveil the {Potential} of {Industry} 4.0 - {How} {Location} and
  {Time} {Can} {Support} {Smart} {Manufacturing}:.
\newblock In \emph{Proceedings of the 12th {International} {Conference} on
  {Informatics} in {Control}, {Automation} and {Robotics}}, pages 463--470,
  Colmar, Alsace, France, 2015. SCITEPRESS - Science and and Technology
  Publications.
\newblock ISBN 978-989-758-122-9 978-989-758-123-6.
\newblock \doi{10.5220/0005510804630470}.
\newblock URL
  \url{http://www.scitepress.org/DigitalLibrary/Link.aspx?doi=10.5220/0005510804630470}.

\bibitem[Schultz et~al.(2018)Schultz, Bhatt, Suchan, and
  Wałęga]{benzmuller_answer_2018}
Carl Schultz, Mehul Bhatt, Jakob Suchan, and Przemysław~Andrzej Wałęga.
\newblock Answer {Set} {Programming} {Modulo} ‘{Space}-{Time}’.
\newblock In Christoph Benzmüller, Francesco Ricca, Xavier Parent, and Dumitru
  Roman, editors, \emph{Rules and {Reasoning}}, volume 11092, pages 318--326.
  Springer International Publishing, Cham, 2018.
\newblock ISBN 978-3-319-99905-0 978-3-319-99906-7.
\newblock URL \url{http://link.springer.com/10.1007/978-3-319-99906-7_24}.

\bibitem[Schulz et~al.(2009)Schulz, Suntisrivaraporn, Baader, and
  Boeker]{schulz_snomed_2009}
Stefan Schulz, Boontawee Suntisrivaraporn, Franz Baader, and Martin Boeker.
\newblock {SNOMED} reaching its adolescence: {Ontologists}’ and logicians’
  health check.
\newblock \emph{International Journal of Medical Informatics}, 78:\penalty0
  S86--S94, April 2009.
\newblock ISSN 1386-5056.
\newblock \doi{10.1016/j.ijmedinf.2008.06.004}.
\newblock URL
  \url{https://www.sciencedirect.com/science/article/pii/S1386505608000919}.

\bibitem[Schwindt and Zimmermann(2015{\natexlab{a}})]{schwindt_handbook_2015}
Christoph Schwindt and Jürgen Zimmermann, editors.
\newblock \emph{Handbook on {Project} {Management} and {Scheduling} {Vol}. 2}.
\newblock Springer International Publishing, Cham, 2015{\natexlab{a}}.
\newblock ISBN 978-3-319-05914-3 978-3-319-05915-0.
\newblock \doi{10.1007/978-3-319-05915-0}.
\newblock URL \url{http://link.springer.com/10.1007/978-3-319-05915-0}.

\bibitem[Schwindt and Zimmermann(2015{\natexlab{b}})]{schwindt_handbook_2015-1}
Christoph Schwindt and Jürgen Zimmermann, editors.
\newblock \emph{Handbook on {Project} {Management} and {Scheduling} {Vol}.1}.
\newblock Springer International Publishing, Cham, 2015{\natexlab{b}}.
\newblock ISBN 978-3-319-05442-1 978-3-319-05443-8.
\newblock \doi{10.1007/978-3-319-05443-8}.
\newblock URL \url{http://link.springer.com/10.1007/978-3-319-05443-8}.

\bibitem[{Sepp Hochreiter} and {Jurgen
  Schmidhuber}(1997)]{sepp_hochreiter_long_1997}
{Sepp Hochreiter} and {Jurgen Schmidhuber}.
\newblock Long {Short} {Term} {Memory}.
\newblock \emph{Neural Computation}, 9\penalty0 (8):\penalty0 1735--1780, 1997.

\bibitem[Shahriar et~al.(2016)Shahriar, Smith, Rahman, Freeman, Hills,
  Rawnsley, Henry, and Bishop-Hurley]{shahriar_detecting_2016}
Md.~Sumon Shahriar, Daniel Smith, Ashfaqur Rahman, Mark Freeman, James Hills,
  Richard Rawnsley, Dave Henry, and Greg Bishop-Hurley.
\newblock Detecting heat events in dairy cows using accelerometers and
  unsupervised learning.
\newblock \emph{Computers and Electronics in Agriculture}, 128:\penalty0
  20--26, October 2016.
\newblock ISSN 01681699.
\newblock \doi{10.1016/j.compag.2016.08.009}.
\newblock URL
  \url{https://linkinghub.elsevier.com/retrieve/pii/S0168169916306093}.

\bibitem[Shi et~al.(2020)Shi, Chen, Ma, Mao, Zhang, and Zhang]{shi_neural_2020}
Shaoyun Shi, Hanxiong Chen, Weizhi Ma, Jiaxin Mao, Min Zhang, and Yongfeng
  Zhang.
\newblock Neural {Logic} {Reasoning}.
\newblock In \emph{Proceedings of the 29th {ACM} {International} {Conference}
  on {Information} \& {Knowledge} {Management}}, {CIKM} '20, pages 1365--1374,
  New York, NY, USA, October 2020. Association for Computing Machinery.
\newblock ISBN 978-1-4503-6859-9.
\newblock \doi{10.1145/3340531.3411949}.
\newblock URL \url{https://doi.org/10.1145/3340531.3411949}.

\bibitem[Singer(2021)]{singer_rise_2021}
Gadi Singer.
\newblock The {Rise} of {Cognitive} {AI}, May 2021.
\newblock URL
  \url{https://towardsdatascience.com/the-rise-of-cognitive-ai-a29d2b724ccc}.

\bibitem[Skarlatidis et~al.(2015)Skarlatidis, Artikis, Filippou, and
  Paliouras]{skarlatidis_probabilistic_2015}
Anastasios Skarlatidis, Alexander Artikis, Jason Filippou, and Georgios
  Paliouras.
\newblock A {Probabilistic} {Logic} {Programming} {Event} {Calculus}.
\newblock \emph{Theory and Practice of Logic Programming}, 15\penalty0
  (2):\penalty0 213--245, March 2015.
\newblock ISSN 1471-0684, 1475-3081.
\newblock \doi{10.1017/S1471068413000690}.
\newblock URL \url{http://arxiv.org/abs/1204.1851}.
\newblock arXiv:1204.1851 [cs].

\bibitem[{SmartX HUB}(2022)]{smartx_hub_industrial_2022}
{SmartX HUB}.
\newblock Industrial – {IIoT} and {RFID} – {Improve} {Tracking},
  {Workflows}, and {Safety} by {SmartX} {HUB}, 2022.
\newblock URL \url{https://smartxhub.com/industrial/}.

\bibitem[Smith et~al.(2020)Smith, McNally, Little, Ingham, and
  Schmoelzl]{smith_automatic_2020}
Daniel Smith, Jody McNally, Bryce Little, Aaron Ingham, and Sabine Schmoelzl.
\newblock Automatic detection of parturition in pregnant ewes using a
  three-axis accelerometer.
\newblock \emph{Computers and Electronics in Agriculture}, 173:\penalty0
  105392, June 2020.
\newblock ISSN 01681699.
\newblock \doi{10.1016/j.compag.2020.105392}.
\newblock URL
  \url{https://linkinghub.elsevier.com/retrieve/pii/S0168169919322872}.

\bibitem[Song et~al.(2021)Song, Li, Yang, Bai, Hu, Zhang, and
  Zhang]{song_intelligent_2021}
Qisong Song, Shaobo Li, Jing Yang, Qiang Bai, Jianjun Hu, Xingxing Zhang, and
  Ansi Zhang.
\newblock Intelligent {Optimization} {Algorithm}-{Based} {Path} {Planning} for
  a {Mobile} {Robot}.
\newblock \emph{Computational Intelligence and Neuroscience}, 2021:\penalty0
  1--17, September 2021.
\newblock ISSN 1687-5273, 1687-5265.
\newblock \doi{10.1155/2021/8025730}.
\newblock URL \url{https://www.hindawi.com/journals/cin/2021/8025730/}.

\bibitem[Song et~al.(2016)Song, Kanasugi, and
  Shibasaki]{song_deeptransport_2016}
Xuan Song, Hiroshi Kanasugi, and Ryosuke Shibasaki.
\newblock {DeepTransport}: {Prediction} and {Simulation} of {Human} {Mobility}
  and {Transportation} {Mode} at a {Citywide} {Level}.
\newblock \emph{International Joint Conference on Artificial Intelligence
  (IJCAI-16)}, pages 2618--2624, 2016.

\bibitem[Stetter(2021)]{stetter_wearable_2021}
Bernd~Josef Stetter.
\newblock Wearable {Sensors} and {Machine} {Learning} based {Human} {Movement}
  {Analysis} – {Applications} in {Sports} and {Medicine}.
\newblock Karlsruher Institut für Technologie (KIT), 2021.
\newblock URL \url{https://publikationen.bibliothek.kit.edu/1000131001}.

\bibitem[Susto et~al.(2015)Susto, Schirru, Pampuri, McLoone, and
  Beghi]{susto_machine_2015}
Gian~Antonio Susto, Andrea Schirru, Simone Pampuri, Sean McLoone, and
  Alessandro Beghi.
\newblock Machine {Learning} for {Predictive} {Maintenance}: {A} {Multiple}
  {Classifier} {Approach}.
\newblock \emph{IEEE Transactions on Industrial Informatics}, 11\penalty0
  (3):\penalty0 812--820, June 2015.
\newblock ISSN 1551-3203, 1941-0050.
\newblock \doi{10.1109/TII.2014.2349359}.
\newblock URL \url{http://ieeexplore.ieee.org/document/6879441/}.

\bibitem[Sutskever et~al.(2014)Sutskever, Vinyals, and
  Le]{sutskever_sequence_2014}
Ilya Sutskever, Oriol Vinyals, and Quoc~V Le.
\newblock Sequence to {Sequence} {Learning} with {Neural} {Networks}.
\newblock In \emph{Advances in {Neural} {Information} {Processing} {Systems}},
  volume~27. Curran Associates, Inc., 2014.
\newblock URL
  \url{https://proceedings.neurips.cc/paper/2014/hash/a14ac55a4f27472c5d894ec1c3c743d2-Abstract.html}.

\bibitem[Sutton and Barto(2018)]{sutton_reinforcement_2018}
Richard~S. Sutton and Andrew~G. Barto.
\newblock \emph{Reinforcement {Learning}: {An} {Introduction}}.
\newblock Adaptive {Computation} and {Machine} {Learning} series. A Bradford
  Book, Cambridge, MA, USA, 2 edition, November 2018.
\newblock ISBN 978-0-262-03924-6.

\bibitem[Syafrudin et~al.(2018)Syafrudin, Alfian, Fitriyani, and
  Rhee]{syafrudin_performance_2018}
Muhammad Syafrudin, Ganjar Alfian, Norma~Latif Fitriyani, and Jongtae Rhee.
\newblock Performance {Analysis} of {IoT}-{Based} {Sensor}, {Big} {Data}
  {Processing}, and {Machine} {Learning} {Model} for {Real}-{Time} {Monitoring}
  {System} in {Automotive} {Manufacturing}.
\newblock \emph{Sensors}, 18\penalty0 (9):\penalty0 2946, September 2018.
\newblock ISSN 1424-8220.
\newblock \doi{10.3390/s18092946}.
\newblock URL \url{https://www.mdpi.com/1424-8220/18/9/2946}.
\newblock Number: 9 Publisher: Multidisciplinary Digital Publishing Institute.

\bibitem[Tao et~al.(2018)Tao, Qi, Liu, and Kusiak]{tao_data-driven_2018}
Fei Tao, Qinglin Qi, Ang Liu, and Andrew Kusiak.
\newblock Data-driven smart manufacturing.
\newblock \emph{Journal of Manufacturing Systems}, 48:\penalty0 157--169, July
  2018.
\newblock ISSN 0278-6125.
\newblock \doi{10.1016/j.jmsy.2018.01.006}.
\newblock URL
  \url{https://www.sciencedirect.com/science/article/pii/S0278612518300062}.

\bibitem[Tao et~al.(2019)Tao, Sui, Liu, Qi, Zhang, Song, Guo, Lu, and
  Nee]{tao_digital_2019}
Fei Tao, Fangyuan Sui, Ang Liu, Qinglin Qi, Meng Zhang, Boyang Song, Zirong
  Guo, Stephen C.-Y. Lu, and A.~Y.~C. Nee.
\newblock Digital twin-driven product design framework.
\newblock \emph{International Journal of Production Research}, 57\penalty0
  (12):\penalty0 3935--3953, June 2019.
\newblock ISSN 0020-7543, 1366-588X.
\newblock \doi{10.1080/00207543.2018.1443229}.
\newblock URL
  \url{https://www.tandfonline.com/doi/full/10.1080/00207543.2018.1443229}.

\bibitem[Tao et~al.(2005)Tao, Cheng, Xiao, Ngai, Kao, and
  Prabhakar]{tao_indexing_2005}
Yufei Tao, Reynold Cheng, Xiaokui Xiao, Wang~Kay Ngai, Ben Kao, and Sunil
  Prabhakar.
\newblock Indexing multi-dimensional uncertain data with arbitrary probability
  density functions.
\newblock In \emph{Proceedings of the 31st international conference on {Very}
  large data bases}, {VLDB} '05, pages 922--933, Trondheim, Norway, August
  2005. VLDB Endowment.
\newblock ISBN 978-1-59593-154-2.

\bibitem[Tawfik and Neufeld(2000)]{tawfik_temporal_2000}
Ahmed~Y. Tawfik and Eric~M. Neufeld.
\newblock Temporal {Reasoning} and {Bayesian} {Networks}.
\newblock \emph{Computational Intelligence}, 16\penalty0 (3):\penalty0
  349--377, August 2000.
\newblock ISSN 0824-7935, 1467-8640.
\newblock \doi{10.1111/0824-7935.00116}.
\newblock URL
  \url{https://onlinelibrary.wiley.com/doi/10.1111/0824-7935.00116}.

\bibitem[Tolmach et~al.(2022)Tolmach, Li, Lin, Liu, and
  Li]{tolmach_survey_2022}
Palina Tolmach, Yi~Li, Shang-Wei Lin, Yang Liu, and Zengxiang Li.
\newblock A {Survey} of {Smart} {Contract} {Formal} {Specification} and
  {Verification}.
\newblock \emph{ACM Computing Surveys}, 54\penalty0 (7):\penalty0 1--38,
  September 2022.
\newblock ISSN 0360-0300, 1557-7341.
\newblock \doi{10.1145/3464421}.
\newblock URL \url{https://dl.acm.org/doi/10.1145/3464421}.

\bibitem[Tran et~al.(2015)Tran, Bourdev, Fergus, Torresani, and
  Paluri]{tran_learning_2015}
Du~Tran, Lubomir Bourdev, Rob Fergus, Lorenzo Torresani, and Manohar Paluri.
\newblock Learning {Spatiotemporal} {Features} with {3D} {Convolutional}
  {Networks}.
\newblock In \emph{2015 {IEEE} {International} {Conference} on {Computer}
  {Vision} ({ICCV})}, pages 4489--4497, Santiago, Chile, December 2015. IEEE.
\newblock ISBN 978-1-4673-8391-2.
\newblock \doi{10.1109/ICCV.2015.510}.
\newblock URL \url{http://ieeexplore.ieee.org/document/7410867/}.

\bibitem[Tsilionis et~al.(2019)Tsilionis, Artikis, and
  Paliouras]{tsilionis_incremental_2019}
Efthimis Tsilionis, Alexander Artikis, and Georgios Paliouras.
\newblock Incremental {Event} {Calculus} for {Run}-{Time} {Reasoning}.
\newblock In \emph{Proceedings of the 13th {ACM} {International} {Conference}
  on {Distributed} and {Event}-based {Systems}}, {DEBS} '19, pages 79--90, New
  York, NY, USA, June 2019. Association for Computing Machinery.
\newblock ISBN 978-1-4503-6794-3.
\newblock \doi{10.1145/3328905.3329504}.
\newblock URL \url{https://doi.org/10.1145/3328905.3329504}.

\bibitem[Uhlemann et~al.(2017)Uhlemann, Schock, Lehmann, Freiberger, and
  Steinhilper]{uhlemann_digital_2017}
Thomas H.-J. Uhlemann, Christoph Schock, Christian Lehmann, Stefan Freiberger,
  and Rolf Steinhilper.
\newblock The {Digital} {Twin}: {Demonstrating} the {Potential} of {Real}
  {Time} {Data} {Acquisition} in {Production} {Systems}.
\newblock \emph{Procedia Manufacturing}, 9:\penalty0 113--120, 2017.
\newblock ISSN 23519789.
\newblock \doi{10.1016/j.promfg.2017.04.043}.
\newblock URL
  \url{https://linkinghub.elsevier.com/retrieve/pii/S2351978917301610}.

\bibitem[Uhlenkamp et~al.(2019)Uhlenkamp, Hribernik, Wellsandt, and
  Thoben]{uhlenkamp_digital_2019}
Jan-Frederik Uhlenkamp, Karl Hribernik, Stefan Wellsandt, and Klaus-Dieter
  Thoben.
\newblock Digital {Twin} {Applications} : {A} first systemization of their
  dimensions.
\newblock In \emph{2019 {IEEE} {International} {Conference} on {Engineering},
  {Technology} and {Innovation} ({ICE}/{ITMC})}, pages 1--8, Valbonne
  Sophia-Antipolis, France, June 2019. IEEE.
\newblock ISBN 978-1-72813-401-7.
\newblock \doi{10.1109/ICE.2019.8792579}.
\newblock URL \url{https://ieeexplore.ieee.org/document/8792579/}.

\bibitem[Vachalek et~al.(2017)Vachalek, Bartalsky, Rovny, Sismisova, Morhac,
  and Loksik]{vachalek_digital_2017}
Jan Vachalek, Lukas Bartalsky, Oliver Rovny, Dana Sismisova, Martin Morhac, and
  Milan Loksik.
\newblock The digital twin of an industrial production line within the industry
  4.0 concept.
\newblock In \emph{2017 21st {International} {Conference} on {Process}
  {Control} ({PC})}, pages 258--262, Strbske Pleso, Slovakia, June 2017. IEEE.
\newblock ISBN 978-1-5386-4011-1.
\newblock \doi{10.1109/PC.2017.7976223}.
\newblock URL \url{http://ieeexplore.ieee.org/document/7976223/}.

\bibitem[Valdés and Güting(2019)]{valdes_framework_2019}
Fabio Valdés and Ralf~Hartmut Güting.
\newblock A framework for efficient multi-attribute movement data analysis.
\newblock \emph{The VLDB Journal}, 28\penalty0 (4):\penalty0 427--449, August
  2019.
\newblock ISSN 1066-8888, 0949-877X.
\newblock \doi{10.1007/s00778-018-0525-6}.
\newblock URL \url{http://link.springer.com/10.1007/s00778-018-0525-6}.

\bibitem[Valle et~al.(2009)Valle, Ceri, Harmelen, and Fensel]{valle_its_2009}
Emanuele~Della Valle, Stefano Ceri, Frank~van Harmelen, and Dieter Fensel.
\newblock It's a {Streaming} {World}! {Reasoning} upon {Rapidly} {Changing}
  {Information}.
\newblock \emph{IEEE Intelligent Systems}, 24\penalty0 (6):\penalty0 83--89,
  November 2009.
\newblock ISSN 1541-1672.
\newblock \doi{10.1109/MIS.2009.125}.
\newblock URL \url{http://ieeexplore.ieee.org/document/5372206/}.

\bibitem[VanDerHorn and Mahadevan(2021)]{vanderhorn_digital_2021}
Eric VanDerHorn and Sankaran Mahadevan.
\newblock Digital {Twin}: {Generalization}, characterization and
  implementation.
\newblock \emph{Decision Support Systems}, 145:\penalty0 113524, June 2021.
\newblock ISSN 01679236.
\newblock \doi{10.1016/j.dss.2021.113524}.
\newblock URL
  \url{https://linkinghub.elsevier.com/retrieve/pii/S0167923621000348}.

\bibitem[Vaswani et~al.(2017)Vaswani, Shazeer, Parmar, Uszkoreit, Jones, Gomez,
  Kaiser, and Polosukhin]{vaswani_attention_2017-1}
Ashish Vaswani, Noam Shazeer, Niki Parmar, Jakob Uszkoreit, Llion Jones,
  Aidan~N Gomez, Łukasz Kaiser, and Illia Polosukhin.
\newblock Attention is {All} you {Need}.
\newblock In \emph{Advances in {Neural} {Information} {Processing} {Systems}},
  volume~30. Curran Associates, Inc., 2017.
\newblock URL
  \url{https://proceedings.neurips.cc/paper/2017/hash/3f5ee243547dee91fbd053c1c4a845aa-Abstract.html}.

\bibitem[{Viterbi, Andrew}(1967)]{viterbi_andrew_error_1967}
{Viterbi, Andrew}.
\newblock Error bounds for convolutional codes and an asymptotically optimum
  decoding algorithm.
\newblock \emph{IEEE Transactions on Information Theory}, 13\penalty0
  (2):\penalty0 260--269, 1967.

\bibitem[Vlasselaer et~al.(2014)Vlasselaer, Meert, Van Den~Broeck, and
  De~Raedt]{vlasselaer_efficient_2014}
Jonas Vlasselaer, Wannes Meert, Guy Van Den~Broeck, and Luc De~Raedt.
\newblock Efficient probabilistic inference for dynamic relational models.
\newblock In \emph{Proceedings of the 13th {AAAI} {Conference} on {Statistical}
  {Relational} {AI}}, {AAAIWS}'14-13, pages 131--132. AAAI Press, January 2014.

\bibitem[Wan and Song(2018)]{wan_neural_2018}
Fang Wan and Chaoyang Song.
\newblock A {Neural} {Network} {With} {Logical} {Reasoning} {Based} on
  {Auxiliary} {Inputs}.
\newblock \emph{Frontiers in Robotics and AI}, 5:\penalty0 86, July 2018.
\newblock ISSN 2296-9144.
\newblock \doi{10.3389/frobt.2018.00086}.
\newblock URL
  \url{https://www.frontiersin.org/article/10.3389/frobt.2018.00086/full}.

\bibitem[Wang et~al.(2020)Wang, Miwa, and Morikawa]{wang_big_2020}
Di~Wang, Tomio Miwa, and Takayuki Morikawa.
\newblock Big {Trajectory} {Data} {Mining}: {A} {Survey} of {Methods},
  {Applications}, and {Services}.
\newblock \emph{Sensors}, 20\penalty0 (16):\penalty0 4571, January 2020.
\newblock ISSN 1424-8220.
\newblock \doi{10.3390/s20164571}.
\newblock URL \url{https://www.mdpi.com/1424-8220/20/16/4571}.
\newblock Number: 16 Publisher: Multidisciplinary Digital Publishing Institute.

\bibitem[Wang et~al.(2021{\natexlab{a}})Wang, Wu, Li, Gu, Das, and
  Zaniolo]{wang_formal_2021}
Jin Wang, Jiacheng Wu, Mingda Li, Jiaqi Gu, Ariyam Das, and Carlo Zaniolo.
\newblock Formal semantics and high performance in declarative machine learning
  using {Datalog}.
\newblock \emph{The VLDB Journal}, 30\penalty0 (5):\penalty0 859--881,
  September 2021{\natexlab{a}}.
\newblock ISSN 1066-8888, 0949-877X.
\newblock \doi{10.1007/s00778-021-00665-6}.
\newblock URL \url{https://link.springer.com/10.1007/s00778-021-00665-6}.

\bibitem[Wang et~al.(2021{\natexlab{b}})Wang, Yang, and
  Zhang]{wang_location_2021}
Peng Wang, Jing Yang, and Jianpei Zhang.
\newblock Location {Prediction} for {Indoor} {Spaces} based on {Trajectory}
  {Similarity}.
\newblock In \emph{2021 4th {International} {Conference} on {Data} {Science}
  and {Information} {Technology}}, {DSIT} 2021, pages 402--407, New York, NY,
  USA, July 2021{\natexlab{b}}. Association for Computing Machinery.
\newblock ISBN 978-1-4503-9024-8.
\newblock \doi{10.1145/3478905.3478983}.
\newblock URL \url{https://doi.org/10.1145/3478905.3478983}.

\bibitem[Wari and Zhu(2016)]{wari_survey_2016}
Ezra Wari and Weihang Zhu.
\newblock A survey on metaheuristics for optimization in food manufacturing
  industry.
\newblock \emph{Applied Soft Computing}, 46:\penalty0 328--343, September 2016.
\newblock ISSN 15684946.
\newblock \doi{10.1016/j.asoc.2016.04.034}.
\newblock URL
  \url{https://linkinghub.elsevier.com/retrieve/pii/S156849461630182X}.

\bibitem[Wałęga et~al.(2015)Wałęga, Bhatt, and
  Schultz]{calimeri_aspmtqs_2015}
Przemysław~Andrzej Wałęga, Mehul Bhatt, and Carl Schultz.
\newblock {ASPMT}({QS}): {Non}-{Monotonic} {Spatial} {Reasoning} with {Answer}
  {Set} {Programming} {Modulo} {Theories}.
\newblock In Francesco Calimeri, Giovambattista Ianni, and Miroslaw
  Truszczynski, editors, \emph{Logic {Programming} and {Nonmonotonic}
  {Reasoning}}, volume 9345, pages 488--501. Springer International Publishing,
  Cham, 2015.
\newblock ISBN 978-3-319-23263-8 978-3-319-23264-5.
\newblock URL \url{http://link.springer.com/10.1007/978-3-319-23264-5_41}.

\bibitem[{Worximity Technology}(2022)]{worximity_technology_oee_2022}
{Worximity Technology}.
\newblock {OEE} monitoring tool - {Tileboard} by {Worximity}, 2022.
\newblock URL \url{http://www.worximity.com/en/oee-monitoring-tool}.

\bibitem[Wu et~al.(2019{\natexlab{a}})Wu, Wu, Tamar, Russell, Gkioxari, and
  Tian]{wu_bayesian_2019}
Yi~Wu, Yuxin Wu, Aviv Tamar, Stuart Russell, Georgia Gkioxari, and Yuandong
  Tian.
\newblock Bayesian {Relational} {Memory} for {Semantic} {Visual} {Navigation}.
\newblock In \emph{2019 {IEEE}/{CVF} {International} {Conference} on {Computer}
  {Vision} ({ICCV})}, pages 2769--2779, Seoul, Korea (South), October
  2019{\natexlab{a}}. IEEE.
\newblock ISBN 978-1-72814-803-8.
\newblock \doi{10.1109/ICCV.2019.00286}.
\newblock URL \url{https://ieeexplore.ieee.org/document/9009539/}.

\bibitem[Wu et~al.(2019{\natexlab{b}})Wu, Pan, Long, Jiang, and
  Zhang]{wu_graph_2019}
Zonghan Wu, Shirui Pan, Guodong Long, Jing Jiang, and Chengqi Zhang.
\newblock Graph wavenet for deep spatial-temporal graph modeling.
\newblock In \emph{Proceedings of the 28th {International} {Joint} {Conference}
  on {Artificial} {Intelligence}}, {IJCAI}'19, pages 1907--1913, Macao, China,
  August 2019{\natexlab{b}}. AAAI Press.
\newblock ISBN 978-0-9992411-4-1.

\bibitem[Xu et~al.(2019)Xu, Güting, Zheng, and Wolfson]{xu_moving_2019}
Jian~Qiu Xu, Ralf~Hartmut Güting, Yu~Zheng, and Ouri Wolfson.
\newblock Moving {Objects} with {Transportation} {Modes}: {A} {Survey}.
\newblock \emph{Journal of Computer Science and Technology}, 34\penalty0
  (4):\penalty0 709--726, July 2019.
\newblock ISSN 18604749.
\newblock \doi{10.1007/s11390-019-1938-4}.
\newblock Publisher: Springer New York LLC.

\bibitem[Yan et~al.(2010)Yan, Parent, Spaccapietra, and
  Chakraborty]{hutchison_hybrid_2010}
Zhixian Yan, Christine Parent, Stefano Spaccapietra, and Dipanjan Chakraborty.
\newblock A {Hybrid} {Model} and {Computing} {Platform} for {Spatio}-semantic
  {Trajectories}.
\newblock In David Hutchison, Takeo Kanade, Josef Kittler, Jon~M. Kleinberg,
  Friedemann Mattern, John~C. Mitchell, Moni Naor, Oscar Nierstrasz,
  C.~Pandu~Rangan, Bernhard Steffen, Madhu Sudan, Demetri Terzopoulos, Doug
  Tygar, Moshe~Y. Vardi, Gerhard Weikum, Lora Aroyo, Grigoris Antoniou, Eero
  Hyvönen, Annette ten Teije, Heiner Stuckenschmidt, Liliana Cabral, and Tania
  Tudorache, editors, \emph{The {Semantic} {Web}: {Research} and
  {Applications}}, volume 6088, pages 60--75. Springer Berlin Heidelberg,
  Berlin, Heidelberg, 2010.
\newblock ISBN 978-3-642-13485-2 978-3-642-13486-9.
\newblock \doi{10.1007/978-3-642-13486-9_5}.
\newblock URL \url{http://link.springer.com/10.1007/978-3-642-13486-9_5}.
\newblock Series Title: Lecture Notes in Computer Science.

\bibitem[Yang et~al.(2019)Yang, Wu, Wang, Jia, and Li]{yang_how_2019}
Dong Yang, Lingxiao Wu, Shuaian Wang, Haiying Jia, and Kevin~X. Li.
\newblock How big data enriches maritime research – a critical review of
  {Automatic} {Identification} {System} ({AIS}) data applications.
\newblock \emph{Transport Reviews}, 39\penalty0 (6):\penalty0 755--773,
  November 2019.
\newblock ISSN 0144-1647, 1464-5327.
\newblock \doi{10.1080/01441647.2019.1649315}.
\newblock URL
  \url{https://www.tandfonline.com/doi/full/10.1080/01441647.2019.1649315}.

\bibitem[Yang et~al.(2022)Yang, Fang, Gao, Zhou, Li, Jin, and
  Song]{yang_obstacle_2022}
Fan Yang, Xi~Fang, Fei Gao, Xianjin Zhou, Hao Li, Hongbin Jin, and Yu~Song.
\newblock Obstacle {Avoidance} {Path} {Planning} for {UAV} {Based} on
  {Improved} {RRT} {Algorithm}.
\newblock \emph{Discrete Dynamics in Nature and Society}, 2022:\penalty0 1--9,
  January 2022.
\newblock ISSN 1607-887X, 1026-0226.
\newblock \doi{10.1155/2022/4544499}.
\newblock URL \url{https://www.hindawi.com/journals/ddns/2022/4544499/}.

\bibitem[Yao et~al.(2017)Yao, Zhang, Zhu, Huang, and Bi]{yao_trajectory_2017}
Di~Yao, Chao Zhang, Zhihua Zhu, Jianhui Huang, and Jingping Bi.
\newblock Trajectory clustering via deep representation learning.
\newblock In \emph{2017 {International} {Joint} {Conference} on {Neural}
  {Networks} ({IJCNN})}, pages 3880--3887, May 2017.
\newblock \doi{10.1109/IJCNN.2017.7966345}.
\newblock ISSN: 2161-4407.

\bibitem[Yoo et~al.(2013)Yoo, Fitch, and Sukkarieh]{yoo_provably-correct_2013}
Chanyeol Yoo, Robert Fitch, and Salah Sukkarieh.
\newblock Provably-correct stochastic motion planning with safety constraints.
\newblock In \emph{2013 {IEEE} {International} {Conference} on {Robotics} and
  {Automation}}, pages 981--986, Karlsruhe, Germany, May 2013. IEEE.
\newblock ISBN 978-1-4673-5643-5 978-1-4673-5641-1.
\newblock \doi{10.1109/ICRA.2013.6630692}.
\newblock URL \url{http://ieeexplore.ieee.org/document/6630692/}.

\bibitem[Yu et~al.(2020{\natexlab{a}})Yu, Ma, Ren, Zhao, and
  Yi]{vedaldi_spatio-temporal_2020}
Cunjun Yu, Xiao Ma, Jiawei Ren, Haiyu Zhao, and Shuai Yi.
\newblock Spatio-{Temporal} {Graph} {Transformer} {Networks} for {Pedestrian}
  {Trajectory} {Prediction}.
\newblock In Andrea Vedaldi, Horst Bischof, Thomas Brox, and Jan-Michael Frahm,
  editors, \emph{Computer {Vision} – {ECCV} 2020}, volume 12357, pages
  507--523. Springer International Publishing, Cham, 2020{\natexlab{a}}.
\newblock ISBN 978-3-030-58609-6 978-3-030-58610-2.
\newblock URL \url{https://link.springer.com/10.1007/978-3-030-58610-2_30}.

\bibitem[Yu et~al.(2021)Yu, Selby, Vlahos, Yadav, and
  Lemp]{yu_feature-oriented_2021}
Jiangbo~Gabe Yu, Brent Selby, Nicholas Vlahos, Vivek Yadav, and Jason Lemp.
\newblock A feature-oriented vehicle trajectory data processing scheme for data
  mining: {A} case study for {Statewide} truck parking behaviors.
\newblock \emph{Transportation Research Interdisciplinary Perspectives},
  11:\penalty0 100401, September 2021.
\newblock ISSN 2590-1982.
\newblock \doi{10.1016/j.trip.2021.100401}.
\newblock URL
  \url{https://www.sciencedirect.com/science/article/pii/S2590198221001081}.

\bibitem[Yu et~al.(2020{\natexlab{b}})Yu, Tang, Wang, Wu, Qian, Sun, and
  Xu]{yu_tulsn_2020}
Yong Yu, Haina Tang, Fei Wang, Lin Wu, Tangwen Qian, Tao Sun, and Yongjun Xu.
\newblock {TULSN}: {Siamese} {Network} for {Trajectory}-user {Linking}.
\newblock In \emph{2020 {International} {Joint} {Conference} on {Neural}
  {Networks} ({IJCNN})}, pages 1--8, July 2020{\natexlab{b}}.
\newblock \doi{10.1109/IJCNN48605.2020.9206609}.
\newblock ISSN: 2161-4407.

\bibitem[Zhang et~al.(2020)Zhang, Liu, Chang, Wang, and
  Gao]{zhang_recurrent_2020}
Jianjing Zhang, Hongyi Liu, Qing Chang, Lihui Wang, and Robert~X. Gao.
\newblock Recurrent neural network for motion trajectory prediction in
  human-robot collaborative assembly.
\newblock \emph{CIRP Annals}, 69\penalty0 (1):\penalty0 9--12, January 2020.
\newblock ISSN 0007-8506.
\newblock \doi{10.1016/j.cirp.2020.04.077}.
\newblock URL
  \url{https://www.sciencedirect.com/science/article/pii/S0007850620300998}.

\bibitem[Zhang et~al.(2017)Zhang, Ren, Liu, Sakao, and
  Huisingh]{zhang_framework_2017}
Yingfeng Zhang, Shan Ren, Yang Liu, Tomohiko Sakao, and Donald Huisingh.
\newblock A framework for {Big} {Data} driven product lifecycle management.
\newblock \emph{Journal of Cleaner Production}, 159:\penalty0 229--240, August
  2017.
\newblock ISSN 0959-6526.
\newblock \doi{10.1016/j.jclepro.2017.04.172}.
\newblock URL
  \url{https://www.sciencedirect.com/science/article/pii/S0959652617309150}.

\bibitem[Zhao et~al.(2019)Zhao, Feng, Xu, Xia, Chen, Sun, Guo, Jin, and
  Li]{zhao_deepmm_2019}
Kai Zhao, Jie Feng, Zhao Xu, Tong Xia, Lin Chen, Funing Sun, Diansheng Guo,
  Depeng Jin, and Yong Li.
\newblock {DeepMM}: {Deep} {Learning} {Based} {Map} {Matching} with {Data}
  {Augmentation}.
\newblock In \emph{Proceedings of the 27th {ACM} {SIGSPATIAL} {International}
  {Conference} on {Advances} in {Geographic} {Information} {Systems}}, pages
  452--455, Chicago IL USA, November 2019. ACM.
\newblock ISBN 978-1-4503-6909-1.
\newblock \doi{10.1145/3347146.3359090}.
\newblock URL \url{https://dl.acm.org/doi/10.1145/3347146.3359090}.

\bibitem[Zheng(2015)]{zheng_trajectory_2015-1}
Yu~Zheng.
\newblock Trajectory {Data} {Mining}: {An} {Overview}.
\newblock \emph{ACM Transactions on Intelligent Systems and Technology},
  6\penalty0 (3):\penalty0 29:1--29:41, May 2015.
\newblock ISSN 2157-6904.
\newblock \doi{10.1145/2743025}.
\newblock URL \url{https://doi.org/10.1145/2743025}.

\bibitem[Zheng et~al.(2010)Zheng, Chen, Li, Xie, and
  Ma]{zheng_understanding_2010}
Yu~Zheng, Yukun Chen, Quannan Li, Xing Xie, and Wei-Ying Ma.
\newblock Understanding transportation modes based on {GPS} data for web
  applications.
\newblock \emph{ACM Transactions on the Web}, 4\penalty0 (1):\penalty0 1--36,
  January 2010.
\newblock ISSN 1559-1131, 1559-114X.
\newblock \doi{10.1145/1658373.1658374}.
\newblock URL \url{https://dl.acm.org/doi/10.1145/1658373.1658374}.

\bibitem[Zhong et~al.(2014)Zhong, Huang, Dai, and Zhang]{zhong_mining_2014}
Ray~Y. Zhong, George~Q. Huang, Q.~Y. Dai, and T.~Zhang.
\newblock Mining {SOTs} and dispatching rules from {RFID}-enabled real-time
  shopfloor production data.
\newblock \emph{Journal of Intelligent Manufacturing}, 25\penalty0
  (4):\penalty0 825--843, August 2014.
\newblock ISSN 1572-8145.
\newblock \doi{10.1007/s10845-012-0721-y}.
\newblock URL \url{https://doi.org/10.1007/s10845-012-0721-y}.

\bibitem[Zhou et~al.(2018)Zhou, Gao, Trajcevski, Zhang, Zhong, and
  Zhang]{zhou_trajectory-user_2018}
Fan Zhou, Qiang Gao, Goce Trajcevski, Kunpeng Zhang, Ting Zhong, and Fengli
  Zhang.
\newblock Trajectory-{User} {Linking} via {Variational} {AutoEncoder}.
\newblock In \emph{Proceedings of the {Twenty}-{Seventh} {International}
  {Joint} {Conference} on {Artificial} {Intelligence}}, pages 3212--3218,
  Stockholm, Sweden, July 2018. International Joint Conferences on Artificial
  Intelligence Organization.
\newblock ISBN 978-0-9992411-2-7.
\newblock \doi{10.24963/ijcai.2018/446}.
\newblock URL \url{https://www.ijcai.org/proceedings/2018/446}.

\bibitem[Zhou et~al.(2021)Zhou, Dai, Gao, Wang, and
  Zhong]{zhou_fan_self-supervised_2021}
Fan Zhou, Yurou Dai, Qiang Gao, Pengyu Wang, and Ting Zhong.
\newblock Self-supervised human mobility learning for next location prediction
  and trajectory classification.
\newblock \emph{Knowledge Based Systems}, 228, 2021.
\newblock \doi{10.1016/j.knosys.2021.107214}.
\newblock URL
  \url{https://reader.elsevier.com/reader/sd/pii/S0950705121004767?token=B74B486F29448DEC07EA95C8DFAED53E42FB117B28D91D93806BFA60F2A2B4DF8086E71D13A7731775B6A0502E301D4B&originRegion=us-east-1&originCreation=20220204011551}.

\bibitem[Zhou et~al.(2022)Zhou, Zheng, Huang, Hao, Li, and
  Zhao]{zhou_graph_2022}
Yu~Zhou, Haixia Zheng, Xin Huang, Shufeng Hao, Dengao Li, and Jumin Zhao.
\newblock Graph {Neural} {Networks}: {Taxonomy}, {Advances}, and {Trends}.
\newblock \emph{ACM Transactions on Intelligent Systems and Technology},
  13\penalty0 (1):\penalty0 15:1--15:54, January 2022.
\newblock ISSN 2157-6904.
\newblock \doi{10.1145/3495161}.
\newblock URL \url{https://doi.org/10.1145/3495161}.

\bibitem[Zhuang et~al.(2021)Zhuang, Miao, Liu, and
  Xiong]{zhuang_connotation_2021}
Cunbo Zhuang, Tian Miao, Jianhua Liu, and Hui Xiong.
\newblock The connotation of digital twin, and the construction and application
  method of shop-floor digital twin.
\newblock \emph{Robotics and Computer-Integrated Manufacturing}, 68:\penalty0
  102075, April 2021.
\newblock ISSN 07365845.
\newblock \doi{10.1016/j.rcim.2020.102075}.
\newblock URL
  \url{https://linkinghub.elsevier.com/retrieve/pii/S0736584520302854}.

\bibitem[Özçep et~al.(2014)Özçep, Möller, and
  Neuenstadt]{lutz_stream-temporal_2014}
Özgür~Lütfü Özçep, Ralf Möller, and Christian Neuenstadt.
\newblock A {Stream}-{Temporal} {Query} {Language} for {Ontology} {Based}
  {Data} {Access}.
\newblock In Carsten Lutz and Michael Thielscher, editors, \emph{{KI} 2014:
  {Advances} in {Artificial} {Intelligence}}, volume 8736, pages 183--194.
  Springer International Publishing, Cham, 2014.
\newblock ISBN 978-3-319-11205-3 978-3-319-11206-0.
\newblock URL \url{http://link.springer.com/10.1007/978-3-319-11206-0_18}.

\end{thebibliography}
}

\end{document}